\documentclass[square,numbers]{article}

\usepackage[final]{neurips_2024}

\usepackage[utf8]{inputenc} 
\usepackage[T1]{fontenc}    
\usepackage{hyperref}       
\usepackage{url}            
\usepackage{booktabs}       
\usepackage{amsfonts}       
\usepackage{nicefrac}       
\usepackage{microtype}      
\usepackage{xcolor}         
\usepackage{times}
\usepackage{soul}
\usepackage{graphicx}
\usepackage{amsmath}
\usepackage{amsthm}
\usepackage{amssymb}
\usepackage{algorithm}
\usepackage{algorithmic}
\usepackage{subfigure}
\usepackage{multirow}
\usepackage{wrapfig}
\usepackage{colortbl}

\title{Skill Path: Unveiling Language Skills from\\ Circuit Graphs}

\author{%
  Hang Chen \\
  School of Computer Science and Technology\\
 Xi'an Jiaotong University\\
  \texttt{albert2123@stu.xjtu.edu.cn} \\
  \And
  Jiaying Zhu\\
  School of Computer Science and Engineering\\
 The Chinese University of Hong Kong\\
  \texttt{zhujy0725@cse.cuhk.edu.hk} \\
  \AND
    Xinyu Yang\\
  School of Computer Science and Technology\\
 Xi'an Jiaotong University\\
  \texttt{yxyphd@mail.xjtu.edu.cn} \\
  \And
  Wenya Wang\thanks{Corresponding author}\\
  School of Computer Science and Engineering\\
 Nanyang Technological University\\
  \texttt{wangwy@ntu.edu.sg} \\
}

\begin{document}

\maketitle

\begin{abstract}
Circuit graph discovery has emerged as a fundamental approach to elucidating the skill mechanistic of language models. Despite the output faithfulness of circuit graphs, they suffer from atomic ablation, which causes the loss of causal dependencies between connected components. In addition, their discovery process, designed to preserve output faithfulness, inadvertently captures extraneous effects other than an isolated target skill. To alleviate these challenges, we introduce \textbf{skill paths}, which offers a more refined and compact representation by isolating individual skills within a linear chain of components.  
To enable skill path extracting from circuit graphs, we propose a three-step framework, consisting of \textbf{decomposition}, \textbf{pruning}, and \textbf{post-pruning causal mediation}. In particular, we offer a complete linear decomposition of the transformer model which leads to a disentangled computation graph. After pruning, we further adopt causal analysis techniques, including counterfactuals and interventions, to extract the final skill paths from the circuit graph.  
To underscore the significance of skill paths, we investigate three generic language skills—Previous Token Skill, Induction Skill, and In-Context Learning Skill—using our framework. Experiments support two crucial properties of these skills, namely stratification and inclusiveness.  Our codes are available at: \url{https://github.com/Zodiark-ch/Language-Skill-of-LLMs}.
\end{abstract}

\section{Introduction}
Circuit Discovery~\citep{elhage2021mathematical,bhaskar2024finding} is pivotal for comprehending language model functionality. Current methods~\citep{yao2024knowledge,syed2023attribution,bhaskar2024finding} achieve this by selectively pruning low-effect edges or nodes within the computational graph, ensuring a  precise circuit graph that preserve task outputs, thereby elucidating the model's task execution mechanism. Typically, these tasks align with specific language model skills, such as the induction task~\citep{conmy2023towards}, greater than task~\citep{hanna2024does}, and reverse task~\citep{lindner2023tracr}.

\begin{figure*}
  \begin{center}
    \includegraphics[width=\textwidth]{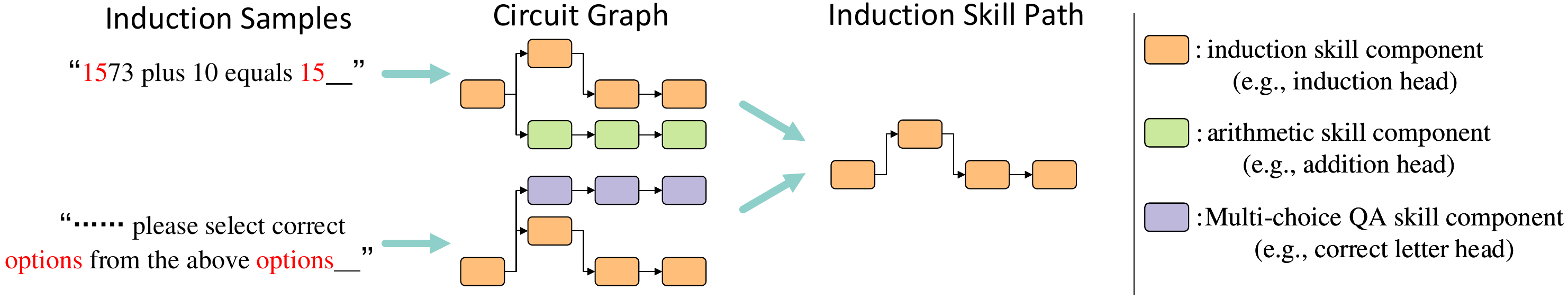}
  \end{center}
  \caption{The difference and correlation between skill path and circuit graph. We use the induction dataset as an example and present two types of samples of induction. When the induction dataset contains a certain number of samples related to arithmetic skills, the final circuit graph may contain parts of the paths for arithmetic skills in addition to the induction skill, as may those samples containing multi-choice skills or other potential skills. We verified in Appendix~\ref{suppObgdf} that even cross-domain randomly sampled datasets still have the confounding of other skills.}
  \label{figintro}
\end{figure*}

Although the circuit graph ensures output faithfulness, capturing the target skill mechanism has two \textbf{constraints}:
\begin{itemize}
    \item \textbf{1)} Current circuit graphs, particularly derived from real-world datasets, include unrelated skills to the intended task skill~\citep{arora2023theoryemergencecomplexskills}, as depicted in Figure~\ref{figintro}. 
    \item \textbf{2)} These graphs use edge- or node-ablation pruning, which ignores causal dependencies between synergistic components crucial for a skill~\citep{mueller2024missedcausesambiguouseffects}. However, due to MLPs, the computational graph is not fully linear, making it difficult to isolate the effect of a path that contains multiple components. 
\end{itemize}
To effectively isolate the target skill mechanism, we introduce the \textbf{skill path}, a compact and finer ``subcircuit'' consisting of a linear sequence of components, which reflects the precise location of a target skill in circuits. Additionally, skill paths naturally facilitate the exploration of skill evolution, such as how advanced skills evolve from basic skills.

Therefore, we propose a novel framework consisting of 3 steps, namely \textbf{Decomposition, Pruning,} and \textbf{Post-pruning Causal Mediation}, to extract skill paths from circuit graphs. 
Specifically, we present the \textbf{compensation component}, enabling a linear breakdown of \text{MLP} ($\text{mlp}(\text{attn}(x)+x)$) inputs into $\text{mlp}(\text{attn}(x))$ and $\text{mlp}(x)$, thus isolating paths within the same \text{MLP} into separate activations. Furthermore, we define \textbf{ functional components} to encapsulate all fundamental components of the computation graph, ensuring each path is constituted by these functionally identical components (\textbf{Decomposition}). Based on the functional components, we collected a large number of samples containing the target skill and obtained the circuit graph sets using existing pruning strategies with activation patching (\textbf{Pruning}). For these circuit graphs, we use counterfactuals and interventions to remove the paths responding to other skill effects (\textbf{Post-pruning Causal Mediation}), and the resulting \textbf{skill paths} represent the complete effect of a target skill. 

The advantage of this three-step framework lies in the following: 
\begin{itemize}
    \item 1) \textbf{Decomposition} provides a fully linear factorization of the computational graph, such that the effects of the underlying functional components are isolated. As a result, ablating a specific path does not affect components outside that path, thereby mitigating constraint 2.
    \item 2) \textbf{Post-pruning causal mediation} introduces counterfactual and interventional techniques to remove subcircuits that are irrelevant to the target skill, thus addressing constraint 1.
\end{itemize}

To show the potential capability of skill paths, we select three generic and progressively complex skills which have been introduced in \citep{crosbie2024induction,ren2024identifying,edelman2024evolution}: a) \emph{Previous Token skill} which is responsible for receiving information from the previous token; b) \emph{Induction Skill} which duplicates tokens with the same prefix; and c) \emph{ICL Skill} which perform inference based on similar patterns presented in demonstrations. Using our 3-step framework, we unveil the skill paths of these skills. These skill paths have better interpretability in skill interaction,  providing stronger evidence to confirm 2 conjectures that have long remained unverified: 
\textbf{Stratification}: Simpler language skills reside in shallower
layers, whereas more complex language skills are in deeper layers. 
\textbf{Inclusiveness}: More complex skill paths are formed on top of simpler skill paths. 

In summary, our contributions are threefold.
\begin{itemize}
\item We propose a theoretical derivation for the complete and lossless linear decomposition of computation graphs, which can decouple all decoder-based language models into finer-grained ``functional components''.

\item We propose a novel three-step framework for discovery of skill paths, including complete paths of a generic skill. 

\item Our analysis and experiments verify 2 properties among the Previous Token Skill, Induction Skill, and ICL Skill, which include stratification and inclusiveness. 
\end{itemize}

\section{Related Work}\label{relatedwork} 

Existing work primarily focuses on discovering circuits responsible for processing specific inputs. Specifically, they provide counterfactual text with slight perturbations to the input as patches~\citep{wang2023interpretability}, use interchange ablation methods to assess the causal effect of each edge on the output~\citep{yao2024knowledge}, and apply various pruning strategies~\citep{conmy2023towards, syed2023attribution,bhaskar2024finding} to identify circuits formed by salient edges. For the resulting circuit graph, further fine-grained exploration (e.g., noising and denoising~\citep{heimersheim2024useinterpretactivationpatching}) is often performed to confirm the mechanisms responsible for different parts of the input text.

Existing work has identified circuits as fine-grained behaviors that respond to specific inputs. In contrast, the circuits we identify encompass the complete global skill. For example, in the IOI~\citep{wang2023interpretability} samples, existing work has identified induction heads. Induction heads may serve as instantiations of the induction skill, but they represent only a partial mechanism derived from specific input samples. Different input samples can lead to the identification of distinct induction heads, highlighting their limited generality. In contrast, our work focuses on uncovering the complete circuits underlying the induction skill by identifying the comprehensive skill paths, offering a more holistic understanding of the mechanism.

\section{Method}\label{secmethod}
In this paper, we propose a novel three-step framework to extract the target skill paths.
\begin{itemize}
    \item \textbf{Step 1} (Section~\ref{secmemorycircuit}, ~\ref{secimplecomputation}): We decouple the architecture of language models into completely linear ``functional components''. This results in a \emph{Computation Graph}, $\mathcal{G}$. 
    \item \textbf{Step 2} (Section~\ref{secgreedysearch}): Utilizing the existing pruning strategies, we transform the computational graph into a \emph{Circuit Graph}, $\mathcal{G}*$. 
    \item \textbf{Step 3} (Section~\ref{seccausaleffect}): We select those paths rendering the most significant causal effect in $\mathcal{G}*$ on the target skill. The final graph formed by the skill paths is named \emph{Skill Graph}, denoted $\mathcal{G}^{S}$. 
\end{itemize}

\subsection{Decomposition}\label{secmemorycircuit}

In each layer of the language model, the input of \text{MLP} includes the input $X$ of this layer and the output $\text{attn}(X)$ of attention, that is, the activation of the \text{MLP} is represented as $\text{mlp}(\text{attn}(X)+X)$. However, due to the presence of nonlinear activations in \text{MLP}, its activation cannot be directly decomposed into $\text{mlp}(\text{attn}(X))$ and $\text{mlp}(X)$, which hinders the ablation of path-level structures. For example, if a path contains a sub-path ``attention $\rightarrow$ \text{MLP}'', its ablation could affect the activations of another path with this \text{MLP}. Therefore, we introduce a \textbf{Compensation Component} $Cps$, which is capable of representing the input to the \text{MLP} in the form of a linear combination, for example:
$\text{mlp}(\text{attn}(X)+X)=\text{mlp}(\text{attn}(X))+\text{mlp}(X)+Cps(X)$. 
In practice, when $\text{attn}(X)$ is decomposed into the sum of each head, there is an additional compensation component (see Table~\ref{tabcircuitindex} and Appendix~\ref{suppdcc} for more details).

Building on the foundation of the Transformer Circuit~\citep{elhage2021mathematical}, we propose a complete decomposition of the transformer model including the \text{MLP} layers with omission of compensation components. Using tensor products ($\otimes$), we can represent any layer of the transformer model:
\begin{align*}\label{eqttranformercircuit}
    \text{output}=&(\text{Id}+\sum_{h \in H}A^{h}\otimes W^{h}_{OV}+\text{Id}\otimes W_{\text{MLP}}+
    \sum_{h \in H}A^{h}\otimes W_{\text{MLP}}W^{h}_{OV})\cdot X
\end{align*}
where $H$ represents the number of attention heads. The matrix $A$ is given by the attention mechanism $A=\text{softmax}((X W_{Q}) (X  W_{K})^{T})$, and $W_{\text{MLP}}$ involves the \text{MLP} operation with activation given by $\text{atv}(X W_{M1}) W_{M2}$. $W_{OV}=W_{O} W_{V}$ refers to an ``output-value'' matrix which computes how each token affects the output if attended to, while $W_{Q}, W_{K}, W_{V}$ are parameter matrices for query, key and value. $W_{M1}$ and $W_{M2}$ are weight parameters in two linear layers. This equation simplifies both the attention and \text{MLP} modules into linear matrix mappings, describing how the paths from input to output for each layer are decoupled into four independent circuits: 
1) $C^{self}= \text{Id} \cdot X$; 
2) $C^{\text{attn}}=\sum_{h \in H}A^{h}\otimes W^{h}_{OV} \cdot X$; 
3) $C^{\text{mlp}}=\text{Id}\otimes W_{\text{MLP}} \cdot X$; 
4) $C^{\text{attn}+\text{mlp}}=\sum_{h \in H}A^{h}\otimes W_{\text{MLP}}W^{h}_{OV} \cdot X$. 

Moreover, $C^{\text{attn}}, C^{\text{mlp}}, C^{\text{attn}+\text{mlp}}$ can be further factorized as:  
\begin{equation}
    C^{\text{attn}/\text{mlp}/\text{attn}+\text{mlp}} = f(X)\cdot W
    \label{eqtmemorycircuitsupp}
\end{equation}
We use $f$ to represent a function that can be considered equivalent to an activation function, for instance, $C^{\text{attn}}=\sum_{h \in H}f^{\text{attn}}_{W_{QK}}(X)\cdot W_{OV}$,  
$f^{\text{attn}}_{W_{QK}}(X)$ represents the \text{softmax}-normalization of the input $X$ through a weighted accumulation performed by $QK$ values, i.e., $f^{\text{attn}}_{W_{QK}}(X)=\text{softmax}((X  W_{Q}) (X W_{K})^{T}) X$. 

The function $f(X)$ possesses the ability for non-linear transformations, while $W$ is an input-agnostic parameter, which can be understood as a memory learned during training~\citep{geva2021transformer}. Therefore, these three components represent the minimum and complete unit for manipulating how much memory to read (i.e., memory-reading function), and are independent of each other, which we refer to as \textbf{``functional component''} (We elaborate the functional component in detail in Appendix~\ref{suppaamc}.)

\subsection{Implementation of Computation Graph}\label{secimplecomputation}
In this paper, we select GPT2-small as the target language model, containing 12 layers ($L=12$) and 12 attention heads ($H=12$). To provide a complete dissection of the the model at each layer which can precisely recover the original output, we introduce \emph{bias components} apart from \emph{functional component} and \emph{compensation component}, to compensate for the remaining information not covered.
Table~\ref{tabcircuitindex} shows the specific components and their implementation for each layer. Notably, our component dissection leads to a lossless decomposition of the original LM layer into fully linear combinations: $LM_l(X) = \sum_{i=0}^{28} C^i$.

We treat functional components as the smallest units and build the computation graph $\mathcal{G}=\{\mathcal{C},\mathcal{E}\}$, where $\mathcal{C}$ stands for the set of 29 components ($C^{0-28}$ shown in Table~\ref{tabcircuitindex}, where Attention and Attention+\text{MLP} has 12 components due to the 12 heads given) and $\mathcal{E}$ represents the edge between any two components in successive layers. Any functional component $C^{i} (0\leqslant i\leqslant 25)$ in any layer $l (0\leqslant l \leqslant 11)$, denoted as $C^{l,i}$, receive information from all components in previous layers, i.e., $\mathcal{E}=\{(C^{l_1,i} \rightarrow C^{l_2,j})\} (0\leqslant l_1 < l_2 \leqslant 11, 0\leqslant i, j \leqslant 25)$. Notably, lossless decomposition ensures that the insights gained from our linear decomposition computation graph accurately reflect the behavior of the original language model.

\subsection{Implementation of Pruning}\label{secgreedysearch}
\begin{table*}[t]
\begin{center}
\resizebox{0.9\linewidth}{!}{
\begin{tabular}{lcl}
\textbf{Index}&\textbf{Category}&\textbf{Implementation ($X$=input representation in each layer)}\\
      \hline
      $C^{0}$&Self&$X$\\
      $C^{1-12}$&Attention&$A^{h} \text{LN}(X) W_{V} W_{O}+ A^{h}  b_{V} W_{O}$\\
      $C^{13}$&\text{MLP}&$\text{atv}(\text{LN}(X) W_{M1}) W_{M2}$\\
      $C^{14-25}$&Attention+\text{MLP}& $\text{atv}(\text{LN}(A^{h} \text{LN}(X) W_{V} W_{O}+ A^{h}  b_{V} W_{O}) W_{M1}) W_{M2}$\\
      $C^{26}$&Compensation&$(\text{atv}(\text{LN}((\sum_{h=1}^{12}C^{h}) W_{M1}))-\sum_{h=1}^{12}\text{atv}(\text{LN}(C^{h}) W_{M1})) W_{M2}$\\
      $C^{27}$& Compensation&$(\text{atv}((\text{LN}(C^{0-13})W_{M1})-\text{atv}(\text{LN}(C^{0})W_{M1})-\text{atv}(\text{LN}(\sum_{h=1}^{12}C^{h})W_{M1})) W_{M2}$\\
      $C^{28}$&Bias&$b_{v}+\text{atv}(b_{M1}) W_{M2}+b_{M2}+\sum_{h=1}^{12}\text{atv}(b_{V} W_{M1}) W_{M2}$\\
\end{tabular}}
\end{center}
\caption{Specific component index and corresponding implementation in each layer of GPT2-small. $W$ and $b$ represent weight and bias parameters, $\text{atv}$ represents the activation of \text{MLP}. $\text{LN}(\cdot)$ is the layernorm function. $A=\text{softmax}(X W_{Q} W_{K}^{T} X^{T}+ b_{Q} W_{K}^{T}  X^{T}+X W_{Q} b_{K}^{T}+b_{Q} b_{K}^{T})$. Functional Component are $C^{1-25}$.}
\label{tabcircuitindex}
\end{table*}
For the computational graph $\mathcal{G}$ obtained in Section~\ref{secimplecomputation}, we default to using the Automated Circuit Discovery~\citep{conmy2023towards} toolkit to construct a circuit graph $\mathcal{G}*$. In simple terms, we eliminate those components that cause significant changes to original output in KL divergence through patching with interchange ablation in the computation graph and finally get the subgraph to completely reflect the task mechanisms as circuit graphs.  To support subsequent post-pruning causal mediation, instead of deriving a single circuit graph from the average of a dataset, we obtain a set of circuit graphs by computing the average of every 10 samples within the dataset. 

\subsection{Post-pruning Causal Mediation}\label{seccausaleffect}
In this section, we employ the post-pruning causal mediation to eliminate the effects of potential skills and other noise effects contained in real-world samples. 
Motivated by efforts in causal effect analysis~\citep{wang2023interpretability,vig2020investigating}, we divide the circuit graphs set (obtained from section~\ref{secgreedysearch}) into 3 sets of paths (sub-circuits): the paths of \textbf{skill effects}, the paths of \textbf{background effects}, and the paths of \textbf{self effects for the last token} (abbreviated as \textbf{self effects}). 
\textbf{Skill effects} refer to the effect of the target skill on the output, which is the focus of this paper. \textbf{Self effects} denote the impact of only using the last token to predict by memorization, which functions like a ``bi-gram model'' (e.g., inputting ``Francis'' and outputting ``Bacon''). \textbf{Background effects} refer to the effects of other potential skills in the input text (we illustrate these three effects through example texts in Appendix~\ref{suppexampleeffect}.).

We use the typical example of the ``Induction'' skill for illustration, which works with an input in the form of ``\textit{...A B...A}'', where \textit{A} and \textit{B} refers to different tokens. Here, the language model is expected to repeat the pattern ``\textit{A B}'' it has seen in the context and predict token ``\textit{B}'' as the output token.  
\begin{figure}
  \begin{center}
    \includegraphics[width=0.6\linewidth]{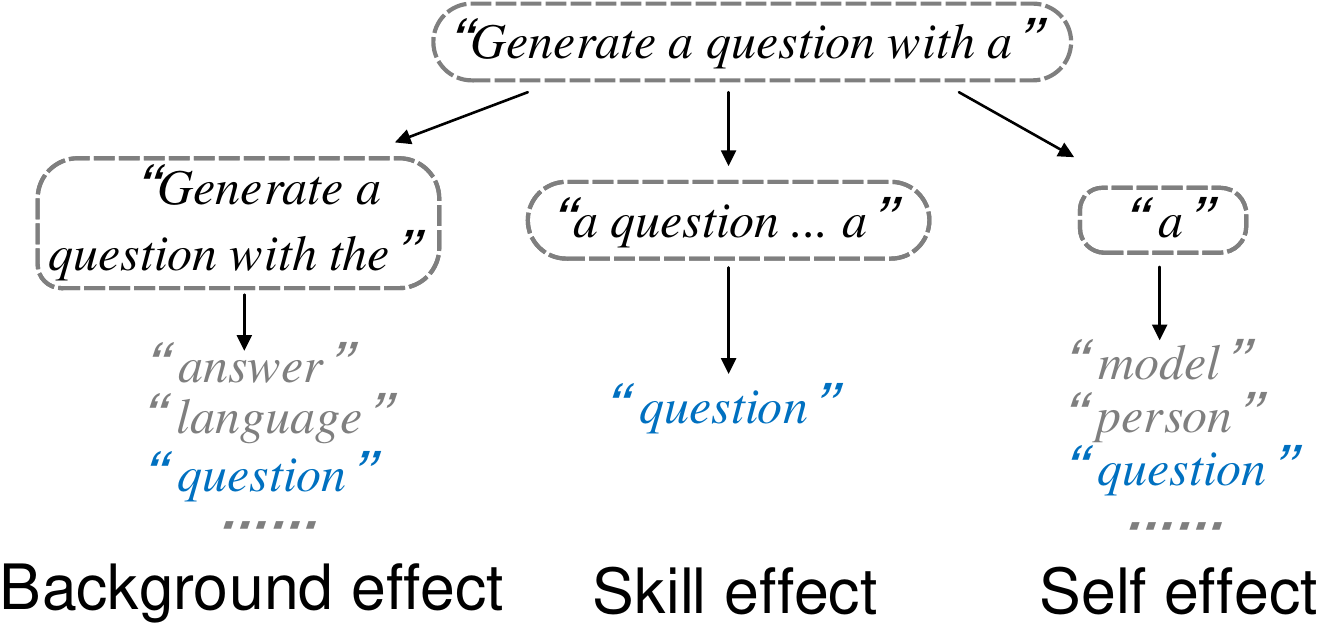}
  \end{center}
  \caption{A case text about causal effects.}
  \label{figcausaleffect}
\end{figure}

Figure~\ref{figcausaleffect} illustrates that the model outputs ``\textit{question}'' when given the input ``\textit{Generate a question with a}''. However, the vocabulary distribution in the output given by the language model does not merely result from the induction skill, but is also confounded by other effects such as the background effect and the self effect. To compute the target effect for a specific skill path, let $\text{Path}^{i}$ be any directed paths in $\mathcal{G}*$ (e.g., $C^{1, 19} \rightarrow C^{2, 14} \rightarrow C^{6, 5}$ $s.t.$ circuit edges $(C^{1, 19}, C^{2, 14})$ and $(C^{2, 14}, C^{6, 5})$ are in $\mathcal{G}*$). $\text{Path}^{i}$ then symbolizes the flow of information across layers amongst the components it encompasses. We use the occurrence rate of $\text{Path}^{i}$ in all samples to compute the effect:
\begin{equation}
\text{Eff}(\text{Path}_{\mathcal{G}*}^{i})=\frac{N_{\text{Path}_{\mathcal{G}*}^{i}=1}}{N_{all}}
\end{equation}
$N_{\text{Path}_{\mathcal{G}*}^{i}=1}$ represents the number of samples encompassing $\text{Path}^{i}$ while $N_{all}$ represents the number of all samples. Each path contributes differently to the three effects. Hence, our aim is to find those paths that contribute to the skill effect rather than the other two effects.

Specifically, for each input text as a sample $s$, we perturb it to create a background text $s_{\text{Bkg}}$ and a self text $s_{\text{Self}}$ (i.e., background and self text are two types of corrupted texts, and the process for generating these two text for all types of skills is described in Appendix \ref{suppdataimple}). Eventually, any sample is augmented with two more perturbed versions, rendering three types of input (i.e., original text, background text, and self text), each of which is subjected to the pruning as discussed in Section~\ref{secgreedysearch}. The pruning produces three distinct circuit graphs: $\mathcal{G}_{\text{Ori}}*$ (from the original input text), $\mathcal{G}_{\text{Bkg}}*$ (from the background text) and $\mathcal{G}_{\text{Self}}*$ (from the self text). Therefore, the skill effect (e.g., \emph{Induction Skill}) of $\text{Path}^{i}$ can be defined as: 
\begin{equation}
\begin{aligned}
    &\text{Eff}_\text{Skill}(\text{Path}^{i})=
    \frac{N_{\text{Path}_{\mathcal{G}_{\text{Ori}}*}^{i}=1, \text{Path}_{\mathcal{G}_{\text{Bkg}}*}^{i}=0, \text{Path}_{\mathcal{G}_{\text{Self}}*}^{i}=0}}{N_{all}}
\end{aligned}
\end{equation}
 Finally, we get all skill paths and compose them into a Skill Graph $\mathcal{G}^{S}=\{\mathcal{C}, \mathcal{E}^{S}\}$. With $\delta$ as the threshold parameter: $\mathcal{E}^{S}=\{\text{Path}^{i} | \text{Eff}_\text{Skill}(\text{Path}^{i})> \delta\}$ (we provide a detailed analysis of $\delta$ in Appendix~\ref{suppthreshold}).

\section{Experimental Design}
This paper focuses on 3 language skills, spanning from basic to advanced levels:
\vspace{-\topsep}
\begin{itemize}
\setlength{\parskip}{0pt}
\setlength{\itemsep}{0pt}
    \item \textbf{Previous Token Skill}: Receive information from the previous token.
    \item \textbf{Induction Skill}: Identify patterns in prefix matching and replicate recurring token sequences.
    \item \textbf{ICL Skill}: Recognize and replicate the demonstration context, thereby producing outputs based on similar patterns.
\end{itemize}

Extensive research has shown that these three skills build on one another in a sequentially encompassing manner~\citep{crosbie2024induction,olsson2022context,ren2024identifying,edelman2024evolution}. The Induction Skill inherently includes the Previous Token Skill. In simple terms, for induction to occur in the sequence ``\textit{A B ... A}'', the token \textit{B} must retrieve information from the preceding token \textit{A}. Likewise, In-Context Learning must be capable of identifying similar patterns across different demonstrations to generate analogous outputs. 

We select over 10k samples encompassing one of the three above-mentioned skills from large corpora and popular datasets such as WIKIQA~\citep{yang-etal-2015-wikiqa}, SST-2~\citep{socher2013recursive}, BIG-BENCH~\citep{srivastava2023beyond}, OpenOrca~\citep{OpenOrca}, and OpenHermes~\citep{OpenHermes}. For each instance, we create a background perturbation and a self perturbation (discussed in Section \ref{seccausaleffect}). 
For simplicity, \textbf{PVT} represents the sample set involving the Previous Token Skill and \textbf{IDT} represents the sample set related to Induction Skill. \textbf{ICL1} represents the ICL sample set from SST-2 datasets; \textbf{ICL2} represents the ICL sample set from object\_counting task; \textbf{ICL3} and \textbf{ICL4} represents those from qawikidata and reasoning\_about\_colored\_objects task. The details of data preparation and implementation are elaborated in Appendix~\ref{suppdataimple}.

\section{Validation}\label{secjustification} 
Since skill paths do not encompass all the skill mechanisms within a circuit graph, they function more as a sub-circuit and are unable to preserve the faithfulness of the original output. In other words, a complete circuit should include three: the background, the skill, and the self effects. However, the skill path only encompasses the skill effect, and therefore, it is unable to fully recover the output. This makes existing circuit justification methods (such as the KL divergence and task accuracy between the skill path and the original output of the model forward) \textbf{ unfeasible}. Fortunately, the linearly decoupled computational graph can recover the model's output losslessly with complete theoretical support, and the adopted pruning strategies have also been thoroughly validated in existing research. Therefore, \textbf{this paper focuses on verifying whether the skill paths can effectively reflect the mechanism of the target skill rather than the task dataset}. Specifically, we conducted the following three \textbf{Validation} experiments. 

\subsection{Path Ablation} 
To understand whether the skill paths are responsible for the corresponding language skills, we design an intervention experiment by removing different sets of paths and observe the output of the LM. 
\begin{table*}
\begin{center}
\resizebox{0.85\linewidth}{!}{
\begin{tabular}{llllllllll}
\textbf{Sample}&\multicolumn{9}{c}{\textbf{Circuit Graph}}\\
&$\mathcal{G}*$&$-R50$&$-R500$&$-\mathcal{G}^{S,\text{PVT}}$&$-\mathcal{G}^{S,\text{IDT}}$&$-\mathcal{G}^{S,\text{ICL1}}$&$-\mathcal{G}^{S,\text{ICL2}}$&$-\mathcal{G}^{S,\text{ICL3}}$&$-\mathcal{G}^{S,\text{ICL4}}$\\
\hline
PVT&1.00&0.46&0.23&0.01&0.00&0.00&0.01&0.00&0.00\\
IDT&1.00&0.58&0.29&0.08&0.00&0.00&0.00&0.01&0.00\\
ICL1&1.00&0.61&0.23&0.01&0.00&0.00&0.00&0.00&0.00\\
ICL2&1.00&0.51&0.18&0.00&0.00&0.01&0.00&0.01&0.01\\
ICL3&1.00&0.54&0.21&0.00&0.00&0.00&0.00&0.00&0.00\\
ICL4&1.00&0.62&0.30&0.07&0.03&0.01&0.02&0.00&0.00\\
\end{tabular}}
\end{center}
\caption{Accuracy of output to original label within different path removing}
\label{tabremoveedge}
\end{table*}
\begin{figure*}
  \centering
  \subfigure[PVT]{
    \includegraphics[width=0.15\linewidth]{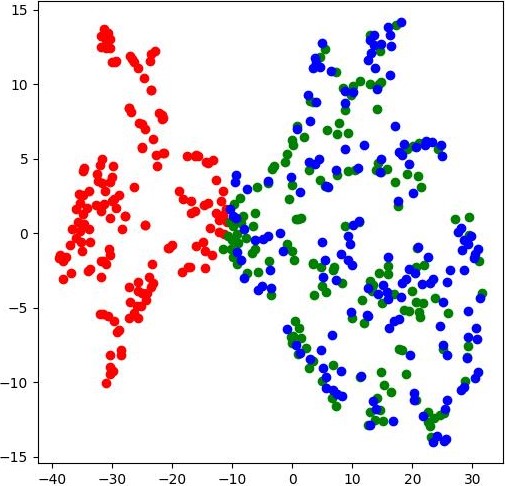}}
  \subfigure[IDT]{
    \includegraphics[width=0.15\linewidth]{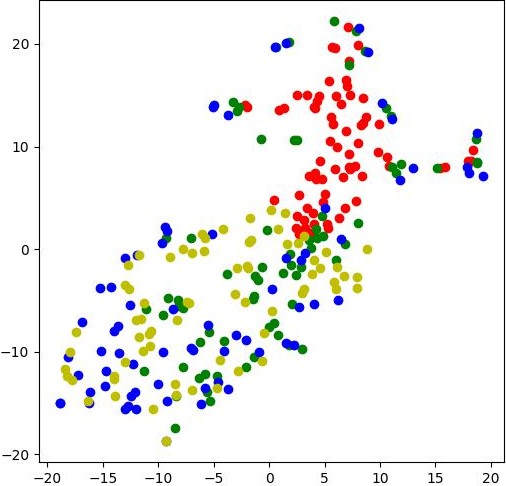}}
  \subfigure[ICL1]{
    \includegraphics[width=0.15\linewidth]{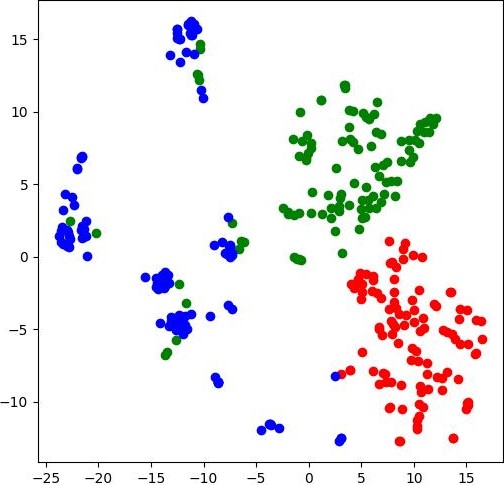}}
  \subfigure[ICL2]{
    \includegraphics[width=0.15\linewidth]{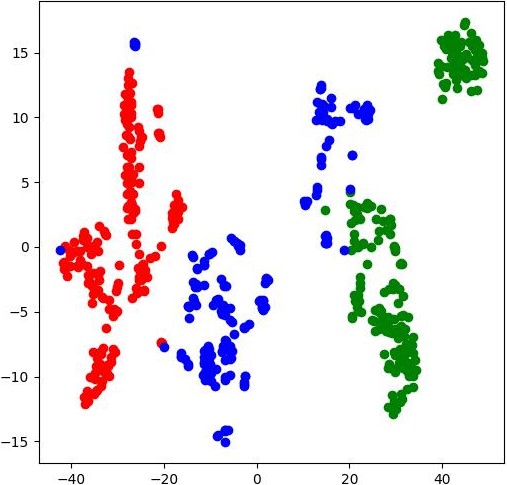}}
  \subfigure[ICL3]{
    \includegraphics[width=0.15\linewidth]{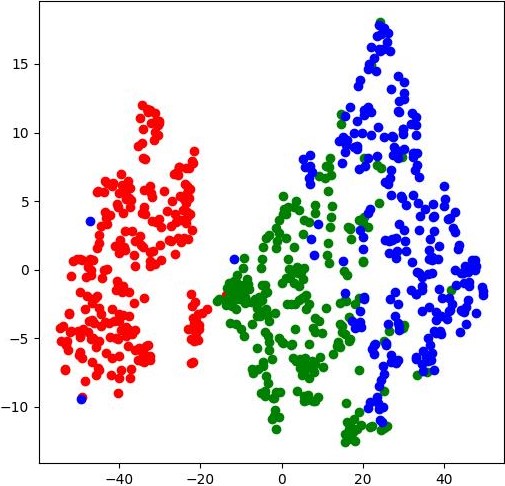}}
  \subfigure[ICL4]{
    \includegraphics[width=0.15\linewidth]{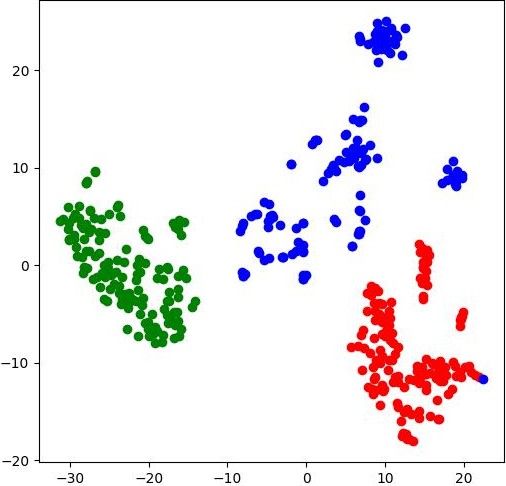}}
  \caption{T-sne visualization of 6 types of samples on vocabulary candidates. \textbf{Red} denotes the original output model ($\mathcal{G}$), while \textbf{blue} signifies the output once a corresponding skill path is removed ($\mathcal{G}-\mathcal{G}^{S}$). The outputs for the background text ($\mathcal{G}_{\text{Bkg}}$) and self text ($\mathcal{G}_{\text{Self}}$) are indicated in \textbf{green} and \textbf{yellow}, respectively.}
  \label{figtsne}
\end{figure*}

Table~\ref{tabremoveedge} displays the accuracy under different configurations of the Circuit Graphs when treating the original output as ground truth. For each language skill $S$, we randomly select 500 samples from its corresponding dataset. As a result, 9 different configurations of Circuit Graphs are tested: $\mathcal{G}*$ which represents the original output; $-R50$ which signifies the removal of 50 paths at random from $\mathcal{G}*$; $-R500$ after the deletion of 500 paths randomly from $\mathcal{G}*$, which approximately equals the number of skill paths\footnote{The exact number of removed paths is: $\mathcal{G}^{S,\text{PVT}}-325$, $\mathcal{G}^{S,\text{IDT}}-466$, $\mathcal{G}^{S,\text{ICL1}}-589$, $\mathcal{G}^{S,\text{ICL2}}-622$, $\mathcal{G}^{S,\text{ICL3}}-603$, $\mathcal{G}^{S,\text{ICL4}}-537$.}. The remaining 6 configurations encompass the removal of paths from $\mathcal{G}*$ that correspond to the skill of Previous Token, Induction, ICL1, ICL2, ICL3, and ICL4, respectively (For additional data for this validation test, please refer to Appendix~\ref{suppdataforvalidation}.).

The results indicate that almost all samples were unable to produce the original token when these skill paths were excluded (as indicated in the last 6 columns), yet random removal of paths does not lead to such significant impact.  
Additionally, Figure~\ref{figtsne} visualizes the t-SNE representation of the outputs associated with different Circuit Graphs. 
It is clear that when a skill path is removed, the output (blue) shifts from red toward green (or yellow), indicating a transition from a text output distribution that includes skills to a distinct space resulted from the removal of these skills. 

\subsection{Validation with Benchmark Task}
 In Appendix~\ref{suppioitask}, we performed validation on samples from the benchmark task, IOI. Using our method, we discovered the paths of some confirmed skills in the IOI circuit graph, including duplication skill, previous token skill, induction skill, inhibition skill, and name mover skill. We verified that the inclusion relationships of these skills are consistent with the edges in the original circuit graph. For instance, the paths of the induction skill include those of the previous token skill, and the paths of the inhibition skill include those of the induction skill, etc.

 \subsection{Validation for Effects of Other skills}\label{secvisualeffect}
 In Appendix~\ref{suppObgdf}, we visualized the probability density functions of the paths corresponding to the three effects (skill effects, background effects, and self effects). The experimental results confirmed the existence of high-frequency paths of background effect and self effect in the original circuit graph, which act as confounders affecting the extraction of skill paths. It also demonstrated that counterfactual and intervention corrupted texts can effectively reduce the influence of background effect and self effect.

\section{Discovery of Language Skills}\label{secdiscovery}
Two conjectures about language skills have been increasingly recognized. They are:
\vspace{-\topsep}
\begin{enumerate}
\setlength{\parskip}{0pt}
\setlength{\itemsep}{0pt}
\item \textbf{Stratification}: Simpler language skills reside in shallower
layers, whereas more complex language skills are found in deeper layers. 
\item \textbf{Inclusiveness}: More complex language skills are formed on top of simpler skills. 
\end{enumerate}

For \textbf{Stratification} and \textbf{Inclusiveness}, many works have already discovered their traces. For instance, \citet{wang2023interpretability} uses causal tracing to show that the previous token head constitutes a subcomponent of the induction circuit in the shallow layers, while \citet{crosbie2024induction,olsson2022context} demonstrates that the ICL capability emerges as a higher-order skill built upon induction. Taken together, these studies support the notion that these three skills form a progression of increasing complexity. 

However, so far, there has been no \textbf{quantitative} evidence for these 2 conjectures. E.g., in which specific layers do simple skills reside? What specific paths in simple skills are encompassed by complex skills? Our work further confirms these via quantitative discoveries.

\subsection{Quantitative Results}\label{sec:language skill}
\begin{table*}
\begin{center}
\resizebox{0.9\textwidth}{!}{
\begin{tabular}{ll}
\textbf{Skill}&\textbf{Receivers receiving more than 10 paths ([\#layer, \#components])}\\
\hline
\textbf{PVT}&[1, 8], [1, 18], [1, 19], [1, 20], [1, 21], [2, 1], [2, 7], [2, 14], [2, 18], [2, 20], [2, 22], [2, 24], [11, 1], [11, 14]\\
\hline
\textbf{IDT}&\textbf{[2, 14]}, \textbf{[2, 18]}, \textbf{[2, 20]}, [3, 14], [3, 17] [4, 5], [4, 12], [5, 11], [6, 5], \textbf{[11, 1]}, \textbf{[11, 14]}\\
\hline
\multirow{2}{*}{\textbf{ICL1}} &\textbf{[2, 14]}, \textbf{[2, 20]}, \textbf{[2, 22]}, \textbf{[2, 24]}, [3, 3], [3, 4], [3, 5], [3, 11], \textbf{[3, 14]}, \textbf{[3, 17]}, [4, 3], \textbf{[4, 5]}, \textbf{[5, 11]}, [8, 5],\\ 
&[10, 10], [11, 8], [11, 9], [11, 10], [11, 11]\\
\hline 
\multirow{2}{*}{\textbf{ICL2}} &\textbf{[1, 19]}, \textbf{[2, 14]}, \textbf{[2, 20]}, \textbf{[2, 24]}, [3, 5], [3, 11], \textbf{[3, 14]}, \textbf{[4, 5]}, [4, 7], [4, 9], [5, 10], \textbf{[6, 5]},[10, 9], [10, 10],\\
&[10, 11], \textbf{[11, 1]}, [11, 5]\\
\hline
\multirow{2}{*}{\textbf{ICL3}} & \textbf{[1, 8]},  \textbf{[1, 18]},  \textbf{[1, 19]},  \textbf{[1, 20]},  \textbf{[1, 21]},  \textbf{[2, 14]},  \textbf{[2, 20]},  \textbf{[2, 24]}, [3,1],  \textbf{[3, 14]}, [4, 3],  \textbf{[4, 5]}, [5, 1], [5, 10],\\&
\textbf{[5, 11]}, [8, 1], [8, 9], [10, 5], [10, 10], [10, 12], \textbf{[11, 1]}, [11, 8]\\
\hline
\multirow{2}{*}{\textbf{ICL4}} &[1, 16], \textbf{[1, 20]}, \textbf{[2, 20]}, [4, 3], \textbf{[4, 5]}, [5, 3], [6, 4], \textbf{[6, 5]}, [8, 9], [9, 4], [9, 5], [10, 2], [10, 10], [10, 12],\\
&[11, 2], [11, 3], [11, 4], [11, 6], [11, 15]\\
\hline

\end{tabular}}
\end{center}
\caption{Key Receivers in skill paths, \textbf{bold} components are presented in the lower skill.}
\label{tabreceivers}
\end{table*}
Table~\ref{tabreceivers} displays the nodes (receivers) receiving more than 10 paths in the skill graphs. We use $[l, i]$ to denote the $l$-th layer and $i$-th components ($C^{l,i}$). The complete Skill Graph can be found in Appendix~\ref{suppskillgraph}. From Table~\ref{tabreceivers}, we provide quantification results for Stratification and Inclusiveness.

\textbf{Quantification of Stratification}: The Previous Token Skill (PVT) is one of the simplest language skills and is located across layers 0-2. The Induction Skill (IDT) is slightly more complex, and thus spreads across layers 0-6. Meanwhile, ICL is the most complex skill and has key receivers across nearly all layers. 

\textbf{Quantification of Inclusiveness}: Components such as [2, 14], [2, 18], and [2, 20] (presented in PVT) can be found in the IDT, indicating that the PVT is an integral part of the IDT. Similarly, the ICL skill encapsulates the PVT and IDT as necessary sub-skills, which is why components that are evident in the PVT (such as [2, 14], [2, 20], [2, 24]) and those identified in the IDT (such as [3, 14], [4, 5]) can be found in the ICL Skill Graph. Furthermore, we list all multi-step paths with inclusive sub-path in Appendix~\ref{suppmultistep}. 

\subsection{Comparison among Pruning Strategies}
\label{suppcomvc} 
\begin{figure*}
  \centering
  \subfigure[E-pruning]{
    \includegraphics[width=0.15\linewidth]{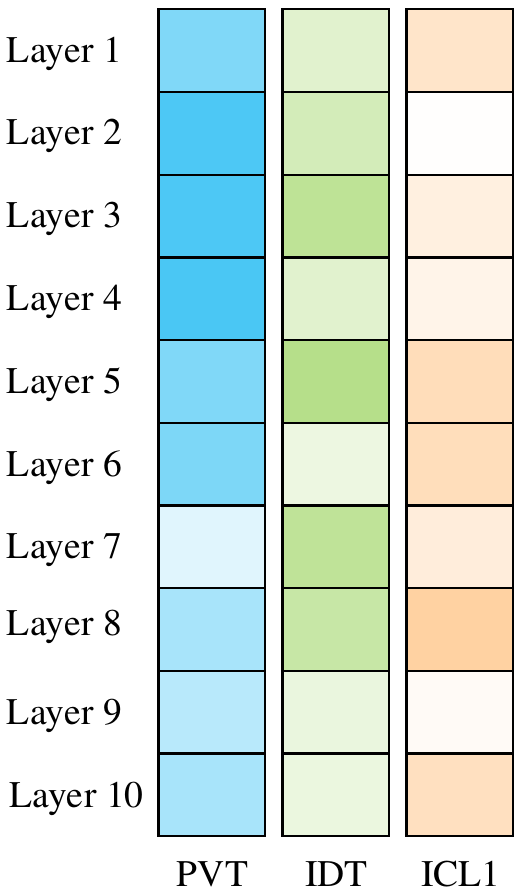}}
  \subfigure[EAP]{
    \includegraphics[width=0.15\linewidth]{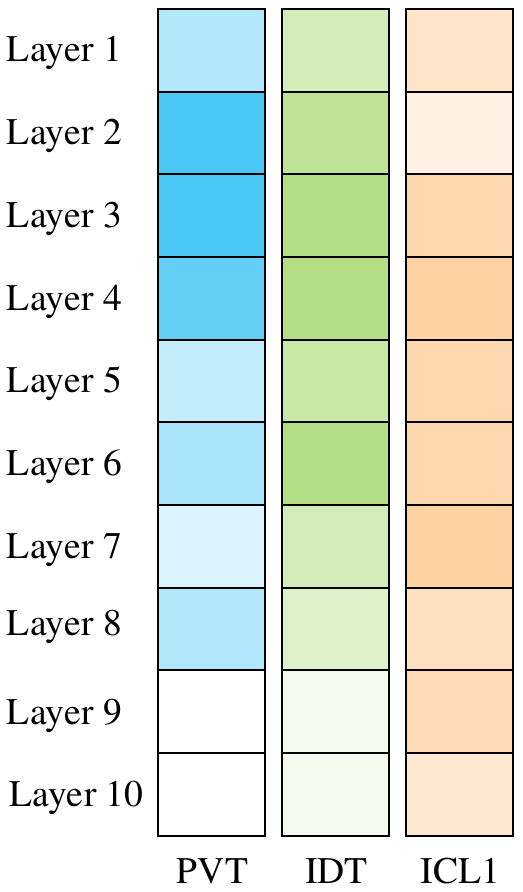}}
  \subfigure[DiscoGP]{
    \includegraphics[width=0.15\linewidth]{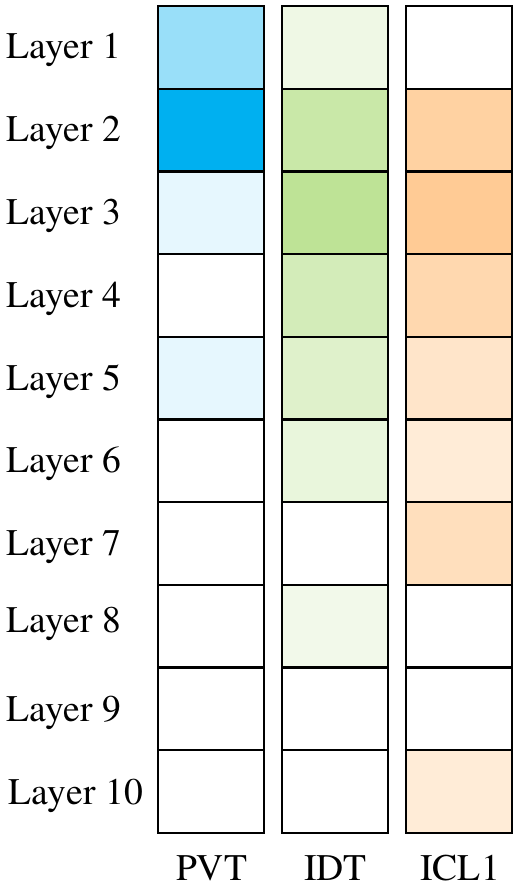}}
  \subfigure[Scrubbing]{
    \includegraphics[width=0.15\linewidth]{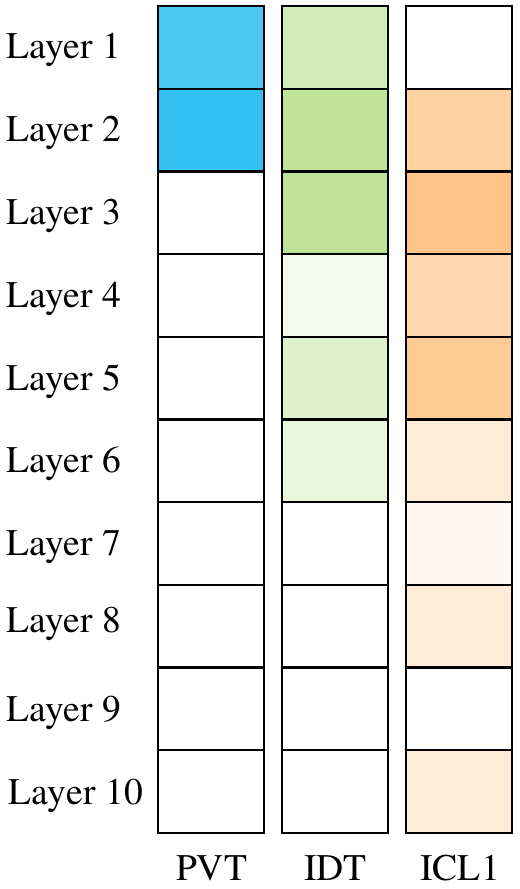}}
  \subfigure[ACDC]{
    \includegraphics[width=0.15\linewidth]{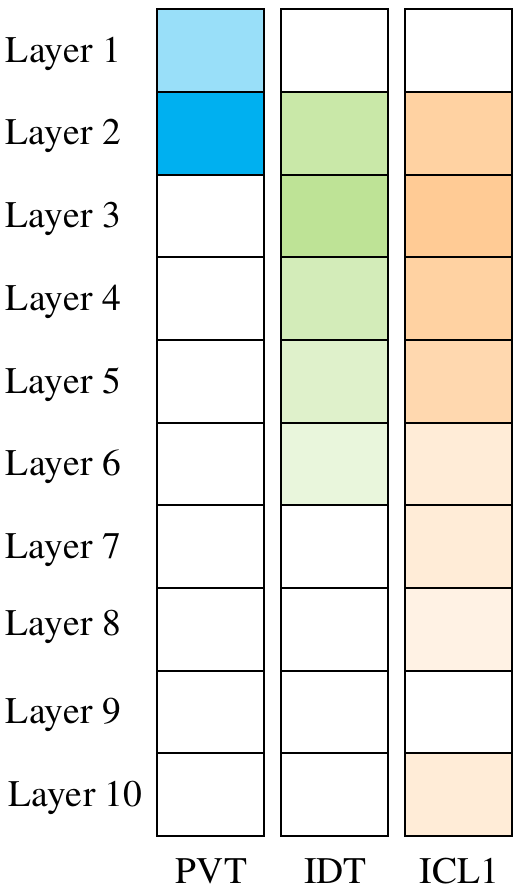}}
  \caption{Visualization of receivers distributed in layer1-10 in 3 increasingly-complex skills (PVT, IDT, and ICL1).}
  \label{figreceivers}
\end{figure*} 
\begin{table*}
\begin{center}
\resizebox{1\textwidth}{!}{
\begin{tabular}{llllllll}

\textbf{Types}&\textbf{Method}&$\text{ovlp(IDT, PVT)}$&$\text{ovlp(ICL1, PVT)}$&$\text{ovlp(ICL1, IDT)}$&$\text{ovlp(GT,PVT)}$&$\text{ovlp(GT,IDT)}$&$\text{ovlp(GT,ICL1)}$\\
\hline
\multirow{5}{*}{Circuit}&ACDC&0.19&0.06&0.17&0.11&0.05&0.11\\
&E-pruning&0.05&0.14&0.17&0.18&0.10&0.05\\
&EAP&0.14&0.05&0.18&0.09&0.15&0.12\\
&DiscoGP&0.18&0.07&0.15&0.08&0.11&0.09\\
&Scrubbing&0.10&0.08&0.17&0.16&0.11&0.10\\
\hline
\multirow{5}{*}{Paths}&ACDC+Ours&0.74&0.81&0.63&0.68&0.13&0.11\\
&E-pruning+Ours&0.67&0.78&0.59&0.67&0.09&0.14\\
&EAP+Ours&0.70&0.68&0.67&0.74&0.17&0.05\\
&DiscoGP+Ours&0.79&0.81&0.71&0.77&0.04&0.13\\
&Scrubbing+Ours&0.74&0.77&0.68&0.71&0.08&0.13\\
\end{tabular}}
\end{center}
\caption{Overlaps between different circuits and between different paths}
\label{taboverlap}
\end{table*}
We investigate the performance of our framework replacing different pruning strategies in validating the 2 conjectures. 
We investigate existing pruning strategies (ACDC~\citep{conmy2023towards}, E-pruning~\citep{bhaskar2024finding}, EAP~\citep{syed2023attribution}, DiscoGP~\citep{yu2024functional}, Scrubbing~\citep{chan2022causal}), which are prevalent methods for circuit pruning, as detailed in Appendix~\ref{suppdetailsbaseline}. 
Specifically, we replace our original pruning process with each pruning strategy mentioned above and test on the following three skills: PVT, IDT, and ICL1. Then, we investigate the distribution of the receiver nodes among layers in these circuit graphs and display the normalized results in Figure~\ref{figreceivers}.

It is clear that in general PVT is more prominent in the shallow layers, while IDT in the shallow to mid layers. ICL1 tends to cluster in the deep layers. Yet, the paths discovered from the circuits via pruning strategies - DiscoGP, Scrubbing, and ACDC - provide distinct patterns from those using E-pruning and EAP. From (c), (d) and (e), PVT circuits appear almost exclusively in layers 1 and 2, and IDT circuits appear only in layers 1-6. The differences in these pruning methods indicate that circuits from greedy search (ACDC), continuous optimization (DiscoGP), and path patching (Scrubbing) are more conducive to reflecting the stratification in skill paths.

\subsection{Skill Path vs. Circuit Graph}
To further interpret how different methods capture \textbf{Inclusiveness} and compare between existing circuit graphs and our skill paths, we investigate the overlap of circuits identified for the 4 skills: PVT, IDT, ICL1, and greater than (GT) in Table~\ref{taboverlap}. GT, with input samples drawn from the \textit{greater than} dataset~\citep{hanna2024does}, assesses the ability of the language model to judge numerical relations. An example from the GT dataset is ``The war lasted from 1517 to 15?''. To demonstrate the difference between skill paths and circuit graphs, we use the circuit graphs obtained by these methods themselves (rows 2-6) and the skill paths obtained using their pruning strategies in combination with our framework (rows 7-11), to calculate the overlap between skills. Here we use $ovlp(A, B)$ (detailed in Appendix~\ref{suppovlp}) to indicate the ratio of edges in circuit/paths $A$ that also exist in $B$. A value of $0$ means that no edges are shared, and $1$ means all edges are shared. 

As discussed in Section \ref{sec:language skill}, the three skills PVT, IDT, and ICL are progressively inclusive, that is, IDT includes PVT, while ICL includes both IDT and PVT. In addition, GT only includes PVT and has no relation to IDT and ICL. 
Table~\ref{taboverlap} shows that the overlap of circuit graphs discovered by existing methods is rather weak. For example, $ovlp(\text{ICL1}, \text{IDT})$ is only 0.17 in ACDC. In contrast, our approach provides clear empirical evidence for the conjecture of inclusiveness: for instance, $ovlp(\text{IDT}, \text{PVT})=0.74$ indicates that 74\% of the edges in the Induction skill exist in the previous token skill. Additionally, from the experimental results related to GT, it can be seen that for skills without inclusiveness, our skill paths can also reflect significant differences. 

In addition to the main experiments, we conducted \textbf{two comparison studies} to further understand the skill paths: one, detailed in Appendix~\ref{suppthreshold}, compared the dynamic results on the number of paths, KL divergence, and task accuracy as the threshold $\delta$ varied from 0 to 1, revealing that the optimal trade-off among these variables occurs when $\delta =0.6-0.7$; the other, described in Appendix~\ref{suppbackground}, involved designing different corrupted texts to assess the background effect, with results indicating that the background effects were similar. 

Finally, we present an analysis of why the model fails to execute the skills in Appendix~\ref{suppexploration}. It demonstrates that through the stratification and inclusiveness of our skill paths, we can explain that the failure of an LLM to execute a high-level skill is due to the failure of a certain basic skill. The limitations are shown in Appendix~\ref{limitations}.


\newpage

\bibliography{iclr2025_conference}
\bibliographystyle{abbrvnat}

\newpage
\appendix
\section{Details about Non-Transitivity and Preeption}
\label{supppitfall}

Initially, we would like to recall the description of causal dependency verification using counterfactuals: ``\textit{If A occurs, then B occurs; and if A does not occur, then B does not occur}''~\citep{lewis2013counterfactuals}. Building upon this, there are two common pitfalls in existing counterfactual operations~\citep{mueller2024missedcausesambiguouseffects}: non-transitivity, which leads to the overlooking of local patterns, and preemption, where redundant causes are disregarded. Since instantiating these concepts at the neural level would introduce significant complexity, we instead provide examples from the perspective of events to illustrate these pitfalls.

\textbf{Non-transitivity:} Consider the following sequence of events (adapted from~\citet{mueller2024missedcausesambiguouseffects}):

\textit{A: A hiker is walking up a mountain, and a large boulder begins rolling down the mountain toward the hiker.}

\textit{B: The hiker, noticing the rolling boulder, ducks out of the way.}

\textit{C: The hiker survives.}

In this sequence, C depends on B, and B depends on A. For instance, the dependency of C on B can be framed as: "If the hiker ducks, they survive; if the hiker does not duck, they do not survive." However, counterfactuals cannot directly verify that C depends on A. This is because A and C are embedded within a more complex causal structure where the mediation between A and C involves more than just B. The same reasoning holds at the neural level: counterfactual mediation can verify the existence of causal relationships such as $A\rightarrow B$ and $B\rightarrow C$, but it cannot directly verify $A\rightarrow C$. In this paper, we address this issue by utilizing path-level circuits: if we can simultaneously verify $A\rightarrow B$, $B\rightarrow C$, and $A\rightarrow B\rightarrow C$, then we can confirm $A\rightarrow C$.

\textbf{Preemption:} Consider the following example:

\textit{A1: Suzy throws a rock at a glass bottle.}

\textit{A2: Simultaneously, Billy throws a rock at the same bottle.}

\textit{B: The rocks shatter the bottle.}

In this case, verifying $A1\rightarrow B$ first would prevent the verification of $A2\rightarrow B$ (since the bottle is already shattered), and vice versa. Similarly, in neural networks, a given neuron may be influenced by multiple causes. Performing counterfactual mediation for a single cause may obscure the causal effects of other causes. To resolve this, we apply counterfactual mediation to all paths \textbf{simultaneously} in the third step of our framework, ensuring that this issue is avoided.

\section{Analysis about Memory Circuits}\label{suppaamc}
\subsection{Full Equations of All Memory Circuits} 

Three of memory circuits can be further factorized as: 
\begin{equation}
\begin{aligned}
     &C^{attn}=\sum_{h \in H}f^{attn}_{W_{QK}}(X)\cdot W_{OV}\\
     &where~f^{attn}_{W_{QK}}(X)=softmax((X  W_{Q}) (X W_{K})^{T}) X\\     
    &C^{mlp}=f^{mlp}_{W_{M1}}(X)\cdot W_{M2}\\
    &where~f^{mlp}_{W_{M1}}(X)=atv(X W_{M1})\\
     &C^{attn+mlp}=\sum_{h \in H}f^{attn+mlp}_{W_{QK},W_{OV},W_{M1}}(X)\cdot W_{M2} \\
     &where~f^{attn+mlp}_{W_{QK},W_{OV},W_{M1}}(X)=atv(f^{attn}_{W_{QK}}(X) W_{OV}  W_{M1}) \nonumber 
 \end{aligned}
\end{equation}

We use $f$ to represent a function that can be considered equivalent to an activation function, for instance, $f^{attn}_{W_{QK}}(X)$ represents the softmax-normalization of the input $X$ through a weighted accumulation performed by $QK$ values. In conclusion, these three types of circuits can be expressed using a common paradigm:
\begin{equation}
    C^{attn/mlp/attn+mlp} = f(X)\cdot W
    \label{eqtmemorycircuit}
\end{equation}

\subsection{Why $A \otimes X$ is not the circuit with complete function?}
We use $X^{l,n}$ to denote the hidden state representation corresponding to the $n$-th token at the $l$-th layer, and $U$ represents the unembedding matrix. Therefore, for any representation $X^{l,n}$, we can obtain its vocabulary distribution, i.e., the logits for each token candidate, using $X^{l,n}U$. We adopt a sample text, \textit{``Beats Music is owned by"}, as the input. Table~\ref{tablogitsembedding} shows the logits corresponding to the words \textit{`` the"} and \textit{`` Apple"} when these tokens are converted to vocabulary embeddings.

Our expected correct output is such that after the last layer's representation is unembedded, the logits for \textit{`` Apple"} reach their peak. However, as shown in Table ~\ref{tablogitsembedding}, after conducting an $A\otimes X$ operation on the 1st layer's representation, the logit range for \textit{`` Apple"} is $[80.49,86.44]$, where $80.49$ corresponds to the attention weight of \textit{`` Music''} to \textit{`` by"} being $1$, and $86.44$ represents the attention weight of \textit{`` Be''} to \textit{`` by"} being 1. 

This situation exposes a significant drawback. In the representations of all previous tokens, the logits for \textit{`` the"} are always higher than those for \textit{`` Apple"}. Hence, no matter how many effects $A\otimes X$ operations performed, it remains impossible for the logits of \textit{`` Apple"} to surpass those of \textit{`` the"}. Therefore, although $A\otimes X$ incorporates an activation function such as $softmax$, it can only be considered as semi-activated~\citep{elhage2021mathematical}. We refer to this as a ``deep constraint", that is, $A\otimes X$ cannot allow the representation of the destination token to exceed the upper and lower boundaries of the previous token’s representation. This is why we assert that $A\otimes X$ lacks full functions, that is, it does not possess memory capability. 

\begin{table*}[t]
\begin{center}
\resizebox{0.7\linewidth}{!}{
\begin{tabular}{lllllll}
\textbf{Logits}&\multicolumn{6}{c}{\textbf{Tokens}}\\
&\textit{``Be''}&\textit{``ats''}&\textit{`` Music''}&\textit{`` is''}&\textit{`` owned''}&\textit{`` by''}\\
      \hline
      \textit{`` the''}&95.45& 89.43&91.20&99.32&94.21&101.52\\
      \textit{`` Apple''}&86.44&82.13&80.49&82.31&82.57&83.41\\
    
\end{tabular}}
\end{center}
\caption{Logits of \textit{`` the''} and \textit{`` Apple''} when the representation in 1-st layer products unembedding matrix, with input \textit{``Beats Music is owned by"}}
\label{tablogitsembedding}
\end{table*}

\subsection{How to explain Memory Circuits?}
Let's likewise map all the Memory Circuits into the vocabulary space: 
\begin{equation}
    V=C\cdot U= f(X) \cdot W \cdot U=f(x)\cdot WU
    \label{eqtmemoryvocabulary}
\end{equation}
Simply put, we assume $X \in \mathbb{R}^{N, D}$, $f(X) \in \mathbb{R}^{N, M}$, $W \in \mathbb{R}^{M, D}$, and $U \in \mathbb{R}^{D, E}$, where $N$ represents the number of tokens, $D$ denotes the dimensions in the residual stream, $M$ refers to the dimensions in the circuit (such as the dimensions in QKV or MLP), and $E$ signifies the length of the vocabulary list. Naturally, $WU \in \mathbb{R}^{M, E}$, which could be seen as a collection of $M$ vocabulary distributions. These vocabulary distributions are unaffected by the input tokens and thus can be considered as the acquired memory from training. 

The function $f(X) \in \mathbb{R}^{N, M}$ acts like a weight which specifies how much each vocabulary distribution contributes to the output. This confirms why MLP is generally regarded as a memory storage, as its dimensions are usually significantly larger than those of QKV. Simultaneously, it also explains the advantage of MoE: providing a wider range of options for vocabulary distribution.

In the final analysis, the inference process of a language model can be seen as constituting 3 key components: \textbf{``memory''}, \textbf{``movement''}, and \textbf{``ensemble''}. \textbf{``Memory"} pertains to acquiring a new distribution from memory distribution, while \textbf{``movement''} involves transferring token information to subsequent tokens. Finally, \textbf{``ensemble''} refers to the process of combining representations from multiple circuits to produce the final representation.
Within this process, Memory Circuits serve as the smallest units responsible for \textbf{``memory''} and also encompass independent operations of \textbf{``movement''} ($C^{1-12}$ and $C^{14-25}$). Furthermore, they form individual elements of the \textbf{``ensemble''}. Therefore, we examine the interrelationships (necessary paths) between Memory Circuits to understand the language skills of language models.

\subsection{Derivation of Compensation Circuits}
\label{suppdcc}
The input of the MLP consists of two parts: the residual stream and the output of the attention. Due to the presence of nonlinear activation functions, the residual stream and attention are coupled in the input, making it impossible to isolate their impact on the MLP, thereby affecting the verification of pruning. To address this, we introduce a compensation circuit, decomposing the MLP into four parts:
\begin{equation}
    \begin{aligned}
        &atv((X+\sum_{h \in H}Attn^{h})W_{M1})W_{M2}=(atv(XW_{M1})+\sum_{h \in H}atv(Attn^{h}W_{M1]}))W_{M2}+Cps^{1}+Cps^{2}\\
        &where:~Cps^{1}=(atv((X+\sum_{h \in H}Attn^{h})W_{M1})-atv(XW_{M1})-atv(\sum_{h \in H}Attn^{h}W_{M1}))W_{M2}\\ 
        &Cps^{2}=(atv(\sum_{h \in H}Attn^{h}W_{M1})-\sum_{h \in H}atv(Attn^{h}W_{M1}))W_{M2}
    \end{aligned}
    \label{eqtlineardecoposition}
\end{equation}
where MLP operation with activation given by $atv((X+\sum_{h \in H}Attn^{h}) W_{M1}) W_{M2}$ ($W_{M1}$ and $W_{M2}$ are weight parameters in two linear layers and $atv$ represents the activation function), $X$ represents the input representation in each layer and $H$ represents the number of attention heads, $Attn^{h}$ represents the output of $h$-th attention head, $Cps^{1}$ and $Cps^{2}$ are compensation circuit, representing the synergy effect of the residual stream ($X+\sum_{h \in H}Attn^{h}$) and the sum of attention head $\sum_{h \in H}Attn^{h}$ respectively. 

The compensation circuit calculates the synergy between the output when linear terms are summed before passing through a non-linear function, and the output passing through a non-linear function before summing. Therefore, the compensation circuit is dynamic and related to the input. From the perspective of the MLP, if we want the compensation circuit to be 0, then the input to the MLP must be reduced to only one or zero linear terms. This is an unlikely occurrence in practical pruning, so we assume that all edges of the compensation circuit always exist.

\section{Example Input-Output Text for Three Effects}
\label{suppexampleeffect}

The skill effect, background effect, and self effect are the three influences we have identified that affect the model's output of the next token, as well as the three sub-circuits of the complete circuit graph. These three effects typically co-exist, but we can identify instances where one effect is particularly prominent. For example, when we set the target skill as the induction skill, the following three situations can occur:

\textbf{Dominant skill effect}: input text: ``\textit{Generate a question with a }'', output: ``\textit{question}''.

\textbf{Dominant background effect}: input text: ``\textit{Today is a really good day to relax; emotion: happy. I got into a car accident when I went out; emotion: sad. Someone actually asked me to have dinner; emotion:}'', output: ``\textit{surprised}''.

\textbf{Dominant self effect}: input text: ``\textit{There is a Barack family, and among them there is a Barack.}'', output: ``\textit{Obama}''.

In the above examples, the input text for all three effects includes the induction skill (in the skill effect, it's ``\textit{...a question... a}'', in the background effect, it's ``\textit{...emotion: sad... emotion:}'', and in the self effect, it's ``\textit{...a Barack family ... a Barack}''). However, only the skill effect correctly outputs the token that the induction skill should output (``\textit{...a question... a question}''). In the background effect, the output of another skill (ICL skill) replaces the output of the induction skill, so we refer to this effect from outside the target skill as the background effect. In the self effect, the output rationale comes from a frequently occurring fixed phrase ``\textit{Barack Obama}'', in other words, the output seems to be triggered solely by the input's last token, so we refer to this output caused by the last token itself as the self effect.

\section{Data Preparation and Implementations}\label{suppdataimple}
\subsection{Data Preparation}
\subsubsection{Previous Token Skill}
We randomly selected 40k text samples comprising two tokens - \textit{``token0 token1"} - from the WIKIQA, OpenOrca, and OpenHermes corpora. In 20k of these samples, the two tokens made up one word, while in the remaining 20k, \textit{``token0''} and \textit{``token1''} belonged to two separate words. For the background text, we chose \textit{``token0''}, and for the self text, we selected \textit{`` token1''}. A complete sample is as follows: 

\textit{\{text: `` that most", backgound\_text: `` that", self\_text: `` most'', GPT2-small\_output: `` of''\}}

\subsubsection{Induction Skill}
The samples for the Induction Skill also come from WIKIQA, OpenOrca, and OpenHermes. We randomly selected 14k samples with the template \textit{``... A1 B ... A2"}, where the destination token \textit{`` A2''} is the same as the preceding token \textit{`` A1''}, and the total token length of the sample does not exceed $30$. For the background text, we removed \textit{`` A2''} and had GPT2-small produce a new but different token to replace \textit{`` A2''}, resulting in \textit{``... A1 B ... C"}. Since \textit{`` C''} is semantically supplemented by the preceding text and differs from \textit{`` A2''}, it preserves semantics as much as possible without the Induction Skill. The self text is still token \textit{`` A2''}. A complete sample is as follows:

\textit{\{text: ``chinese lesson 1.2: chinese", backgound\_text: ``chinese lesson 1.2: The", self\_text: `` chinese'', GPT2-small\_output: `` lesson''\}}

\subsubsection{ICL Skill}
The 4 types of ICL skill samples come from SST-2 dataset and the object\_counting, qawikidata, reasoning\_about\_colored\_objects datasets in BIGBENCH. These samples have been named by us as \textit{icl\_sst2}, \textit{icl\_oc}, \textit{icl\_qa}, \textit{icl\_raco}, with quantities of \textit{1000}, \textit{284}, \textit{1000}, and \textit{135} respectively. Each sample is required to contain two different labelled demonstrations and should be answerable correctly by GPT2-small. Here are examples of the four types of samples:

\textit{icl\_sst2}:

\textit{\{text: ``, nor why he keeps being cast in action films when none of them are ever any good Sentiment: negative$\backslash$nfunny , even punny 6 Sentiment: positive$\backslash$nis that secret ballot is a comedy , both gentle and biting . Sentiment:", backgound\_text: ``is that secret ballot is a comedy , both gentle and biting . Sentiment:", self\_text: `` Sentiment:'', GPT2-small\_output: `` positive''\}}

\textit{icl\_oc}:

\textit{\{text: ``I have a piano, a trombone, a violin, and a flute. How many musical instruments do I have?A: four$\backslash$nI have a banana, a plum, a strawberry, a nectarine, an apple, a raspberry, an orange, a peach, a grape, and a blackberry. How many fruits do I have?A: ten$\backslash$nI have a head of broccoli, a cauliflower, a stalk of celery, a cabbage, a potato, an onion, a yam, a garlic, a lettuce head, and a carrot. How many vegetables do I have?A:", backgound\_text: ``I have a head of broccoli, a cauliflower, a stalk of celery, a cabbage, a potato, an onion, a yam, a garlic, a lettuce head, and a carrot. How many vegetables do I have?A:", self\_text: `` A:'', GPT2-small\_output: `` ten''\}}

\textit{icl\_qa}:

\textit{\{text: ``The country of University of Tsukuba is A: Japan$\backslash$nThe sport played by Judit Polgár is A: chess$\backslash$nThe country of citizenship of Théophile Gautier is A:", backgound\_text: ``The country of citizenship of Théophile Gautier is A:", self\_text: `` A:'', GPT2-small\_output: `` France''\}}

\textit{icl\_raco}:

\textit{\{text: ``On the nightstand, you see the following objects arranged in a row: a black bracelet, a pink booklet, a blue cup, and a silver cat toy. What is the color of the object directly to the left of the pink object? A: black$\backslash$nOn the floor, you see a bunch of objects arranged in a row: a red cup, a gold bracelet, a fuchsia puzzle, a purple stress ball, and a burgundy fidget spinner. What is the color of the object directly to the right of the cup? A: gold$\backslash$nOn the table, you see a set of things arranged in a row: a black keychain, a purple mug, a blue dog leash, and a teal sheet of paper. What is the color of the left-most thing? A:", backgound\_text: ``On the table, you see a set of things arranged in a row: a black keychain, a purple mug, a blue dog leash, and a teal sheet of paper. What is the color of the left-most thing? A:", self\_text: `` A:'', GPT2-small\_output: `` black''\}}

\subsection{Implementation of Extracting Skill Paths}\label{suppimple}
Following the 3-step process from Section~\ref{secmethod}, we obtained the skill path $\mathcal{G}^{S}$. We found that the skill effect values in $\mathcal{G}^{S}$ for the Previous Token Skill and the Induction Skill were not high, with the highest $\text{Eff}_\text{Skill}$ being only 0.54 and 0.61, respectively. However, the highest $\text{Eff}_\text{Skill}$ for the ICL Skill reached 0.98. We speculated that because the Previous Token Skill and the Induction Skill are overly simple, there were a significant number of samples that happened to output the correct answers without triggering the corresponding skill paths. For instance, in the text \textit{``In China [mainland]"}, it is challenging to confidently determine whether \textit{``mainland"} was influenced by the bi-gram model of \textit{``China"} or if \textit{``China"} received information from \textit{``In"}. As such, we attempted to perform bisection clustering for each sample in the Previous Token Skill and Induction Skill, based on the paths with top 10\% $\text{Eff}_\text{Skill}$.

\begin{figure*}
  \centering
  \subfigure[Previous Token Skill]{
    \includegraphics[width=0.48\linewidth]{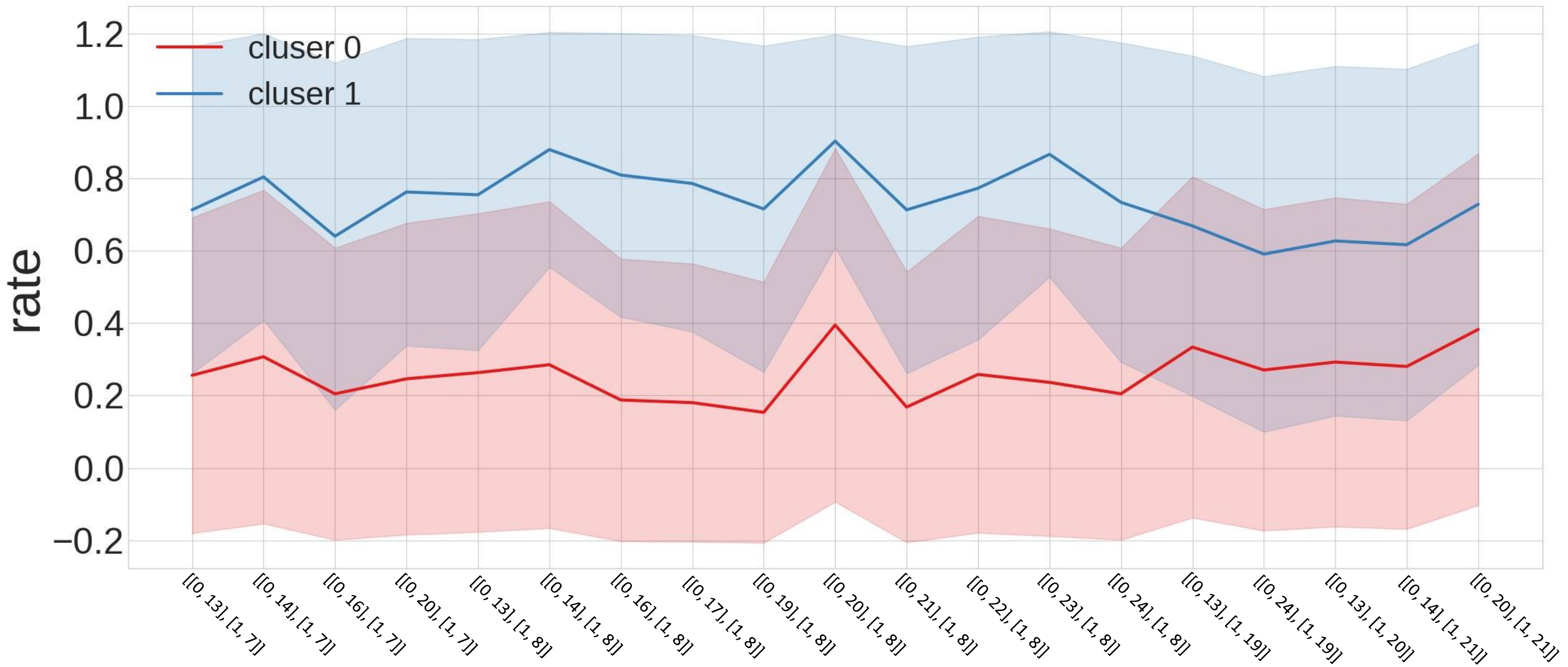}}
  \subfigure[Induction Skill]{
    \includegraphics[width=0.48\linewidth]{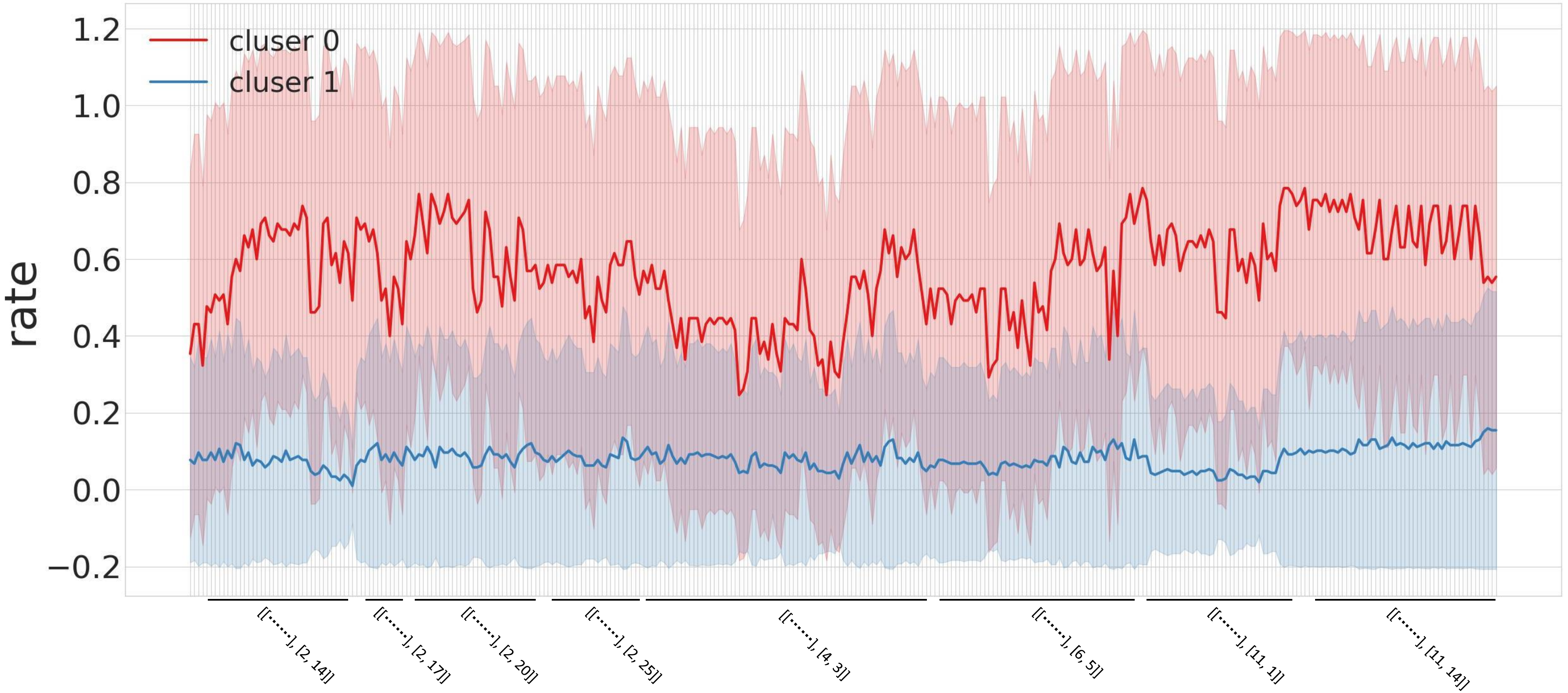}}\\
  \subfigure[ICL1 Skill (icl\_sst2)]{
    \includegraphics[width=0.48\linewidth]{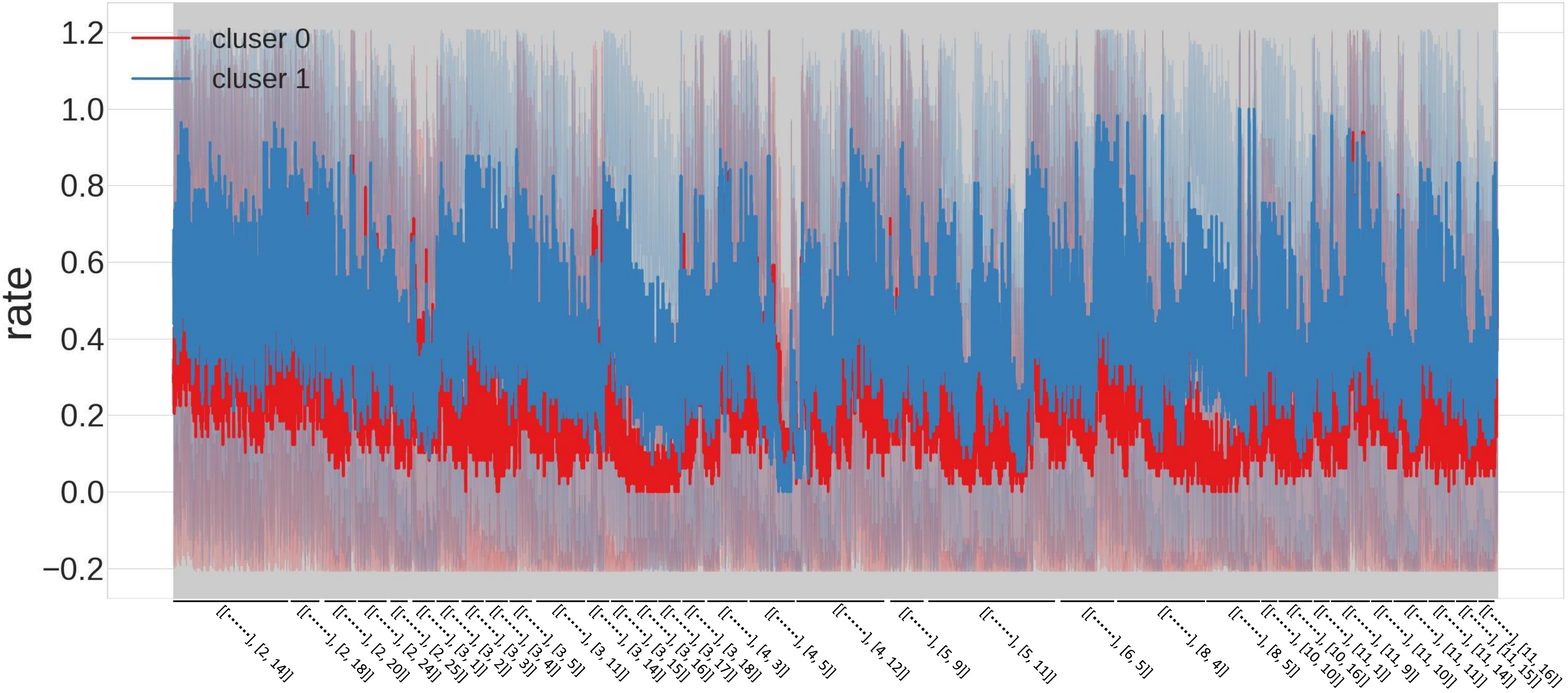}}
  \subfigure[ICL2 Skill (icl\_oc)]{
    \includegraphics[width=0.48\linewidth]{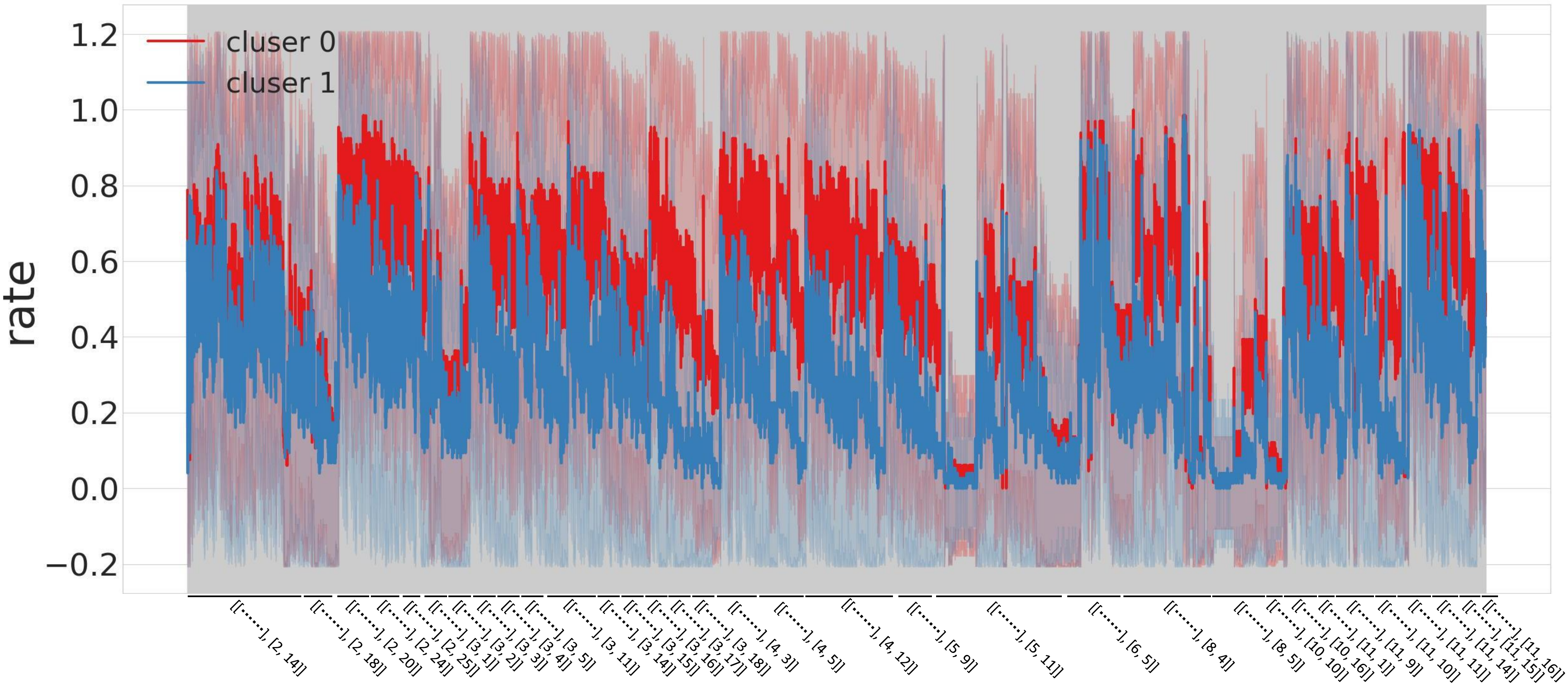}}
  \subfigure[ICL3 Skill (icl\_qa)]{
    \includegraphics[width=0.48\linewidth]{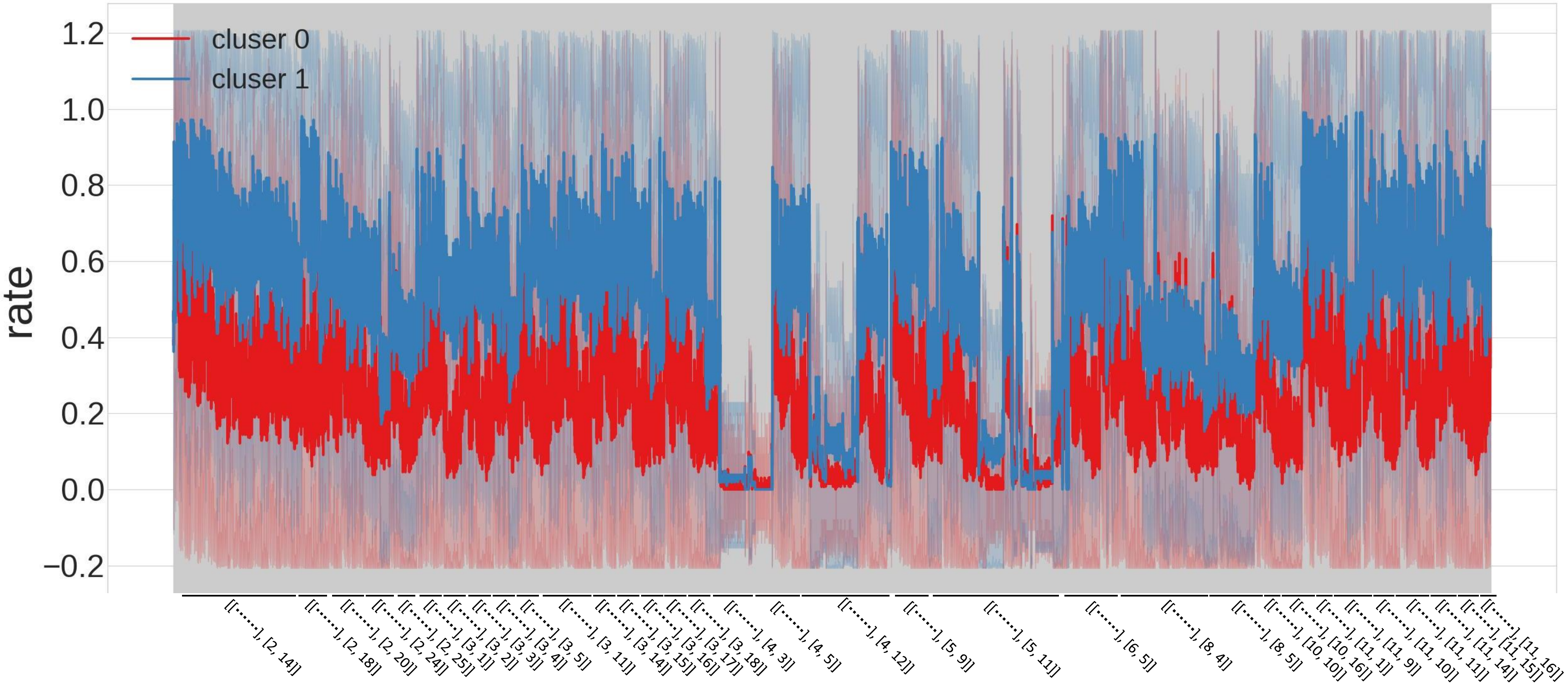}}
  \subfigure[ICL4 Skill (icl\_raco)]{
    \includegraphics[width=0.48\linewidth]{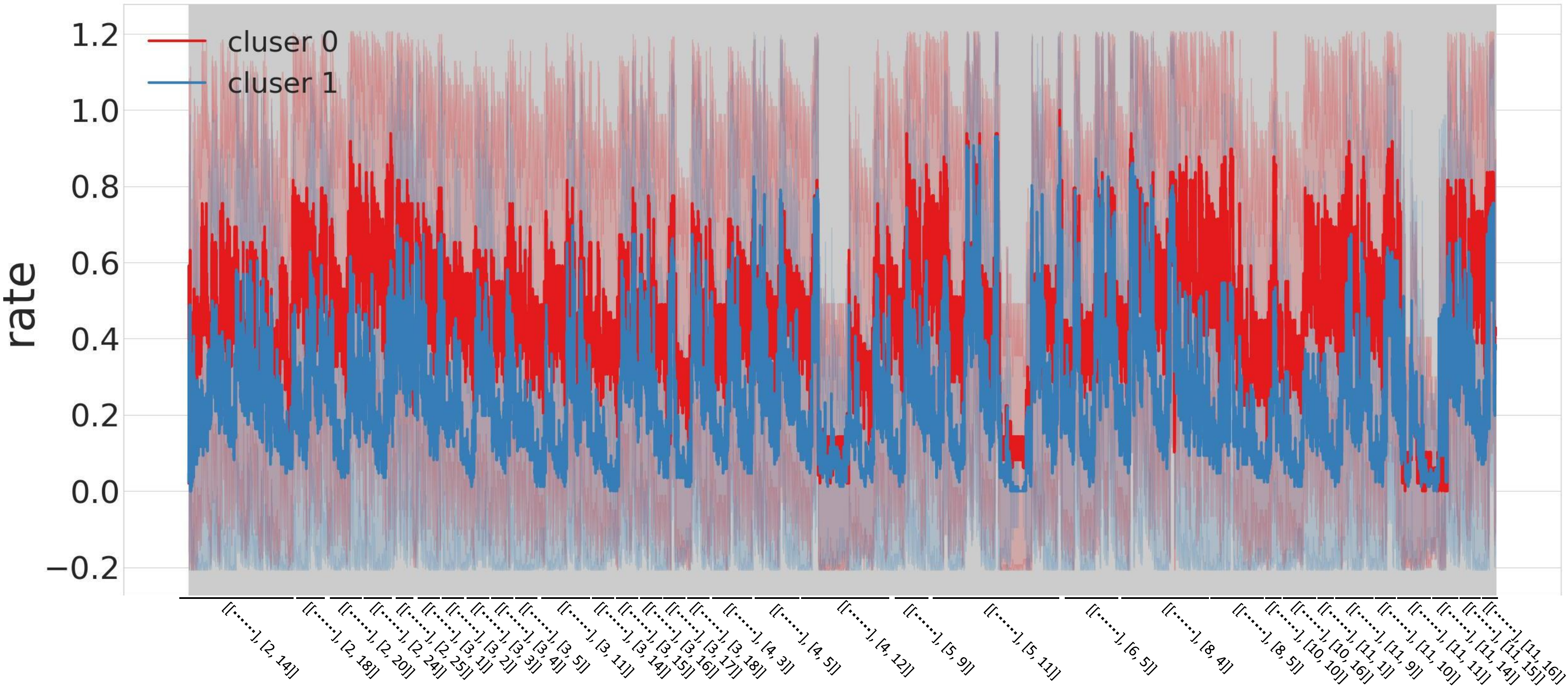}}
  \caption{bisection clustering on paths with top 10\% $Eff_{Skill}$ for 3 skills}
  \label{figcluster2}
\end{figure*}

Figure~\ref{figcluster2} shows the results of our clustering on the $\mathcal{G}^{S}$ for the 3 skills. The x-axis sequentially arranges the top 10\% of paths on $Eff_{Skill}$ from shallow to deep, and the y-axis indicates the mean $Eff_{Skill}$ of these paths. It's striking that two clusters in the Previous Skill and Induction Skill: one consistently showing a high $Eff_{Skill}$, and the other showing little to no $Eff_{Skill}$. This suggests that these low $Eff_{Skill}$ samples hardly share common paths or trigger common language skills. Meanwhile, the ICL skill does not showcase discriminable clustering, further corroborating our speculation.

\begin{figure*}
  \centering
  \subfigure[1-st round clustering]{
    \includegraphics[width=0.48\linewidth]{previouscluster20.pdf}}
  \subfigure[2-nd round clustering]{
    \includegraphics[width=0.48\linewidth]{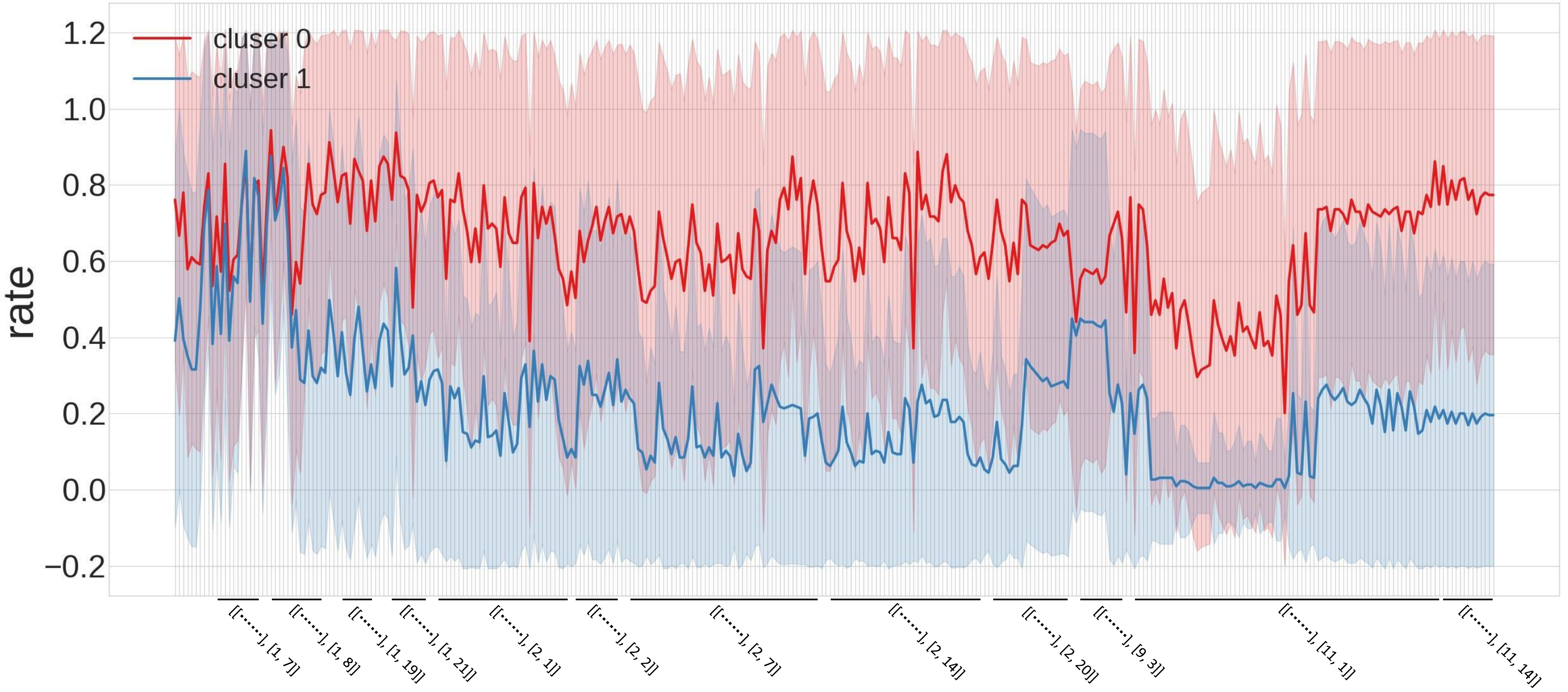}}
  \subfigure[3-rd round clustering]{
    \includegraphics[width=0.48\linewidth]{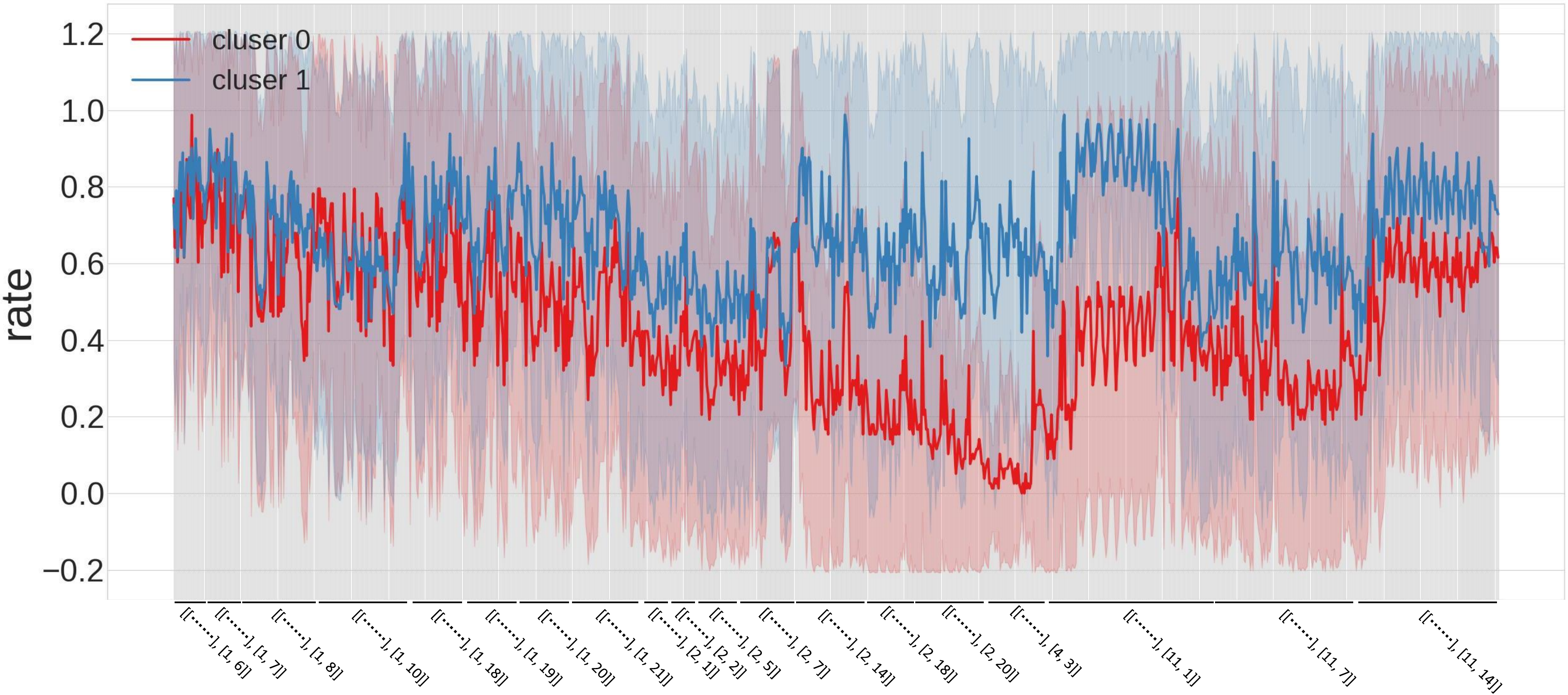}}
  
  \caption{3 rounds clustering in Previous Token Skill}
  \label{figprevious3rounds}
\end{figure*}

\begin{figure*}
  \centering
  \subfigure[1-st round clustering]{
    \includegraphics[width=0.48\linewidth]{inductioncluster20.pdf}}
  \subfigure[2-nd round clustering]{
    \includegraphics[width=0.48\linewidth]{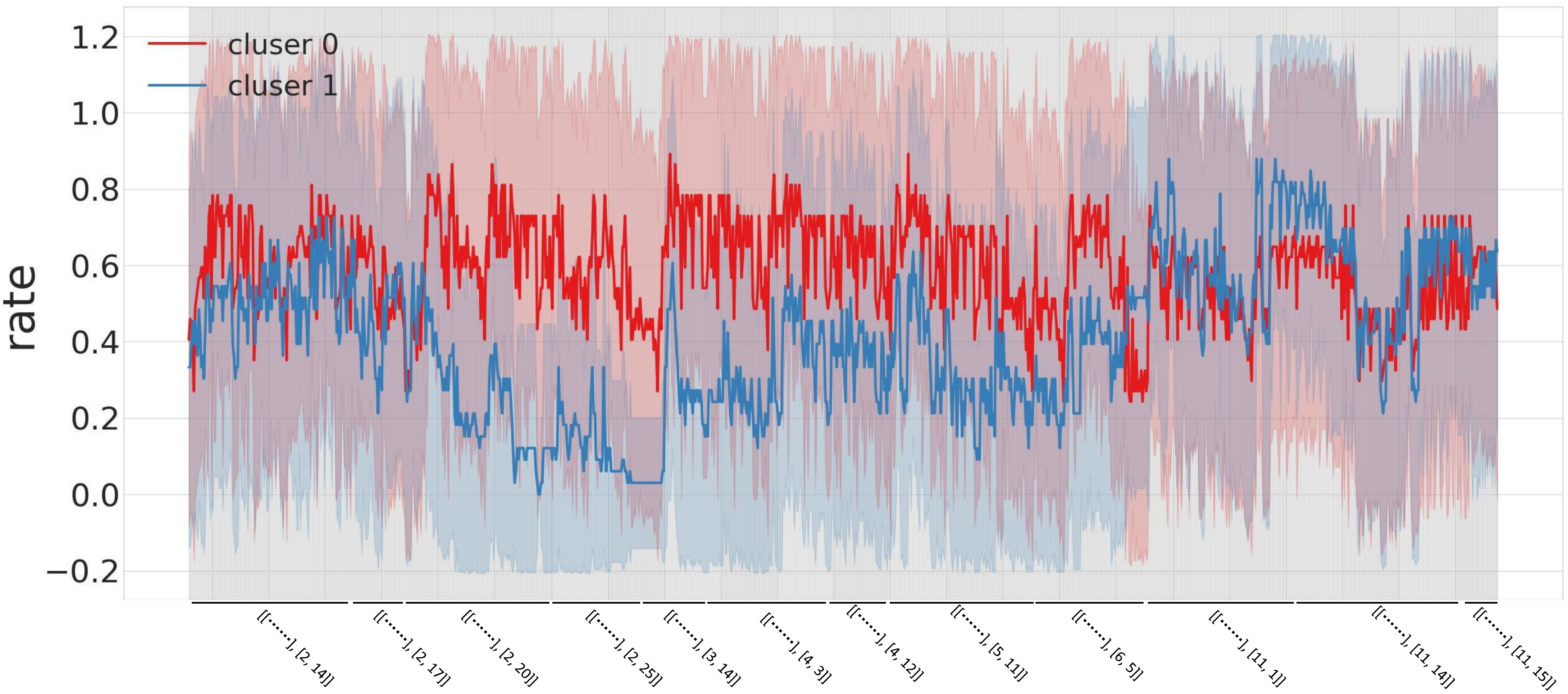}}\\
  
  \caption{2 rounds clustering in Induction Skill}
  \label{figinduction2rounds}
\end{figure*}

Going a step further, we would like to ascertain whether the Previous Token Skill and Induction Skill, after undergoing multiple rounds of ``purification" through clustering, could still be divided into two clusters. Therefore, we recursively performed bisection clustering on the higher $Eff_{Skill}$ cluster each time. Figure~\ref{figprevious3rounds} and~\ref{figinduction2rounds} presents the results after each round of clustering. Notably, the Previous Token could not be divided after 2 rounds of clustering, while the Induction Token hit the dividing limit after just 1 round. Considering that the number of clustering rounds for ICL Skill was 0, we believe this supports our hypothesis: the more complicated the skill, the fewer instances of coincidental samples.

\begin{figure*}
  \centering
  \subfigure[bisection cluster]{
    \includegraphics[width=0.48\linewidth]{inductioncluster20.pdf}}
  \subfigure[trisection cluster]{
    \includegraphics[width=0.48\linewidth]{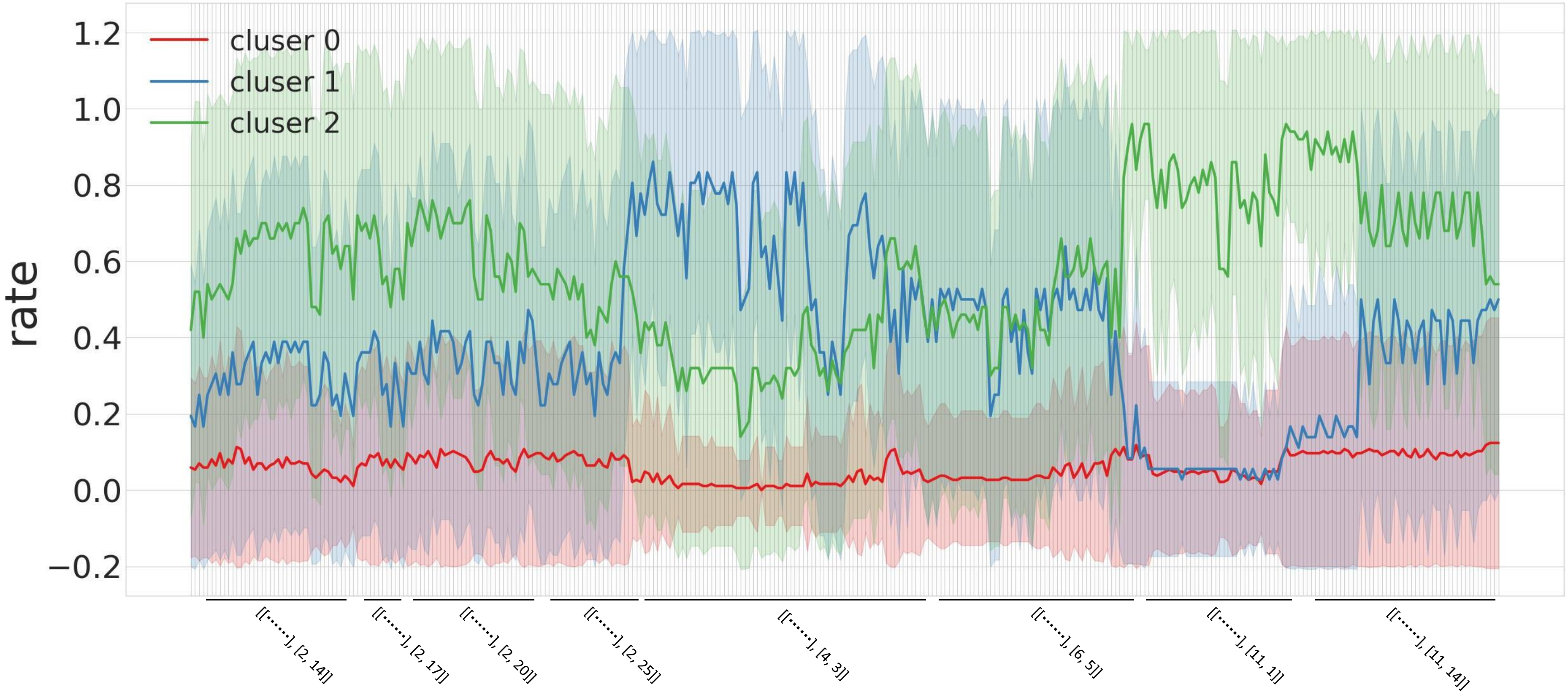}}\\
  \subfigure[quadsection cluster]{
    \includegraphics[width=0.48\linewidth]{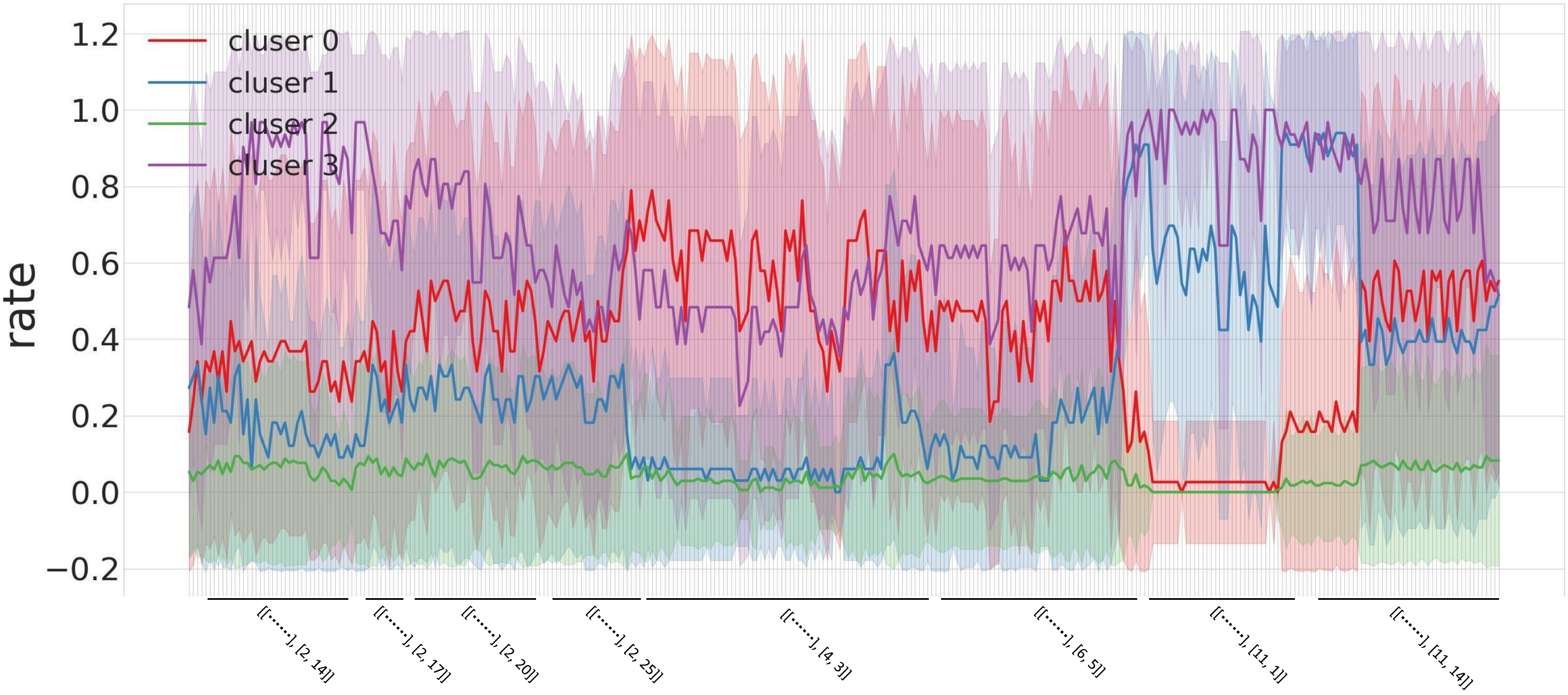}}
  \subfigure[quintisection cluster]{
    \includegraphics[width=0.48\linewidth]{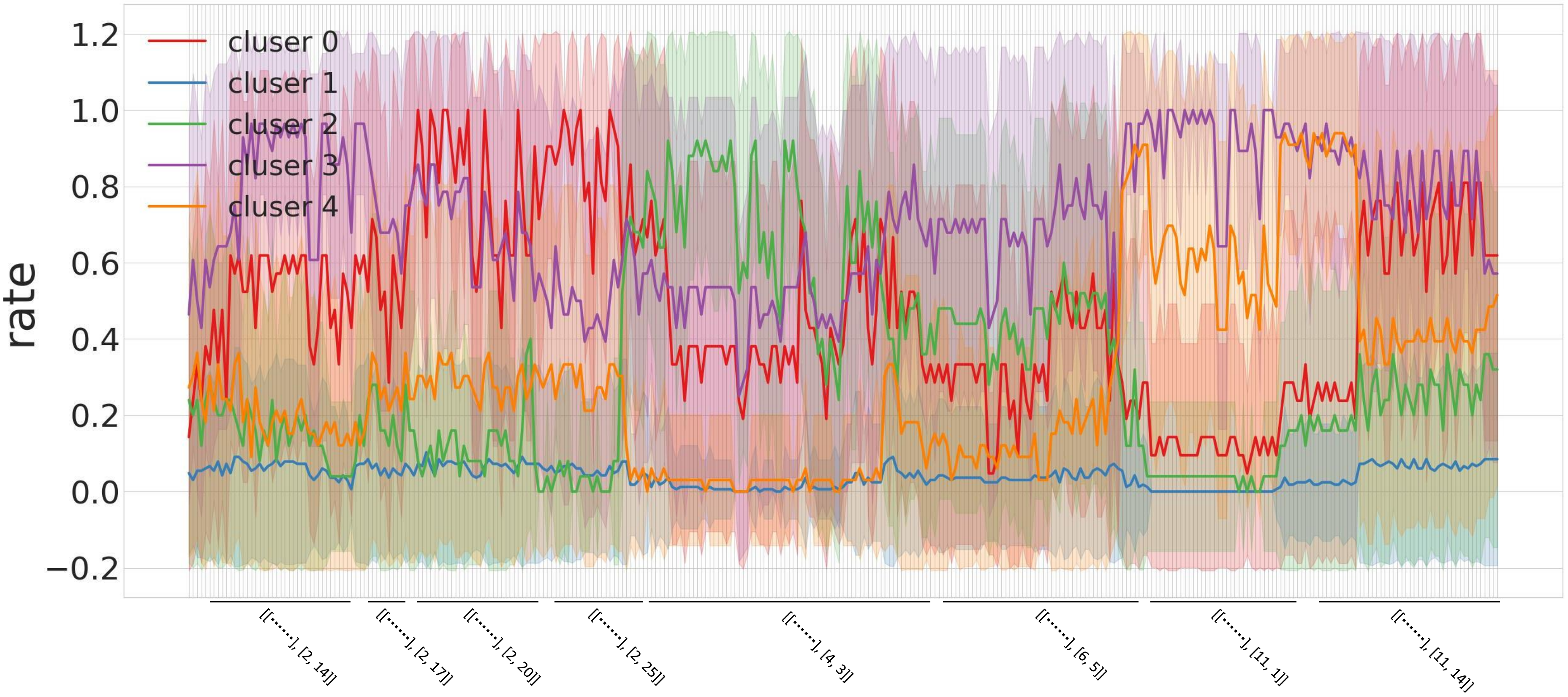}}
  \caption{different clustering on Induction Skill}
  \label{figdiffercluster}
\end{figure*}

Lastly, we verified that bisection clustering significantly outperformed trisection, quad-section, and quintisection clustering. As illustrated in Figure~\ref{figdiffercluster}, out of all the clusterings, only bisection clustering was able to distinctly segregate two mutually exclusive clusters categorized by high and low $Eff_{Skill}$.

\section{More Details about Path Ablation}~\label{suppdataforvalidation}
To enhance the transparency and validity of the validation experiment, we have supplemented it with some additional data.

\begin{figure*}
  \centering
  \includegraphics[width=0.8\linewidth]{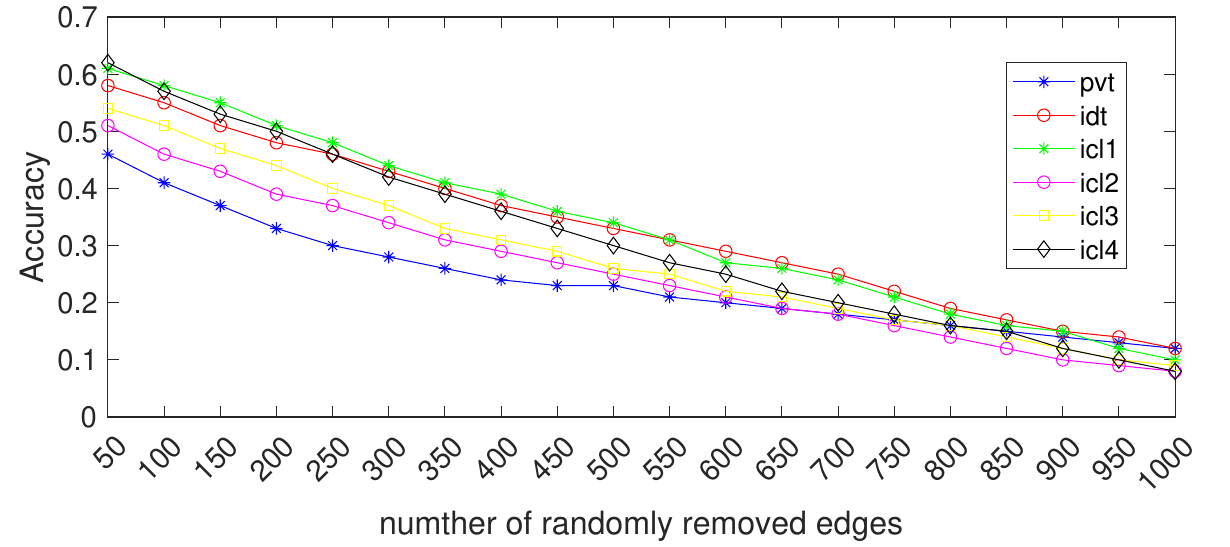}
  \caption{Accuracy with the number of removed edges increasing.} 
\label{figremovededges}
\end{figure*}

Firstly, Table~\ref{tabremoveedge} only provides the accuracy of randomly deleting 50 and 500 edges; however, the dynamics of accuracy is not disclosed as the number of deleted edges changes. Therefore, we demonstrate the dynamics of accuracy in Figure~\ref{figremovededges} when the number of randomly deleted edges ranges from 50 to 1000. Notably, even with 1000 randomly deleted edges, the accuracy still remains above 0.1 (the total number of edges being considered is 6875). However, deleting the skill paths directly leads to an accuracy close to 0, even if the skill paths contain only around 500 edges. This further illustrates that the skill paths contain more mechanisms that significantly determine the final output.

\begin{table*}[t]
\begin{center}
\resizebox{1\linewidth}{!}{
\begin{tabular}{llllllll}
\textbf{Sample}&\multicolumn{7}{c}{\textbf{Circuit Graph}}\\
&$\mathcal{G}*$&$-(\mathcal{G}^{S,\text{PVT}}-\mathcal{G}*)$&$-(\mathcal{G}^{S,\text{IDT}}-\mathcal{G}*)$&$-(\mathcal{G}^{S,\text{ICL1}}-\mathcal{G}*)$&$-(\mathcal{G}^{S,\text{ICL2}}-\mathcal{G}*)$&$-(\mathcal{G}^{S,\text{ICL3}}-\mathcal{G}*)$&$-(\mathcal{G}^{S,\text{ICL4}}-\mathcal{G}*)$\\
\hline
PVT&1.00&1.00&0.88&0.89&0.89&0.83&0.89\\
IDT&1.00&0.93&1.00&0.81&0.82&0.85&0.81\\
ICL1&1.00&0.95&0.81&1.00&0.95&0.93&0.97\\
ICL2&1.00&0.93&0.84&1.00&0.92&0.95&0.92\\
ICL3&1.00&0.94&0.86&1.00&0.93&0.91&0.94\\
ICL4&1.00&0.96&0.83&1.00&0.93&0.94&0.96\\
\end{tabular}}
\end{center}
\caption{Accuracy of output to original label within different Circuit Graph}
\label{tabremovededgessupp}
\end{table*}

Secondly, in Table~\ref{tabremoveedge}, we only showed the situation where low-level skill graphs remove those paths contained in high-level skill graphs. To reinforce the validation, we additionally provide in Table~\ref{tabremovededgessupp} the scenario where samples of low-level skills are only deleted from those paths that exist in the high-level skill graph but not in the low-level skills. 

Herein, $-(\mathcal{G}^{S,\text{PVT}}-\mathcal{G}*)$ represents the deletion of paths in the previous token skill graph that do not exist in the target graph for the target sample, while $-(\mathcal{G}^{S,\text{IDT}}-\mathcal{G}*)$ represents the deletion of paths in the Induction skill graph that do not exist in the target graph. $-(\mathcal{G}^{S,\text{ICL1}}-\mathcal{G}*)$, $-(\mathcal{G}^{S,\text{ICL2}}-\mathcal{G}*)$, $-(\mathcal{G}^{S,\text{ICL3}}-\mathcal{G}*)$, and $-(\mathcal{G}^{S,\text{ICL4}}-\mathcal{G}*)$ respectively represent the deletion of paths in the ICL1, ICL2, ICL3, and ICL4 skill graphs that do not exist in the target graph for the target sample.

Recall that a portion of the paths in the high-level skill graph is identical to a portion of the paths in the low-level skill graph. Table~\ref{tabremovededgessupp} clearly shows that when the target samples delete the paths that exist in other skills but not in their own, the accuracy is not significantly affected. For instance, $-(\mathcal{G}^{S,\text{IDT}}-\mathcal{G}^{S,\text{PVT}})$ deletes 129 paths, but only reduces the sample accuracy of the previous token skill to 0.88, while the accuracy corresponding to randomly deleting 100 edges is only 0.42 (see Figure~\ref{figremovededges}). In conjunction with Table~\ref{tabremoveedge}, it explains that only the overlap of the induction skill graph paths with the previous token skill graph affects the previous token skill. Additionally, when the ICL series skills output paths that exist in other ICLs but not in themselves, their accuracy is somewhat higher (more than 0.9). This is due to the skill graphs in the ICL series being more similar to each other, resulting in fewer paths in the complement.

\section{Grounded in Benchmark: IOI Task}\label{suppioitask}
\begin{table*}
\begin{center}
\resizebox{0.9\textwidth}{!}{
\begin{tabular}{ll}
\textbf{Skill}&\textbf{Receivers with receiving more than 10 paths ([\#layer, \#circuit])}\\
\hline
\textbf{DLT}&[0, 2], [0, 13], [0, 14], [0, 16], [0, 20], [1, 8], [1, 9], [1, 18], [1, 19], [11, 1], [11, 14]\\
\hline
\textbf{PVT}&[1, 8], [1, 18], [1, 19], [1, 20], [1, 21], [2, 1], [2, 7], [2, 14], [2, 18], [2, 20], [2, 22], [2, 24], [11, 1], [11, 14]\\
\hline
\textbf{IDT}&\textcolor{blue}{[2, 14]}, \textcolor{blue}{[2, 18]}, \textcolor{blue}{[2, 20]}, [3, 14], [3, 17] [4, 5], [4, 12], [5, 11], [6, 5], \textcolor{blue}{[11, 1]}, \textcolor{blue}{[11, 14]}\\
\hline
\textbf{SIB}& \textcolor{blue}{[0, 20]}, \textcolor{blue}{[1, 8]}, \textcolor{blue}{[1, 18]}, \textcolor{blue}{[2, 14]}, \textcolor{blue}{[3, 14]}, \textcolor{blue}{[5, 11]}, [5, 14], [7, 9], [7, 20], [7, 21], [8, 7], [8, 18], \textcolor{blue}{[11, 1]}, \textcolor{blue}{[11, 14]}\\
\hline
\textbf{NMV}&\textcolor{blue}{[0, 2]}, \textcolor{blue}{[1, 8]}, \textcolor{blue}{[1, 20]}, \textcolor{blue}{[3, 14]}, [3, 20], \textcolor{blue}{[5, 11]}, \textcolor{blue}{[8, 7]}, [9, 14], [9, 18], [9, 20], [10, 1], [10, 14], [10, 22], \textcolor{blue}{[11, 14]}\\
\hline
\textbf{LPM}& \textcolor{blue}{[1, 8]}, \textcolor{blue}{[2, 18]}, \textcolor{blue}{[2, 20]}, [3, 11], [4, 14], [6, 7], [8, 4], [8, 17], [8, 24], [9, 9]\\
\hline

\end{tabular}}
\end{center}
\caption{Key Receivers in subgraphs of IOI task, blue circuits are presented in the lower skill}
\label{tabreceiversioi}
\end{table*}
Building on the results from Table~\ref{tabreceivers}, we continue to explore the skills required for IOI, which include duplicate token (DLT), previous token (PVT), induction (IDT), S-inhibition (SIB), name mover (NMV), and back-up head~\citep{wang2023interpretability}. For DLT, we found another distinct cluster within the circuit samples of the induction skill. For SIB, we obtained it by replacing ``S2'' with ``IO'' as the background text. For NMV, we obtained it by using a random name as a substitute for ``IO'' and ``S'' in the background text. Interestingly, our method was unable to detect the presence of a back-up head. A reasonable conjecture is that the back-up head acts more like a preemption mechanism, effectively circumvented in path-level causal analysis. Additionally, we present the key nodes of other skills in Table~\ref{tabreceiversioi}. It is evident that the skill paths we demonstrate possess strong inclusivity. For instance, the S-inhibition skill encompasses crucial nodes of the duplicate token, previous token, and induction skills, while the name mover almost includes nodes from all previous skills. Beyond this, we also discovered a long-position mapping (LPM) skill, obtained through a large number of long sentence samples and background text with deleted commas. It represents another advanced skill that extends PVT.

Moreover, based on the paths, attention weights, and cosine similarities of the representations, we have identified several circuits with distinct characteristics (We demonstrate the performances of other circuit discovery methods in validating these conclusions in Appendix~\ref{suppcomvc}.):

\textbf{Preceding Token Circuit}: Circuit [4, 12] performs a unique function, namely, when any token serves as a query token to attend other tokens, this circuit is shown to consistently carry significant information from its preceding token to the query token. 

\textbf{Key Token Circuit}: Circuit [3, 14] exhibits a significantly different function from the others. This circuit consistently focuses on certain key tokens in the preceding text -- such as the beginning, ending, and label prompts -- and transmits this information to subsequent query tokens. Additionally, other key circuits in layers 3 and 4 partially undertake these functionalities.

\textbf{Opposite Circuit}: 
When using the last token of each input to produce the embedding for a specific circuit, we notice that the cosine similarity between Circuit [11, 14] and other key circuits is usually less than $0$, especially with Circuit [11, 1], where the cosine similarity reaches to $-0.92$. Previous work~\citep{wang2023interpretability} has mentioned this phenomenon, hypothesizing the reason to be controlling the variance of the loss function.

We have observed some differences in the receivers of different ICL tasks. Combined with the insights provided by~\citet{bayazit2023discovering} and~\citet{bricken2023monosemanticity}, we suspect that these differences arise from distinct circuits required to process domain-specific knowledge across different tasks.


\section{Validation for Effects from Other Skills}\label{suppObgdf}

\begin{figure*}
  \centering
  \subfigure[PVT $\mathcal{G}_{Ori}* \& \mathcal{G}_{Bkg}*$]{
    \includegraphics[width=0.23\linewidth]{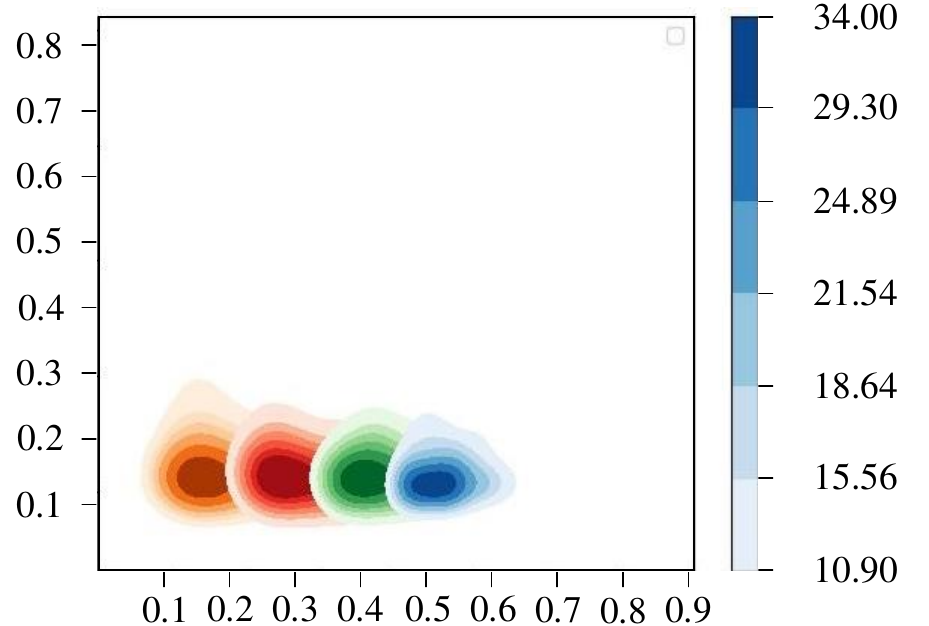}}
  \subfigure[PVT $\mathcal{G}_{Ori}* \& \mathcal{G}_{Self}*$]{
    \includegraphics[width=0.23\linewidth]{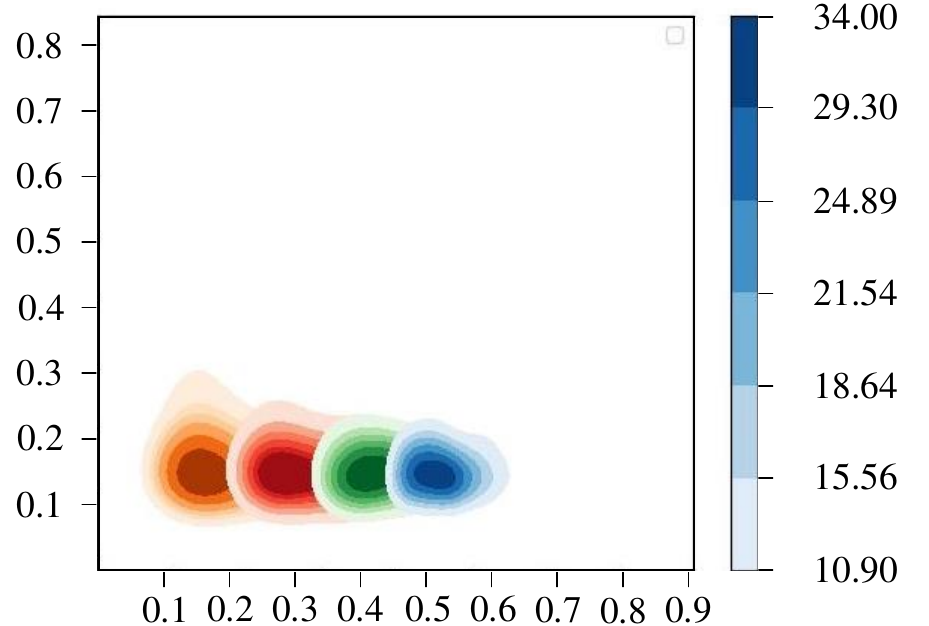}}
  \subfigure[IDT $\mathcal{G}_{Ori}* \& \mathcal{G}_{Bkg}*$]{
    \includegraphics[width=0.23\linewidth]{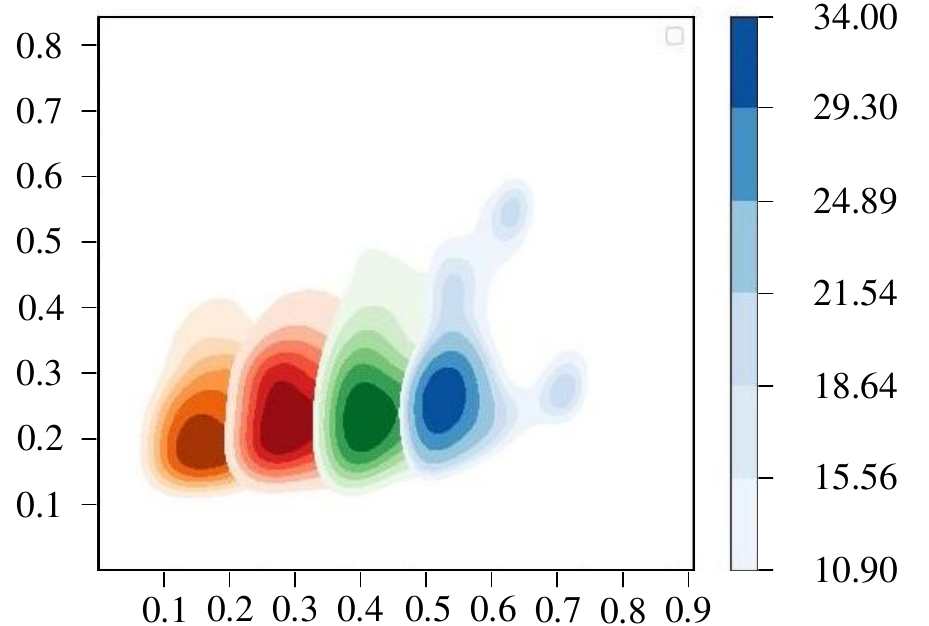}}
  \subfigure[IDT $\mathcal{G}_{Ori}* \& \mathcal{G}_{Self}*$]{
    \includegraphics[width=0.23\linewidth]{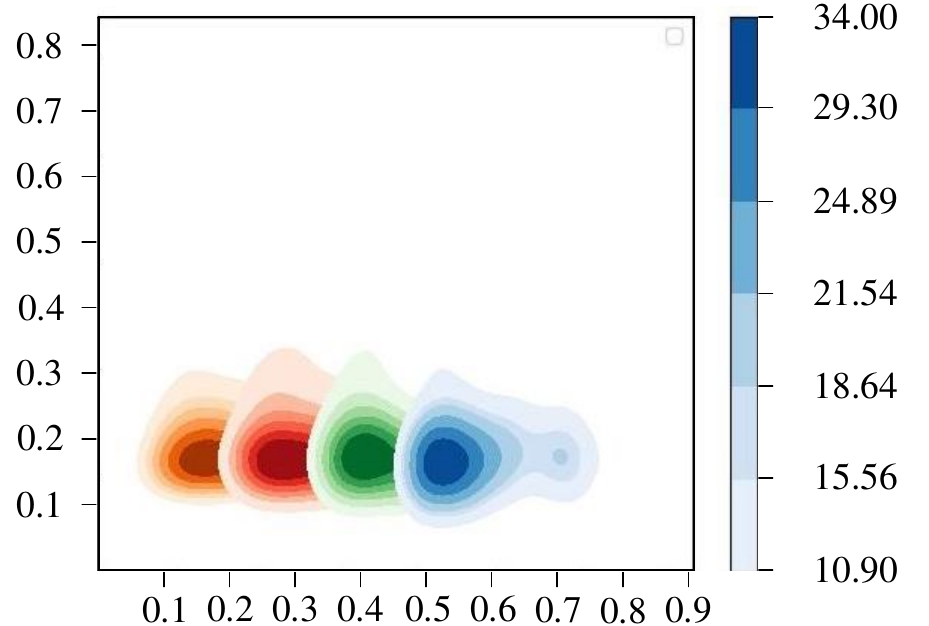}}
  \subfigure[ICL2 $\mathcal{G}_{Ori}* \& \mathcal{G}_{Bkg}*$]{
    \includegraphics[width=0.23\linewidth]{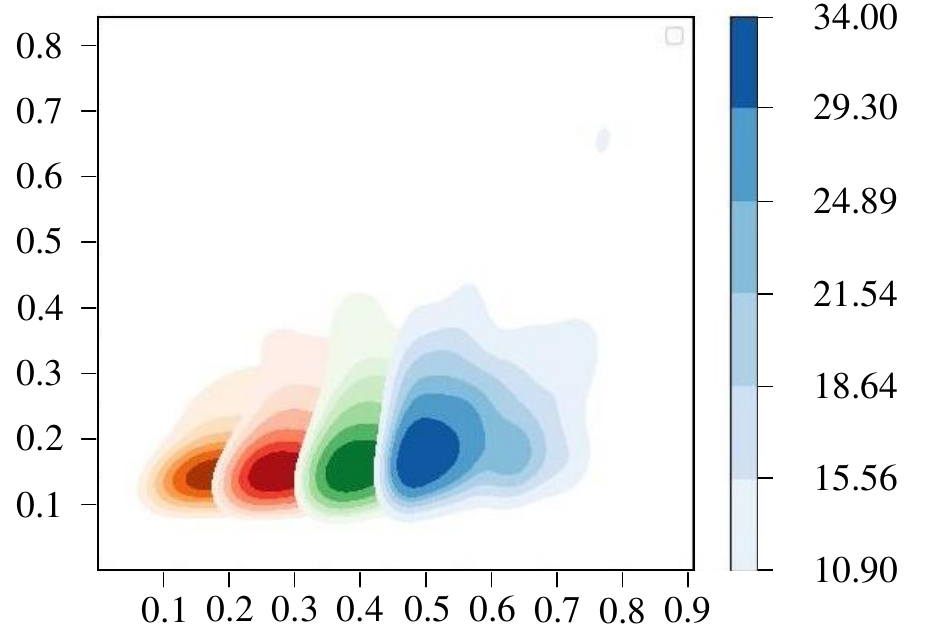}}
  \subfigure[ICL2 $\mathcal{G}_{Ori}* \& \mathcal{G}_{Self}*$]{
    \includegraphics[width=0.23\linewidth]{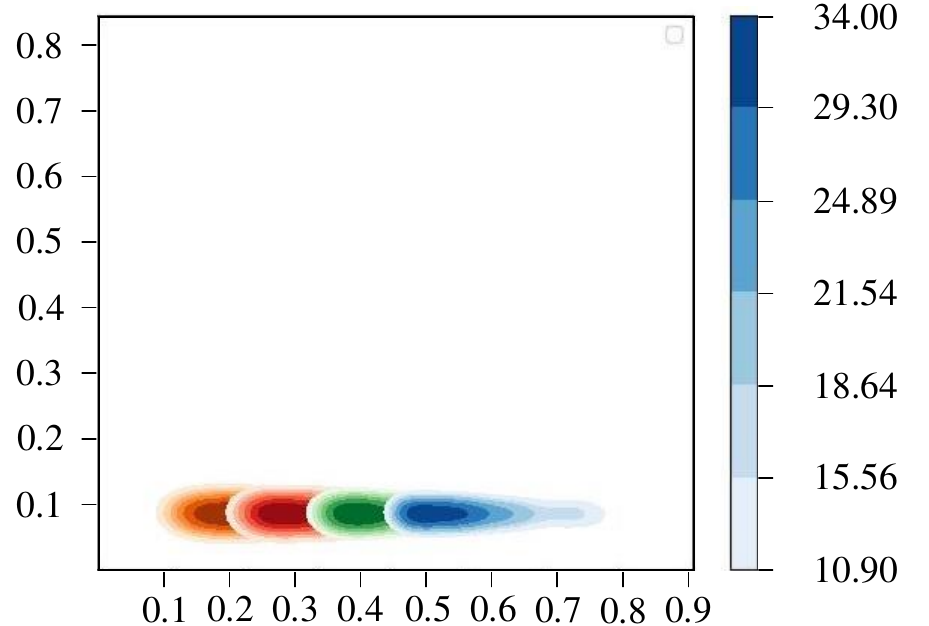}}
  \subfigure[ICL1 $\mathcal{G}_{Ori}* \& \mathcal{G}_{Bkg}*$]{
    \includegraphics[width=0.23\linewidth]{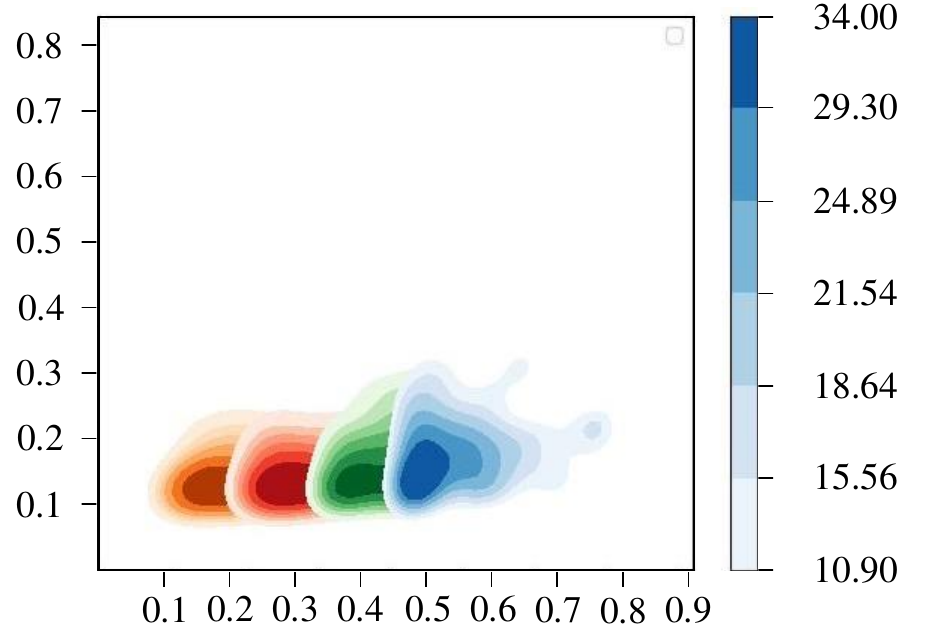}}
  \subfigure[ICL1 $\mathcal{G}_{Ori}* \& \mathcal{G}_{Self}*$]{
    \includegraphics[width=0.23\linewidth]{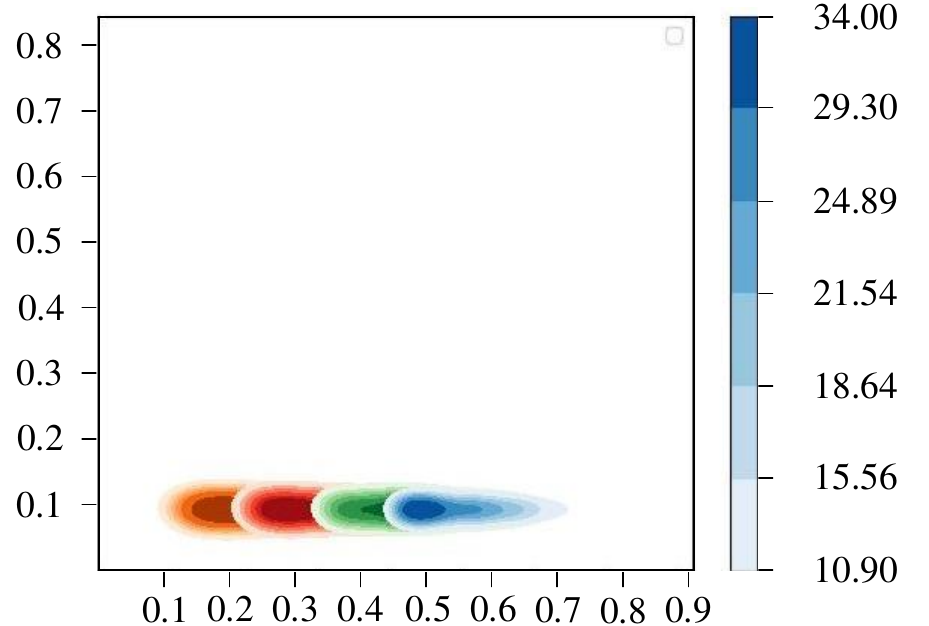}}
  \subfigure[ICL3 $\mathcal{G}_{Ori}* \& \mathcal{G}_{Bkg}*$]{
    \includegraphics[width=0.23\linewidth]{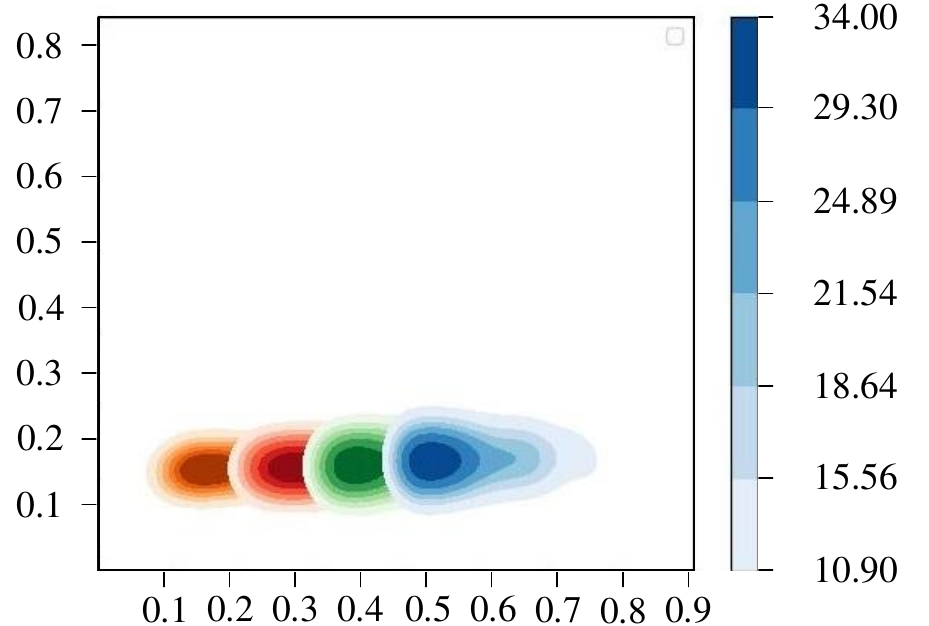}}
  \subfigure[ICL3 $\mathcal{G}_{Ori}* \& \mathcal{G}_{Self}*$]{
    \includegraphics[width=0.23\linewidth]{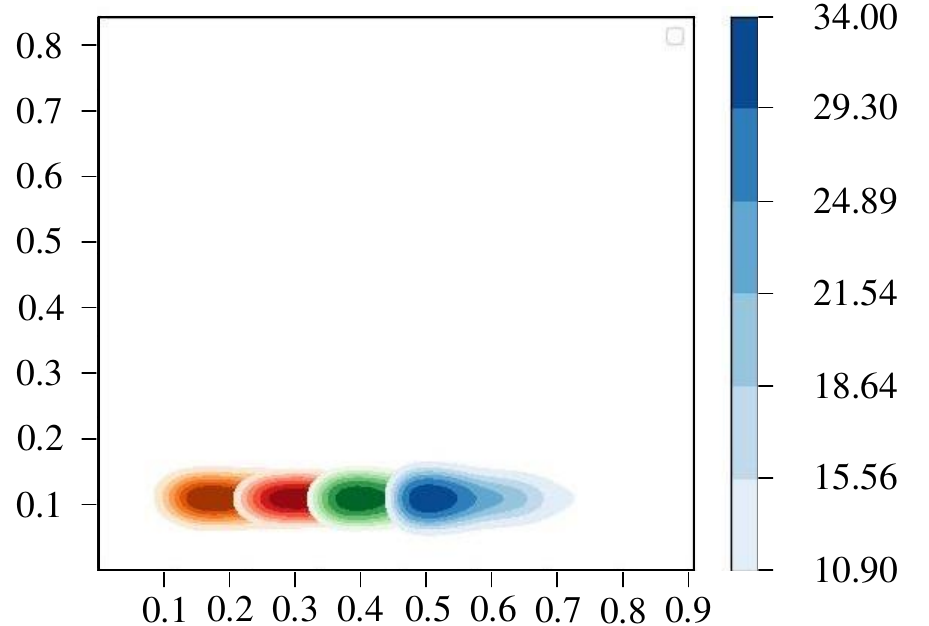}}
  \subfigure[ICL4 $\mathcal{G}_{Ori}* \& \mathcal{G}_{Bkg}*$]{
    \includegraphics[width=0.23\linewidth]{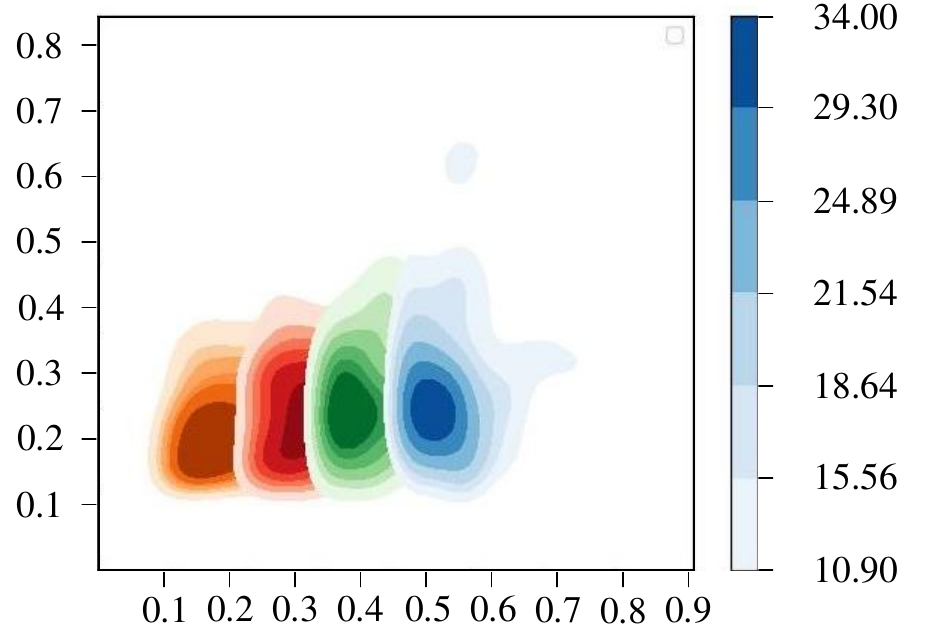}}
  \subfigure[ICL4 $\mathcal{G}_{Ori}* \& \mathcal{G}_{Self}*$]{
    \includegraphics[width=0.23\linewidth]{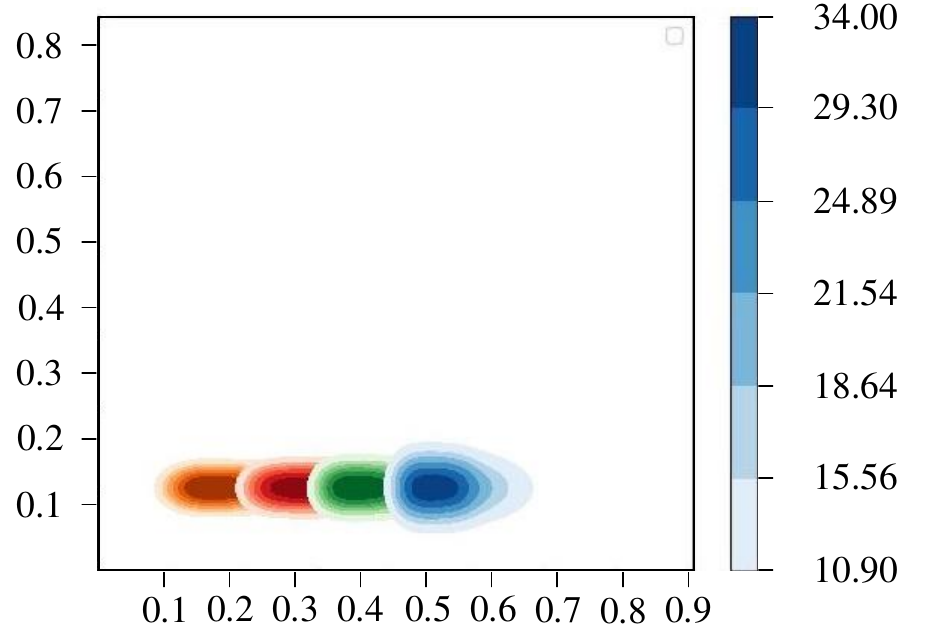}}
  \caption{Bivariate probability density function (PDF) of path effects on Previous Token,Induction, ICL1 ICL2, ICL3, and ICL4 Skills. The x-axis represents the first variable, the path effect in the origin text ($\mathcal{G}_{Ori}*$) while the y-axis represents the second variable, the path effect in the background/self text ($\mathcal{G}_{Bkg}*/\mathcal{G}_{Self}*$). Orange, red, green, and blue respectively represent the distribution of paths with $Eff>0.2, 0.3, 0.4, 0.5$ in the origin text.}
  \label{figbigramdistributionmore}
\end{figure*}

Another question is whether the background effect and the self effect (e.g., potential skills other than the target skill), mentioned in Section~\ref{seccausaleffect}, potentially exist as confounders or share the skill paths. To answer this question, we performed two experiments to show the overlap of three effects. Initially, Table~\ref{tabratiopath} checks the overlap between the paths with $Eff > 0.5$ in the background/self text and the skill paths, illustrating that a small portion (approximately 10\%-20\%) of those paths does not belong to any observed skill. This corresponds to the confounding that originates from other latent skills that we envisioned. Secondly, Figure~\ref{figbigramdistributionmore} visualizes the bivariate probability density function with the original input and background/self text of these paths under different effects. One intriguing discovery is that the confounding skills are more likely to present in the background text than in the self text, and the more complex the skill, the subtler the confounding effect introduced by the self text.

Another question is whether the background effect and self effect, mentioned in Section~\ref{seccausaleffect}, potentially exist as confounders or share the circuits with observed skills? To answer this question, we examine the paths in background/self text with $Eff > 0.5$. Table~\ref{tabratiopath} categorizes these paths into 7 types and displays their ratios. Here, $\mathcal{G}^{S}_{PVT}$ signifies the ratio of those paths found in the Previous Token Skill graph, $\mathcal{G}^{S}_{IDT}$ refers to the ratio of those located in the Induction skill graph, similarly, $\mathcal{G}^{S}_{ICL1}$ to $\mathcal{G}^{S}_{ICL4}$ represents the ratio of paths in corresponding ICL skill graphs, and ``Others'' represents the ratio of paths that do not exist in either skill graphs. Notably, a small fraction of high-effect paths does not belong to any observed skill (approximately 0.1-0.2 in ``Others"); these are the confounding paths we mentioned before. Additionally, we demonstrated the bivariate probability density function (PDF) in Figure~\ref{figbigramdistributionmore}. Bivariate PDF constructed from the origin text as one variable,  and background text or self text as another one variable. Evidently, across all skills, the paths that have a high effect ($Eff>0.5$) in the origin text include a part of paths with a relatively high effect ($Eff>0.5$) in the background text. However, there are nearly ignorable high-effect paths in the self text in ICL skills. We guess that within the ICL skill, the background text and the origin text possess a significantly higher number of tokens compared to the self text, thereby leading to an insignificant effect of the self text.

\begin{table*}
\begin{center}
\resizebox{1\textwidth}{!}{
\begin{tabular}{lllllllllllllll}
\textbf{Skills}&\multicolumn{7}{c}{\textbf{$\mathcal{G}_{Bkg}*$}}&\multicolumn{7}{c}{\textbf{$\mathcal{G}_{Self}*$}}\\
&$\mathcal{G}^{S}_{PVT}$&$\mathcal{G}^{S}_{IDT}$&$\mathcal{G}^{S}_{ICL1}$&$\mathcal{G}^{S}_{ICL2}$&$\mathcal{G}^{S}_{ICL3}$&$\mathcal{G}^{S}_{ICL4}$&Others&$\mathcal{G}^{S}_{PVT}$&$\mathcal{G}^{S}_{IDT}$&$\mathcal{G}^{S}_{ICL1}$&$\mathcal{G}^{S}_{ICL2}$&$\mathcal{G}^{S}_{ICL3}$&$\mathcal{G}^{S}_{ICL4}$&Others\\
\hline
Induction&0.76&-&-&-&-&-&0.24&0.84&-&-&-&-&-&0.16\\
ICL1&0.43&0.38&0.29&0.19&0.25&0.23&0.18&0.51&0.33&0.24&0.16&0.18&0.15&0.15\\
ICL2&0.46&0.37&0.25&0.16&0.19&0.21&0.17&0.61&0.24&0.25&0.14&0.19&0.18&0.15\\
ICL3&0.45&0.35&0.23&0.21&0.15&0.19&0.20&0.60&0.28&0.25&0.16&0.18&0.19&0.11\\
ICL4&0.49&0.36&0.25&0.19&0.26&0.14&0.16&0.61&0.25&0.23&0.19&0.16&0.13&0.13\\
\end{tabular}}
\end{center}
\caption{Ratio of high $Eff$ path ($Eff>0.5$) in $\mathcal{G}_{Bkg}*$ and $\mathcal{G}_{Self}*$ (The sum of ratios $>$ 1 due to overlaps in each item).}
\label{tabratiopath}
\end{table*}
Additionally, Table~\ref{tabratiopath} also shows that a part of high-effect paths in the background/self text is common with the corresponding skill graph. Fortunately, we need not worry that removing these paths would render the final Skill Graph (paths) incomplete. Appendix~\ref{suppmultistep} provides evidence that these removed but common paths can always be restored through multi-step paths (We explain this phenomenon as `Inclusiveness' in Section~\ref{secdiscovery}.).

We have supplemented the bivariate distribution figures for Previous Token, ICL2, ICL3, and ICL4, as depicted in Figure~\ref{figbigramdistributionmore}. 

\section{Detailed Implementation of Substitute Other Strategies for Comparison}
\label{suppdcomvc} 

\subsection{Details about Baselines}\label{suppdetailsbaseline}

\textbf{ACDC}~\citep{conmy2023towards}, Automatic Circuit DisCovery, which calculates the importance score of each edge and performs a greedy search based on the score.

\textbf{E-pruning }~\citep{bhaskar2024finding}, which converts the importance score into an optimization function and assigns a learnable parameter to each edge to indicate whether an edge needs to be deleted.

\textbf{EAP}~\citep{syed2023attribution}, or Edge Attribution Patching, which makes a linear approximation of activation patching to assign an importance score to each edge, and retains the top-$k$ edges. 

\textbf{DiscoGP}~\citep{yu2024functional}, a differentiable circuit discovery algorithm with joint weight and edge computation graph pruning, by placing a set of learnable binary mask parameters at the weights of LM components and their interconnecting edges. 

\textbf{Scrubbing}~\citep{chan2022causal}, which starts from the output and recursively finds all the invariances of parts of the neural network that are implied by the hypothesis, and then replaces the activations of the neural network with the maximum entropy distribution subject to certain natural constraints implied by the hypothesis and the data distribution. 

For these pruning strategies, we adopted the interchange ablation method by default, with the construction of the corrupted text being identical to the background text outlined in Appendix~\ref{suppdataimple}. Additionally, we controlled the granularity of the components to be consistent with the granularity used in Section~\ref{secmemorycircuit}. For example, in the ACDC algorithm, we ignored the qkv edges and used a complete attention component as a replacement. The causal effect metrics were calculated using KL divergence, with a threshold set such that the number of retained edges in the circuit graph was above 4000 (as pruning too many edges would lead to excessive elimination of path-level causal effects).




\subsection{Definition of $ovlp(A, B)$}\label{suppovlp}
The rule for calculating overlap is as follows: Let $ovlp(A, B)$ represent the rate of edges in the skill graph $A$ that is also present in the skill graph $B$. For any edge $e^{i}$ on the skill graph $A$, we set an overlap flag $f_{A,B}(e^{i})$. If $e^{i}$ in $A$ also exists in the skill graph $B$, then $f_{A,B}(e^{i})=1$, otherwise $f_{A,B}(e^{i})=0$. For a circuit graph $A$ with $N_{A}$ edges, its set of edges is $\mathcal{E}_{A}$. Our overlap is calculated as $ovlp(A, B)=\frac{1}{N_{A}}\sum^{\mathcal{E}_{A}}_{e^{i}\in \mathcal{E}_{A}}f_{A,B}(e^{i})$.

\section{Threshold and Faithfulness}\label{suppthreshold}
\begin{figure*}
  \centering
  \includegraphics[width=0.8\linewidth]{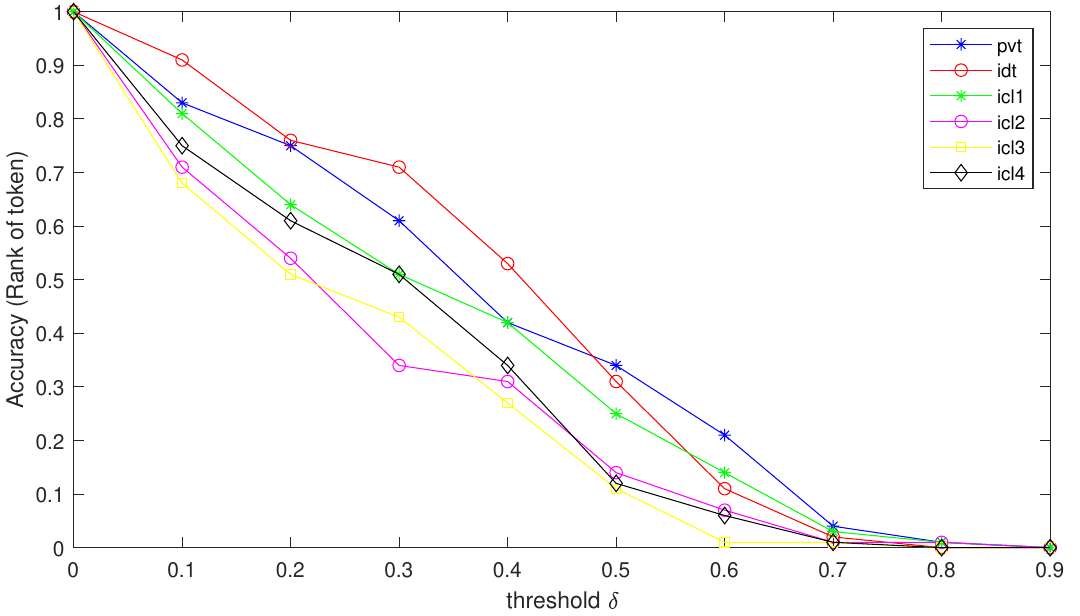}
  \caption{Faithfulness ranging from the $\delta$} 
\label{figthreshold}
\end{figure*}
Although we maintain faithfulness on $\mathcal{G}*$, it is difficult to maintain it on $\mathcal{G}^{S}$. In other words, the bias introduced by counterfactuals and interventions is indeed hard to completely avoid, while the faithfulness of pruning is avoidable. Therefore, a circuit graph that clearly reflects the final result will certainly discard some edges of unclear significance. This is usually accomplished through a threshold. We show in Figure~\ref{figthreshold} the change in accuracy when the threshold $\delta$ mentioned in Section~\ref{seccausaleffect} ranges from 0 to 0.9 (there are almost no circuits left when $\delta >0.9$, so we ignore this part). It can be clearly seen that faithfulness can only be fully guaranteed when $\delta =0$. However, such edges are not sparse enough to reflect some specific interpretable functions. When $\delta >0.7$, it is almost impossible to recover from the original input, but the skill paths obtained can well correspond to the previous methods. Additionally, in this paper, we default the $\delta$ for each skill to PVT: 0.6, IDT: 0.7, ICL1-4: 0.8. 
\begin{figure*}
  \centering
  \includegraphics[width=0.8\linewidth]{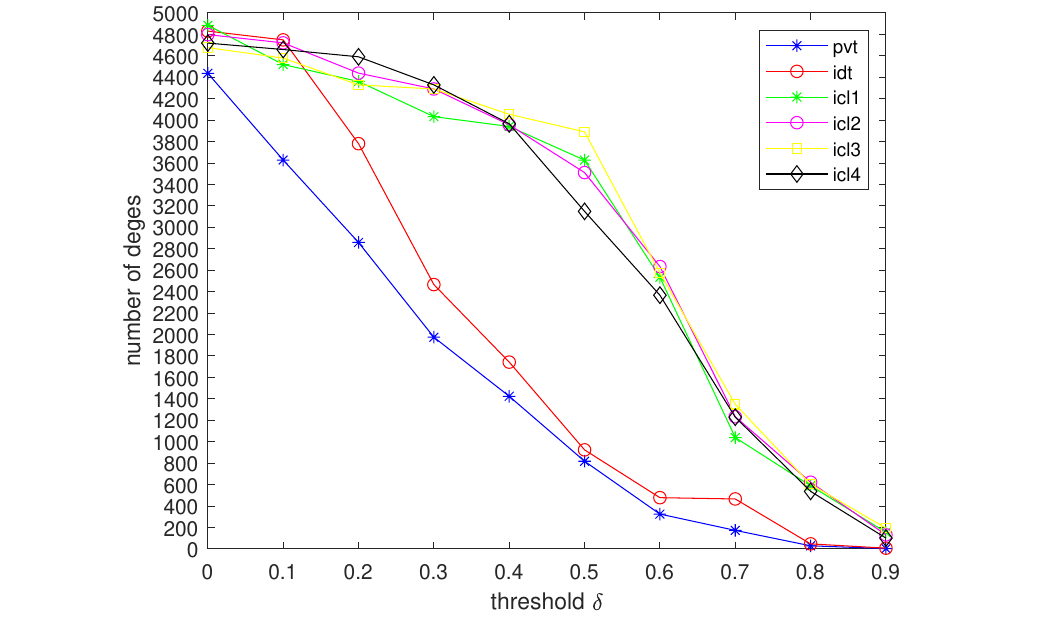}
  \caption{number of edges ranging from the $\delta$} 
\label{figedgedynamic}
\end{figure*}
\begin{figure*}
  \centering
  \includegraphics[width=0.8\linewidth]{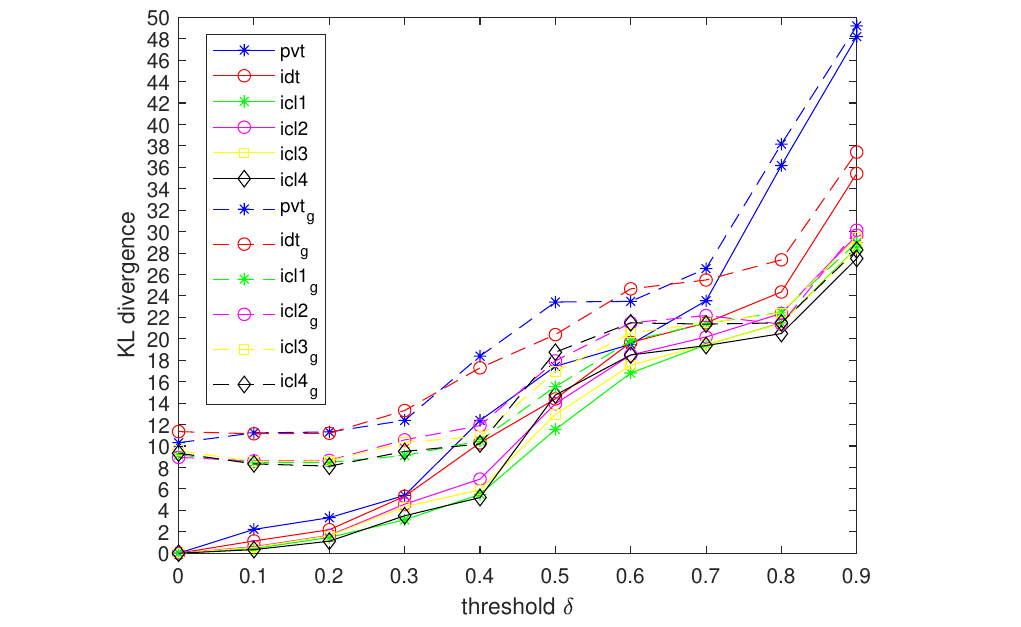}
  \caption{KL divergence ranging from the $\delta$, the solid lines represents KL between $\mathcal{G}^{S}$ and $\mathcal{G}*$, and the dash lines represents KL between $mathcal{G}^{S}$ and $\mathcal{G}$.} 
\label{figKLdynamic}
\end{figure*}
Furthermore, in Figures~\ref{figedgedynamic} and~\ref{figKLdynamic} the changes in the number of edges and the continuous KL divergence metric with varying thresholds $\delta$. Specifically, Figure~\ref{figedgedynamic} presents the total number of edges in the circuit graph (excluding the compensation component and bias circuit) under different thresholds, while Figure~\ref{figKLdynamic} shows the KL divergence between $\mathcal{G}^{S}$ and $\mathcal{G}*$ (solid lines) and $mathcal{G}^{S}$ and $\mathcal{G}$ (dashed lines) obtained at different thresholds. Figure~\ref{figedgedynamic} clearly indicates that the edges with high causal effects of the previous token skill are the fewest, and the most are from the series of ICL skills, which corroborates the conclusion drawn from the clustering in Appendix~\ref{suppimple}. Moreover, the changes in KL divergence (Figure~\ref{figKLdynamic}) can be roughly divided into four phases (steady, burst, steady, burst). In conjunction with Figure~\ref{figedgedynamic}, the two bursts are due to the rapid decrease in edges and the number of edges being too few, approaching zero. The default $\delta$ we selected (PVT 0.6, IDT 0.7, ICL1-4 0.8) are each in the second steady phase. Combining Figures~\ref{figedgedynamic} and\ref{figKLdynamic}, it suggests that when a large number of edges are deleted, the circuit graph enters a phase of minimal change, which we believe best achieves the ``balance between faithfulness and sparsity". 

Furthermore, we can observe that the KL divergence between $\mathcal{G}*$ and $\mathcal{G}$ is approximately 10 (as can be seen from the solid and dashed lines corresponding to $\delta=0$), and generally the KL divergence between $\mathcal{G}^{S}$ and $\mathcal{G}$ ($KL(\mathcal{G}^{S}, \mathcal{G})$) is greater than the KL divergence between $\mathcal{G}^{S}$ and $\mathcal{G}*$ ($KL(\mathcal{G}^{S}, \mathcal{G}*)$). Interestingly, as $\delta$ increases, the values between $KL(\mathcal{G}^{S}, \mathcal{G})$ and $KL(\mathcal{G}^{S}, \mathcal{G})$ get closer and are almost the same at the default threshold.

\section{Sensitivity about Background Text}\label{suppbackground}
To compare the sensitivity brought about by different background texts, we designed four different background text formats on the induction skill and compared the changes between the irreducible circuit graph ($\mathcal{G}^{*}$) of these background texts and the final skill graph ($\mathcal{G}^{S}$). These formats are as follows:

\textbf{Bkg1}: For the induction text ``\textit{......A1 B......A2}", we replace \textit{A2} with the output of the large model for ``\textit{......A1 B......}". For example, if the induction text is ``\textit{Chinese lesson 1.2: Chinese}", the background text is ``\textit{Chinese lesson 1.2: The}".

\textbf{Bkg2}: For the induction text ``\textit{......A1 B......A2}", we directly delete \textit{A2}. For example, if the induction text is ``\textit{Chinese lesson 1.2: Chinese}", the background text is ``\textit{Chinese lesson 1.2: }".

\textbf{Bkg3}: For the induction text ``\textit{......A1 B......A2}", we directly delete \textit{A1}. For example, if the induction text is ``\textit{Chinese lesson 1.2: Chinese}", the background text is `` \textit{lesson 1.2: Chinese}".

\textbf{Bkg4}: For the induction text ``\textit{......A1 B......A2}", we replace \textit{B} with the output of the large model for ``\textit{......A1}". For example, if the induction text is ``\textit{Chinese lesson 1.2: Chinese}", the background text is ``\textit{Chinese people 1.2: Chinese}".

To intuitively feel these changes, we introduced a metric of percentage Hamming distance, \textit{HP}, specifically $HP(G_1, G_2)=hamming distance(G_1, G_2)/(\sum_{G_1}\mathcal{E}+\sum_{G_2}\mathcal{E})*100\%$, i.e., when HP=0\%, it means that the two graphs $G_1$ and $G_2$ completely overlap, and when HP=100\%, it means that the two graphs do not overlap at all. We show the HP between $\mathcal{G}^{*}_{\text{Bkg}}$ and the HP between $\mathcal{G}^{S}$ in any two background texts in Tables~\ref{tabdifferentdatatype}. 

\begin{table*}
  \centering
\resizebox{1\linewidth}{!}{
  \resizebox{0.4\linewidth}{!}{
 
		\centering
            \subtable[HP on $\mathcal{G}^{*}_{\text{Bkg}}$]{
		\begin{tabular}{c|cccc}
		&Bkg1&Bkg2&Bkg3&Bkg4\\
            \hline
            Bkg1&0\% &12.54\% &9.33\% &11.42\% \\
            Bkg2&12.54\% &0\% &6.42\% &9.52\% \\
            Bkg3& 9.33\%&6.42\% &0\% & 12.91\%\\
            Bkg4&11.42\%  &9.52\% &12.91\% &0\% \\
		\end{tabular}}}
 \resizebox{0.4\linewidth}{!}{
 
		\centering
            \subtable[HP on $\mathcal{G}^{S}$]{
		\begin{tabular}{c|cccc}
		&Bkg1&Bkg2&Bkg3&Bkg4\\
            \hline
            Bkg1&0\% & 4.37\%&5.75\% & 4.62\%\\
            Bkg2&4.37\% &0\% &3.51\% &4.03\% \\
            Bkg3&5.75\%  &3.51\% &0\% &3.72\% \\
            Bkg4& 4.62\%&4.03\% &3.72\% &0\% \\
		\end{tabular}}}
 }
 \caption{HP between different background text. For example, the value in the second row and third column of Figure a is 6.42\%, which means $HP(\mathcal{G}^{*}_{\text{Bkg2}}, \mathcal{G}^{*}_{\text{Bkg3}})=6.42\%$ ($\mathcal{G}^{*}_{\text{Bkg2}}$ and $ \mathcal{G}^{*}_{\text{Bkg3}}$ has 6.42\% edges different).}
   \label{tabdifferentdatatype}
\end{table*}

\section{Inclusive Path}\label{suppmultistep}
we have listed all the paths for the previous token skills, all the multi-step paths for the induction and ICL1 skills in the following, with index of the sender component, the first receive component, the second receive component.... The \textcolor{blue}{blue} represents the paths that involve inclusive paths.

\textbf{Previous Token Skill}

\textit{
layer 0 circuit 13, layer 1 circuit 6, with effect 0.71\\
layer 0 circuit 14, layer 1 circuit 7, with effect 0.82\\
layer 0 circuit 16, layer 1 circuit 7, with effect 0.7\\
layer 0 circuit 20, layer 1 circuit 7, with effect 0.86\\
layer 0 circuit 14, layer 1 circuit 8, with effect 0.79\\
layer 0 circuit 16, layer 1 circuit 8, with effect 0.78\\
layer 0 circuit 17, layer 1 circuit 8, with effect 0.81\\
layer 0 circuit 19, layer 1 circuit 8, with effect 0.72\\
layer 0 circuit 20, layer 1 circuit 8, with effect 0.88\\
layer 0 circuit 22, layer 1 circuit 8, with effect 0.81\\
layer 0 circuit 23, layer 1 circuit 8, with effect 0.87\\
layer 0 circuit 24, layer 1 circuit 8, with effect 0.75\\
layer 0 circuit 13, layer 1 circuit 18, with effect 0.79\\
\textcolor{blue}{layer 0 circuit 13, layer 1 circuit 19}, with effect 0.89\\
\textcolor{blue}{layer 0 circuit 14, layer 1 circuit 19}, with effect 0.83\\
\textcolor{blue}{layer 0 circuit 15, layer 1 circuit 19}, with effect 0.74\\
\textcolor{blue}{layer 0 circuit 16, layer 1 circuit 19}, with effect 0.81\\
layer 0 circuit 20, layer 1 circuit 19, with effect 0.82\\
\textcolor{blue}{layer 0 circuit 24, layer 1 circuit 19}, with effect 0.84\\
layer 0 circuit 13, layer 1 circuit 20, with effect 0.84\\
\textcolor{blue}{layer 0 circuit 14, layer 1 circuit 20}, with effect 0.81\\
\textcolor{blue}{layer 0 circuit 20, layer 1 circuit 20}, with effect 0.8\\
layer 0 circuit 13, layer 1 circuit 21, with effect 0.78\\
layer 0 circuit 14, layer 1 circuit 21, with effect 0.83\\
layer 0 circuit 16, layer 1 circuit 21, with effect 0.79\\
layer 0 circuit 17, layer 1 circuit 21, with effect 0.75\\
layer 0 circuit 20, layer 1 circuit 21, with effect 0.87\\
layer 0 circuit 22, layer 1 circuit 21, with effect 0.77\\
layer 0 circuit 23, layer 1 circuit 21, with effect 0.77\\
layer 0 circuit 24, layer 1 circuit 21, with effect 0.75\\
layer 0 circuit 23, layer 2 circuit 1, with effect 0.8\\
layer 0 circuit 24, layer 2 circuit 1, with effect 0.81\\
layer 1 circuit 13, layer 2 circuit 1, with effect 0.76\\
layer 1 circuit 15, layer 2 circuit 1, with effect 0.79\\
layer 1 circuit 16, layer 2 circuit 1, with effect 0.75\\
layer 1 circuit 17, layer 2 circuit 1, with effect 0.75\\
layer 1 circuit 20, layer 2 circuit 1, with effect 0.82\\
layer 0 circuit 13, layer 1 circuit 20, layer 2 circuit 1, with effect 0.74\\
layer 1 circuit 21, layer 2 circuit 1, with effect 0.8\\
layer 0 circuit 20, layer 1 circuit 21, layer 2 circuit 1, with effect 0.77\\
layer 1 circuit 22, layer 2 circuit 1, with effect 0.76\\
layer 1 circuit 23, layer 2 circuit 1, with effect 0.79\\
layer 1 circuit 24, layer 2 circuit 1, with effect 0.8\\
\textcolor{blue}{layer 0 circuit 20, layer 2 circuit 14}, with effect 0.74\\
\textcolor{blue}{layer 0 circuit 21, layer 2 circuit 14}, with effect 0.75\\
\textcolor{blue}{layer 0 circuit 22, layer 2 circuit 14}, with effect 0.77\\
\textcolor{blue}{layer 0 circuit 23, layer 2 circuit 14}, with effect 0.72\\
\textcolor{blue}{layer 0 circuit 24, layer 2 circuit 14}, with effect 0.84\\
\textcolor{blue}{layer 1 circuit 13, layer 2 circuit 14}, with effect 0.72\\
\textcolor{blue}{layer 1 circuit 15, layer 2 circuit 14}, with effect 0.8\\
\textcolor{blue}{layer 1 circuit 16, layer 2 circuit 14}, with effect 0.72\\
\textcolor{blue}{layer 1 circuit 17, layer 2 circuit 14}, with effect 0.8\\
\textcolor{blue}{layer 1 circuit 18, layer 2 circuit 14}, with effect 0.74\\
\textcolor{blue}{layer 1 circuit 20, layer 2 circuit 14}, with effect 0.79\\
\textcolor{blue}{layer 1 circuit 21, layer 2 circuit 14}, with effect 0.79\\
layer 0 circuit 14, layer 1 circuit 21, layer 2 circuit 14, with effect 0.71\\
layer 0 circuit 20, layer 1 circuit 21, layer 2 circuit 14, with effect 0.77\\
\textcolor{blue}{layer 1 circuit 22, layer 2 circuit 14}, with effect 0.81\\
\textcolor{blue}{layer 1 circuit 23, layer 2 circuit 14}, with effect 0.76\\
\textcolor{blue}{layer 1 circuit 24, layer 2 circuit 14}, with effect 0.86\\
layer 0 circuit 13, layer 2 circuit 18, with effect 0.82\\
layer 1 circuit 13, layer 2 circuit 18, with effect 0.88\\
\textcolor{blue}{layer 0 circuit 19, layer 2 circuit 20}, with effect 0.72\\
\textcolor{blue}{layer 0 circuit 20, layer 2 circuit 20}, with effect 0.79\\
\textcolor{blue}{layer 0 circuit 21, layer 2 circuit 20}, with effect 0.72\\
\textcolor{blue}{layer 0 circuit 22, layer 2 circuit 20}, with effect 0.77\\
\textcolor{blue}{layer 1 circuit 19, layer 2 circuit 20}, with effect 0.75\\
\textcolor{blue}{layer 1 circuit 20, layer 2 circuit 20}, with effect 0.76\\
\textcolor{blue}{layer 1 circuit 21, layer 2 circuit 20}, with effect 0.7\\
\textcolor{blue}{layer 1 circuit 22, layer 2 circuit 20}, with effect 0.76\\
layer 1 circuit 23, layer 11 circuit 1, with effect 0.74\\
layer 1 circuit 24, layer 11 circuit 1, with effect 0.75\\
layer 2 circuit 24, layer 11 circuit 1, with effect 0.73\\
layer 4 circuit 23, layer 11 circuit 1, with effect 0.74\\
layer 0 circuit 24, layer 11 circuit 14, with effect 0.77\\
layer 1 circuit 13, layer 11 circuit 14, with effect 0.74\\
layer 1 circuit 16, layer 11 circuit 14, with effect 0.74\\
layer 1 circuit 24, layer 11 circuit 14, with effect 0.82\\
layer 2 circuit 13, layer 11 circuit 14, with effect 0.75\\
layer 2 circuit 16, layer 11 circuit 14, with effect 0.76\\
layer 2 circuit 24, layer 11 circuit 14, with effect 0.81\\
layer 3 circuit 13, layer 11 circuit 14, with effect 0.75\\
layer 3 circuit 16, layer 11 circuit 14, with effect 0.75\\
layer 3 circuit 24, layer 11 circuit 14, with effect 0.81\\
layer 4 circuit 13, layer 11 circuit 14, with effect 0.76\\
layer 4 circuit 24, layer 11 circuit 14, with effect 0.81\\
layer 5 circuit 24, layer 11 circuit 14, with effect 0.82\\
layer 6 circuit 16, layer 11 circuit 14, with effect 0.76\\
layer 6 circuit 24, layer 11 circuit 14, with effect 0.79\\
layer 7 circuit 24, layer 11 circuit 14, with effect 0.77\\
layer 8 circuit 24, layer 11 circuit 14, with effect 0.78\\
layer 9 circuit 24, layer 11 circuit 14, with effect 0.77\\
layer 10 circuit 24, layer 11 circuit 14, with effect 0.77\\
}

\textbf{Multi-Step Paths in Induction Skill}

\textit{
\textcolor{blue}{layer 0 circuit 20, layer 2 circuit 14, layer 5 circuit 11}, with effect 0.6\\
\textcolor{blue}{layer 0 circuit 21, layer 2 circuit 14, layer 5 circuit 11}, with effect 0.6\\
\textcolor{blue}{layer 1 circuit 16, layer 2 circuit 14, layer 5 circuit 11}, with effect 0.6\\
\textcolor{blue}{layer 1 circuit 18, layer 2 circuit 14, layer 5 circuit 11}, with effect 0.6\\
\textcolor{blue}{layer 1 circuit 20, layer 2 circuit 14, layer 5 circuit 11}, with effect 0.6\\
\textcolor{blue}{layer 1 circuit 21, layer 2 circuit 14, layer 5 circuit 11}, with effect 0.6\\
\textcolor{blue}{layer 1 circuit 22, layer 2 circuit 14, layer 5 circuit 11}, with effect 0.61\\
layer 0 circuit 13, layer 2 circuit 20, layer 5 circuit 11, with effect 0.6\\
\textcolor{blue}{layer 0 circuit 20, layer 2 circuit 14, layer 11 circuit 1}, with effect 0.61\\
\textcolor{blue}{layer 0 circuit 21, layer 2 circuit 14, layer 11 circuit 1}, with effect 0.63\\
\textcolor{blue}{layer 1 circuit 18, layer 2 circuit 14, layer 11 circuit 1}, with effect 0.61\\
\textcolor{blue}{layer 1 circuit 20, layer 2 circuit 14, layer 11 circuit 1}, with effect 0.61\\
\textcolor{blue}{layer 1 circuit 21, layer 2 circuit 14, layer 11 circuit 1}, with effect 0.61\\
\textcolor{blue}{layer 1 circuit 22, layer 2 circuit 14, layer 11 circuit 1}, with effect 0.63
}

\textbf{Multi-Step Paths in ICL1 Skill}

\textit{
\textcolor{blue}{layer 0 circuit 13, layer 1 circuit 19, layer 3 circuit 11}, with effect 0.81\\
\textcolor{blue}{layer 0 circuit 14, layer 1 circuit 19, layer 3 circuit 11}, with effect 0.85\\
\textcolor{blue}{layer 0 circuit 15, layer 1 circuit 19, layer 3 circuit 11}, with effect 0.84\\
\textcolor{blue}{layer 0 circuit 16, layer 1 circuit 19, layer 3 circuit 11}, with effect 0.85\\
\textcolor{blue}{layer 0 circuit 21, layer 1 circuit 19, layer 3 circuit 11}, with effect 0.82\\
\textcolor{blue}{layer 0 circuit 22, layer 1 circuit 19, layer 3 circuit 11}, with effect 0.85\\
\textcolor{blue}{layer 0 circuit 23, layer 1 circuit 19, layer 3 circuit 11}, with effect 0.84\\
\textcolor{blue}{layer 0 circuit 24, layer 1 circuit 19, layer 3 circuit 11}, with effect 0.85\\
layer 0 circuit 13, layer 2 circuit 14, layer 3 circuit 11, with effect 0.81\\
\textcolor{blue}{layer 0 circuit 20, layer 2 circuit 14, layer 3 circuit 11}, with effect 0.81\\
\textcolor{blue}{layer 0 circuit 21, layer 2 circuit 14, layer 3 circuit 11}, with effect 0.83\\
\textcolor{blue}{layer 0 circuit 22, layer 2 circuit 14, layer 3 circuit 11}, with effect 0.83\\
\textcolor{blue}{layer 1 circuit 20, layer 2 circuit 14, layer 3 circuit 11}, with effect 0.81\\
\textcolor{blue}{layer 1 circuit 21, layer 2 circuit 14, layer 3 circuit 11}, with effect 0.82\\
\textcolor{blue}{layer 1 circuit 22, layer 2 circuit 14, layer 3 circuit 11}, with effect 0.83\\
\textcolor{blue}{layer 1 circuit 23, layer 2 circuit 14, layer 3 circuit 11}, with effect 0.8\\
\textcolor{blue}{layer 0 circuit 13, layer 2 circuit 20, layer 3 circuit 11}, with effect 0.86\\
\textcolor{blue}{layer 0 circuit 14, layer 2 circuit 20, layer 3 circuit 11}, with effect 0.85\\
\textcolor{blue}{layer 0 circuit 15, layer 2 circuit 20, layer 3 circuit 11}, with effect 0.81\\
\textcolor{blue}{layer 0 circuit 16, layer 2 circuit 20, layer 3 circuit 11}, with effect 0.85\\
\textcolor{blue}{layer 0 circuit 17, layer 2 circuit 20, layer 3 circuit 11}, with effect 0.85\\
\textcolor{blue}{layer 0 circuit 18, layer 2 circuit 20, layer 3 circuit 11}, with effect 0.81\\
\textcolor{blue}{layer 0 circuit 19, layer 2 circuit 20, layer 3 circuit 11}, with effect 0.82\\
\textcolor{blue}{layer 0 circuit 20, layer 2 circuit 20, layer 3 circuit 11}, with effect 0.85\\
\textcolor{blue}{layer 0 circuit 21, layer 2 circuit 20, layer 3 circuit 11}, with effect 0.83\\
\textcolor{blue}{layer 0 circuit 22, layer 2 circuit 20, layer 3 circuit 11}, with effect 0.86\\
\textcolor{blue}{layer 0 circuit 24, layer 2 circuit 20, layer 3 circuit 11}, with effect 0.81\\
\textcolor{blue}{layer 1 circuit 13, layer 2 circuit 20, layer 3 circuit 11}, with effect 0.86\\
\textcolor{blue}{layer 1 circuit 14, layer 2 circuit 20, layer 3 circuit 11}, with effect 0.84\\
\textcolor{blue}{layer 1 circuit 15, layer 2 circuit 20, layer 3 circuit 11}, with effect 0.82\\
\textcolor{blue}{layer 1 circuit 16, layer 2 circuit 20, layer 3 circuit 11}, with effect 0.85\\
\textcolor{blue}{layer 1 circuit 17, layer 2 circuit 20, layer 3 circuit 11}, with effect 0.85\\
\textcolor{blue}{layer 1 circuit 18, layer 2 circuit 20, layer 3 circuit 11}, with effect 0.85\\
\textcolor{blue}{layer 1 circuit 19, layer 2 circuit 20, layer 3 circuit 11}, with effect 0.85\\
\textcolor{blue}{layer 0 circuit 14, layer 1 circuit 19, layer 2 circuit 20, layer 3 circuit 11}, with effect 0.83\\
\textcolor{blue}{layer 0 circuit 15, layer 1 circuit 19, layer 2 circuit 20, layer 3 circuit 11}, with effect 0.83\\
\textcolor{blue}{layer 0 circuit 16, layer 1 circuit 19, layer 2 circuit 20, layer 3 circuit 11}, with effect 0.83\\
\textcolor{blue}{layer 0 circuit 22, layer 1 circuit 19, layer 2 circuit 20, layer 3 circuit 11}, with effect 0.83\\
\textcolor{blue}{layer 0 circuit 23, layer 1 circuit 19, layer 2 circuit 20, layer 3 circuit 11}, with effect 0.82\\
\textcolor{blue}{layer 0 circuit 24, layer 1 circuit 19, layer 2 circuit 20, layer 3 circuit 11}, with effect 0.84\\
\textcolor{blue}{layer 1 circuit 20, layer 2 circuit 20, layer 3 circuit 11}, with effect 0.85\\
\textcolor{blue}{layer 1 circuit 21, layer 2 circuit 20, layer 3 circuit 11}, with effect 0.84\\
\textcolor{blue}{layer 1 circuit 22, layer 2 circuit 20, layer 3 circuit 11}, with effect 0.86\\
\textcolor{blue}{layer 1 circuit 23, layer 2 circuit 20, layer 3 circuit 11}, with effect 0.82\\
\textcolor{blue}{layer 1 circuit 24, layer 2 circuit 20, layer 3 circuit 11}, with effect 0.81\\
\textcolor{blue}{layer 0 circuit 21, layer 2 circuit 14, layer 3 circuit 14}, with effect 0.8\\
\textcolor{blue}{layer 0 circuit 22, layer 2 circuit 14, layer 3 circuit 14}, with effect 0.81\\
\textcolor{blue}{layer 1 circuit 21, layer 2 circuit 14, layer 3 circuit 14}, with effect 0.81\\
\textcolor{blue}{layer 1 circuit 22, layer 2 circuit 14, layer 3 circuit 14}, with effect 0.81\\
layer 0 circuit 13, layer 1 circuit 16, layer 10 circuit 9, with effect 0.84\\
layer 0 circuit 14, layer 1 circuit 16, layer 10 circuit 9, with effect 0.81\\
layer 0 circuit 15, layer 1 circuit 16, layer 10 circuit 9, with effect 0.8\\
layer 0 circuit 22, layer 1 circuit 16, layer 10 circuit 9, with effect 0.81\\
layer 0 circuit 14, layer 1 circuit 20, layer 10 circuit 9, with effect 0.83\\
layer 0 circuit 24, layer 1 circuit 20, layer 10 circuit 9, with effect 0.81\\
\textcolor{blue}{layer 0 circuit 13, layer 2 circuit 20, layer 10 circuit 9}, with effect 0.92\\
\textcolor{blue}{layer 0 circuit 14, layer 2 circuit 20, layer 10 circuit 9}, with effect 0.9\\
\textcolor{blue}{layer 0 circuit 15, layer 2 circuit 20, layer 10 circuit 9}, with effect 0.85\\
\textcolor{blue}{layer 0 circuit 16, layer 2 circuit 20, layer 10 circuit 9}, with effect 0.91\\
\textcolor{blue}{layer 0 circuit 17, layer 2 circuit 20, layer 10 circuit 9}, with effect 0.89\\
\textcolor{blue}{layer 0 circuit 18, layer 2 circuit 20, layer 10 circuit 9}, with effect 0.86\\
\textcolor{blue}{layer 0 circuit 19, layer 2 circuit 20, layer 10 circuit 9}, with effect 0.86\\
\textcolor{blue}{layer 0 circuit 20, layer 2 circuit 20, layer 10 circuit 9}, with effect 0.9\\
\textcolor{blue}{layer 0 circuit 21, layer 2 circuit 20, layer 10 circuit 9}, with effect 0.87\\
\textcolor{blue}{layer 0 circuit 22, layer 2 circuit 20, layer 10 circuit 9}, with effect 0.92\\
\textcolor{blue}{layer 0 circuit 23, layer 2 circuit 20, layer 10 circuit 9}, with effect 0.85\\
\textcolor{blue}{layer 0 circuit 24, layer 2 circuit 20, layer 10 circuit 9}, with effect 0.86\\
\textcolor{blue}{layer 1 circuit 13, layer 2 circuit 20, layer 10 circuit 9}, with effect 0.92\\
\textcolor{blue}{layer 1 circuit 14, layer 2 circuit 20, layer 10 circuit 9}, with effect 0.89\\
\textcolor{blue}{layer 1 circuit 15, layer 2 circuit 20, layer 10 circuit 9}, with effect 0.85\\
\textcolor{blue}{layer 1 circuit 16, layer 2 circuit 20, layer 10 circuit 9}, with effect 0.9\\
\textcolor{blue}{layer 0 circuit 13, layer 1 circuit 16, layer 2 circuit 20, layer 10 circuit 9}, with effect 0.83\\
\textcolor{blue}{layer 1 circuit 17, layer 2 circuit 20, layer 10 circuit 9}, with effect 0.9\\
\textcolor{blue}{layer 1 circuit 18, layer 2 circuit 20, layer 10 circuit 9}, with effect 0.91\\
\textcolor{blue}{layer 0 circuit 14, layer 1 circuit 18, layer 2 circuit 20, layer 10 circuit 9}, with effect 0.81\\
\textcolor{blue}{layer 0 circuit 23, layer 1 circuit 18, layer 2 circuit 20, layer 10 circuit 9}, with effect 0.83\\
\textcolor{blue}{layer 1 circuit 19, layer 2 circuit 20, layer 10 circuit 9}, with effect 0.9\\
\textcolor{blue}{layer 0 circuit 13, layer 1 circuit 19, layer 2 circuit 20, layer 10 circuit 9,} with effect 0.83\\
\textcolor{blue}{layer 0 circuit 14, layer 1 circuit 19, layer 2 circuit 20, layer 10 circuit 9}, with effect 0.87\\
\textcolor{blue}{layer 0 circuit 15, layer 1 circuit 19, layer 2 circuit 20, layer 10 circuit 9}, with effect 0.86\\
\textcolor{blue}{layer 0 circuit 16, layer 1 circuit 19, layer 2 circuit 20, layer 10 circuit 9}, with effect 0.87\\
\textcolor{blue}{layer 0 circuit 20, layer 1 circuit 19, layer 2 circuit 20, layer 10 circuit 9}, with effect 0.82\\
\textcolor{blue}{layer 0 circuit 21, layer 1 circuit 19, layer 2 circuit 20, layer 10 circuit 9}, with effect 0.82\\
\textcolor{blue}{layer 0 circuit 22, layer 1 circuit 19, layer 2 circuit 20, layer 10 circuit 9}, with effect 0.87\\
\textcolor{blue}{layer 0 circuit 23, layer 1 circuit 19, layer 2 circuit 20, layer 10 circuit 9}, with effect 0.86\\
\textcolor{blue}{layer 0 circuit 24, layer 1 circuit 19, layer 2 circuit 20, layer 10 circuit 9}, with effect 0.88\\
\textcolor{blue}{layer 1 circuit 20, layer 2 circuit 20, layer 10 circuit 9}, with effect 0.9\\
\textcolor{blue}{layer 0 circuit 14, layer 1 circuit 20, layer 2 circuit 20, layer 10 circuit 9}, with effect 0.81\\
\textcolor{blue}{layer 1 circuit 21, layer 2 circuit 20, layer 10 circuit 9}, with effect 0.89\\
\textcolor{blue}{layer 1 circuit 22, layer 2 circuit 20, layer 10 circuit 9}, with effect 0.92\\
\textcolor{blue}{layer 1 circuit 23, layer 2 circuit 20, layer 10 circuit 9}, with effect 0.86\\
\textcolor{blue}{layer 0 circuit 14, layer 1 circuit 19, layer 10 circuit 10}, with effect 0.81\\
\textcolor{blue}{layer 0 circuit 16, layer 1 circuit 19, layer 10 circuit 10}, with effect 0.81\\
layer 0 circuit 22, layer 1 circuit 19, layer 10 circuit 10, with effect 0.81\\
layer 0 circuit 23, layer 1 circuit 19, layer 10 circuit 10, with effect 0.81\\
\textcolor{blue}{layer 0 circuit 24, layer 1 circuit 19, layer 10 circuit 10}, with effect 0.82\\
\textcolor{blue}{layer 0 circuit 14, layer 1 circuit 19, layer 11 circuit 5}, with effect 0.81\\
\textcolor{blue}{layer 0 circuit 16, layer 1 circuit 19, layer 11 circuit 5}, with effect 0.8\\
layer 0 circuit 22, layer 1 circuit 19, layer 11 circuit 5, with effect 0.81\\
\textcolor{blue}{layer 0 circuit 24, layer 1 circuit 19, layer 11 circuit 5}, with effect 0.81\\
layer 0 circuit 13, layer 2 circuit 14, layer 11 circuit 5, with effect 0.87\\
layer 0 circuit 14, layer 2 circuit 14, layer 11 circuit 5, with effect 0.81\\
\textcolor{blue}{layer 0 circuit 20, layer 2 circuit 14, layer 11 circuit 5}, with effect 0.86\\
\textcolor{blue}{layer 0 circuit 21, layer 2 circuit 14, layer 11 circuit 5}, with effect 0.89\\
\textcolor{blue}{layer 0 circuit 22, layer 2 circuit 14, layer 11 circuit 5}, with effect 0.89\\
\textcolor{blue}{layer 0 circuit 23, layer 2 circuit 14, layer 11 circuit 5}, with effect 0.86\\
\textcolor{blue}{layer 0 circuit 24, layer 2 circuit 14, layer 11 circuit 5}, with effect 0.84\\
\textcolor{blue}{layer 1 circuit 13, layer 2 circuit 14, layer 11 circuit 5}, with effect 0.85\\
layer 1 circuit 14, layer 2 circuit 14, layer 11 circuit 5, with effect 0.86\\
\textcolor{blue}{layer 1 circuit 15, layer 2 circuit 14, layer 11 circuit 5}, with effect 0.85\\
\textcolor{blue}{layer 1 circuit 16, layer 2 circuit 14, layer 11 circuit 5}, with effect 0.84\\
\textcolor{blue}{layer 1 circuit 17, layer 2 circuit 14, layer 11 circuit 5}, with effect 0.85\\
\textcolor{blue}{layer 1 circuit 18, layer 2 circuit 14, layer 11 circuit 5}, with effect 0.86\\
layer 1 circuit 19, layer 2 circuit 14, layer 11 circuit 5, with effect 0.8\\
\textcolor{blue}{layer 1 circuit 20, layer 2 circuit 14, layer 11 circuit 5}, with effect 0.87\\
layer 1 circuit 21, layer 2 circuit 14, layer 11 circuit 5, with effect 0.89\\
\textcolor{blue}{layer 1 circuit 22, layer 2 circuit 14, layer 11 circuit 5}, with effect 0.89\\
\textcolor{blue}{layer 1 circuit 23, layer 2 circuit 14, layer 11 circuit 5}, with effect 0.86\\
\textcolor{blue}{layer 1 circuit 24, layer 2 circuit 14, layer 11 circuit 5}, with effect 0.81\\
layer 0 circuit 13, layer 2 circuit 24, layer 11 circuit 5, with effect 0.84\\
layer 0 circuit 14, layer 2 circuit 24, layer 11 circuit 5, with effect 0.82\\
layer 0 circuit 15, layer 2 circuit 24, layer 11 circuit 5, with effect 0.85\\
layer 0 circuit 16, layer 2 circuit 24, layer 11 circuit 5, with effect 0.85\\
layer 0 circuit 17, layer 2 circuit 24, layer 11 circuit 5, with effect 0.85\\
layer 0 circuit 22, layer 2 circuit 24, layer 11 circuit 5, with effect 0.85\\
layer 0 circuit 23, layer 2 circuit 24, layer 11 circuit 5, with effect 0.85\\
layer 0 circuit 24, layer 2 circuit 24, layer 11 circuit 5, with effect 0.82\\
layer 1 circuit 13, layer 2 circuit 24, layer 11 circuit 5, with effect 0.83\\
layer 1 circuit 14, layer 2 circuit 24, layer 11 circuit 5, with effect 0.81\\
layer 1 circuit 15, layer 2 circuit 24, layer 11 circuit 5, with effect 0.82\\
layer 1 circuit 16, layer 2 circuit 24, layer 11 circuit 5, with effect 0.81\\
layer 1 circuit 17, layer 2 circuit 24, layer 11 circuit 5, with effect 0.81\\
layer 1 circuit 22, layer 2 circuit 24, layer 11 circuit 5, with effect 0.85\\
layer 1 circuit 23, layer 2 circuit 24, layer 11 circuit 5, with effect 0.82\\
layer 1 circuit 24, layer 2 circuit 24, layer 11 circuit 5, with effect 0.81\\
layer 0 circuit 13, layer 3 circuit 14, layer 11 circuit 5, with effect 0.81\\
layer 0 circuit 23, layer 3 circuit 14, layer 11 circuit 5, with effect 0.85\\
layer 1 circuit 23, layer 3 circuit 14, layer 11 circuit 5, with effect 0.81\\
layer 2 circuit 23, layer 3 circuit 14, layer 11 circuit 5, with effect 0.8}

Almost all 3-step paths are composed of paths from lower-level skills. For instance, in the ICL skill, the sequence \textit{``layer 0 circuit 20, layer 2 circuit 14, layer 5 circuit 11"} encompasses the path \textit{``layer 0 circuit 20, layer 2 circuit 14"} from the previous token skill. Furthermore, it is apparent that the more complex a skill, the more multi-step paths it encompasses.

\section{Exploration - Why Wrong Outputs?}\label{suppexploration}

\begin{table*}
\begin{center}
\resizebox{0.8\textwidth}{!}{
\begin{tabular}{ll}
\textbf{Type}&\textbf{Top-5 circuits with absence rate}\\
\hline
\textbf{F\_IDT}& [2, 18] ($\downarrow$0.37), [2, 14] ($\downarrow$0.32), [11, 1] ($\downarrow$0.28), [2, 20] ($\downarrow$0.26), [2, 24] ($\downarrow$0.26)\\
\textbf{F1\_ICL}& [2, 24] ($\downarrow$0.45), [2, 20] ($\downarrow$ 0.42), [2, 22] ($\downarrow$ 0.41), [1, 20] ($\downarrow$0.39), [2, 14] ($\downarrow$ 0.32)\\
\textbf{F2\_ICL}&[3, 14] ($\downarrow$0.29), [4, 5] ($\downarrow$0.28), [10, 10] ($\downarrow$0.28), [8, 9]($\downarrow$0.24), [4, 12] ($\downarrow$0.22)\\
\hline

\end{tabular}}
\end{center}
\caption{Top 5 Receiver circuits appearing most frequently in skill paths presented in correct output samples but not incorrect samples.}
\label{tablosscircuit}
\end{table*}

In this section, we present a new direction for explaining and exploring common erroneous answers using Skill Circuit Graphs. Specifically, by contrasting the Skill Graphs of ``incorrect'' outputs with those of correct outputs, we can further diagnose what leads to the failure in skill execution. 
Table~\ref{tablosscircuit} illustrates the key circuits exhibiting the highest absent rate\footnote{Let $N^{+}_{C^{l,j}}$ and $N^{-}_{C^{l,j}}$ be the number of paths received by $C^{l,j}$ in correct and incorrect samples. The absence rate for each circuit is calculated as  $(N^{+}_{C^{l,j}}-N^{-}_{C^{l,j}})/N^{+}_{C^{l,j}}\in$ [0, 1].} between 3 ``incorrect'' and correct output types. Specifically, we investigate one erroneous type of output from an induction skill sample (F\_IDT), and two types from ICL skill samples (F1\_ICL, F2\_ICL). 

F\_IDT refers to those samples wherein the input possesses an Induction pattern (``\textit{A B ... A}''), but ultimately does not output \textit{B}. 
F1\_ICL denotes those samples wherein the output includes a word outside of the label options from the demonstrations, for example, a case where the input text ``\textit{[review1], label: positive, [review2], label: negative, [review3], label:}'' unexpectedly produces ``\textit{the}''. Such an error indicates that the language model did not capture the ICL template pattern in this case. 
F2\_ICL involves samples that capture the template pattern yet still produce incorrect outputs, for example, cases where the correct output should be ``\textit{positive}'', but the prediction is ``\textit{negative}''. We compare the circuit graphs of these ``incorrect" samples with the correct samples and identify the top 5 circuits with the highest absence rate.

Table~\ref{tablosscircuit} exhibits several interesting phenomena where the largest discrepancies between correct and incorrect samples in both F\_IDT and F1\_ICL occur on key circuits at layer 2. These circuits originate from the previous token skill, which handles the skill of receiving information from the previous token, such as the ``A $\rightarrow$ B'' in the induction template \textit{``A B ... A"}, as well as patterns such as ``label $\rightarrow$ positive'' in ICL. 
The loss of this skill—failure during the execution of the previous token skill—means that both the Induction skill and ICL skill cannot pass the duplicated prefix information to the next token, leading to template-based errors. 

To further understand why these samples do not successfully execute the previous token skill, we perform a bi-clustering operation on the Previous Token Skill (experiment details are shown in Appendix~\ref{suppimple}), yielding a cluster with $Eff<0.2$ across most of all paths. We compared this cluster (termed the low-effect cluster) with another cluster (named high-effect cluster), with some samples as follows (All samples are from the original text of the Previous Token Skill, tokenized into two tokens): 

\textbf{Low-effect cluster}:\textit{``About to''}, \textit{`` all these''}, \textit{`` am a''}, \textit{`` and win''}, \textit{`` and select''}, \textit{`` care over''}, \textit{``In Singapore''}, \textit{`` in the''}, \textit{`` is a''}, \textit{`` it was''}, \textit{`` than they''}, \textit{``The language''}, \textit{``The country''}, \textit{`` the movie''}

\textbf{High-effect cluster}: \textit{`` 2002''}, \textit{``Adriano''}, \textit{``Ajinomoto''}, \textit{`` becomes''}, \textit{``Could you''}, \textit{`` don’t''}, \textit{`` ended up''},  \textit{``If the''}, \textit{`` iPhone''}, \textit{`` Knowledge''}, \textit{`` stressful''}, \textit{``Windows''}, \textit{`` Youtube’s''}

It becomes obvious that in the context of an experimental setting lacking enough context, the previous token skill is performed only when there is a strong semantic relationship between the two tokens. For pairs of tokens where the semantic relation is not strong, there tends to be a reliance on the bi-gram model decision from the destination token. 

Furthermore, for F2\_ICL, the absence rate is relatively lower, suggesting that the source of the error might not be due to a single explicit cause. These circuits generally reside in the middle or even deeper layers, incorporating functions such as induction and summarization. However, to further analyze this, we would need to delve into the representational level, which for the moment goes beyond the scope of this paper.

\section{Skill Circuit Graphs}\label{suppskillgraph}
Due to large size constraints, we have only displayed the circuit graph for the Previous Token Skill. For additional skill graphs, please refer to our repository.
\begin{figure*}
  \centering
  \includegraphics[width=1\linewidth]{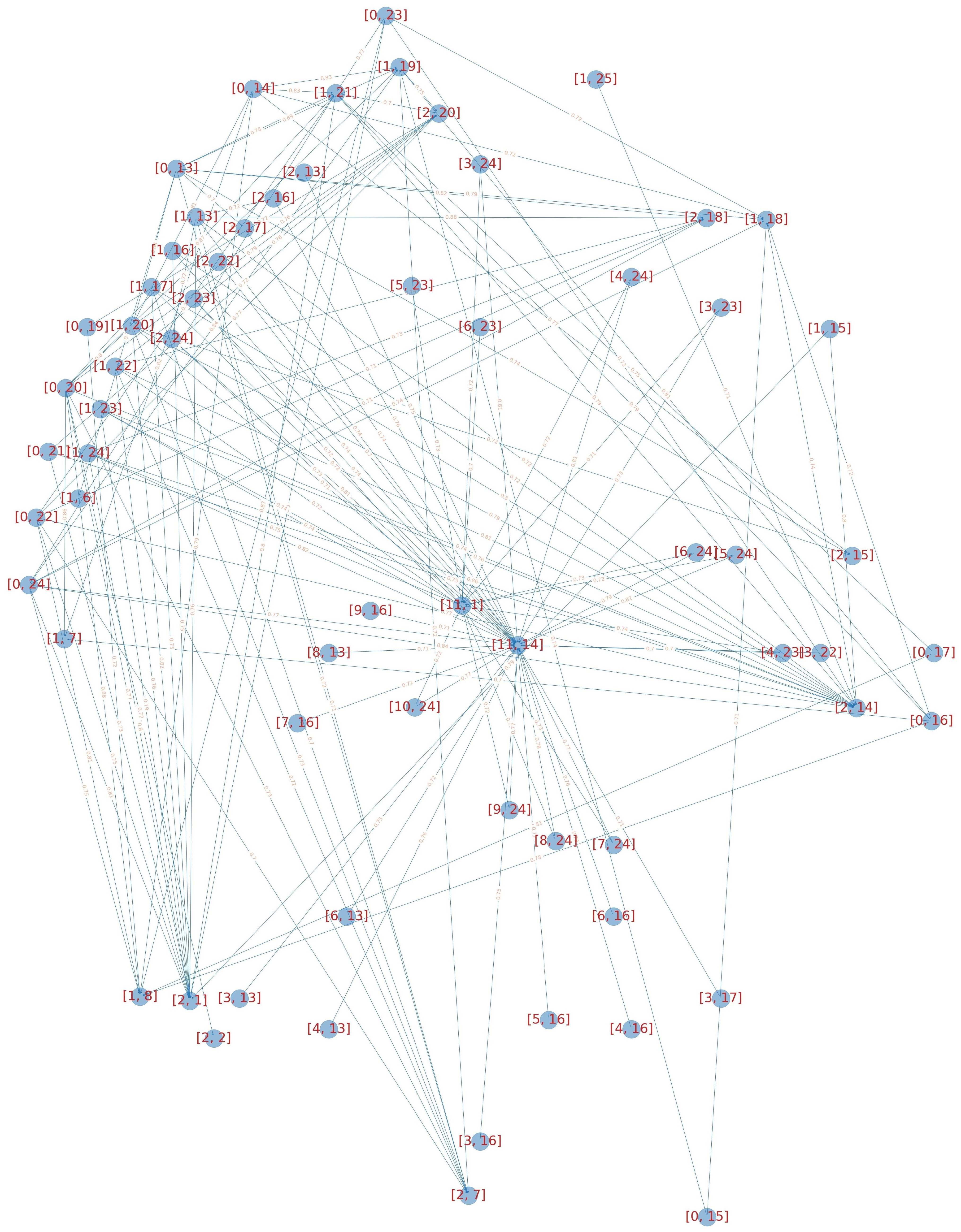}
  \caption{Skill Circuit Graph of Previous Token Skill, all paths with $Eff>0.7$ are labeled.} 
\label{figskillgraphprevioustoken}
\end{figure*}





\section{Limitation}\label{limitations}
We identify 3 limitations that need to be addressed in the future: 1) the \textbf{time complexity} of our framework; 2) \textbf{scalability}; 3) the lack of further examination of the \textbf{representational study}.

Assuming the time for an inference using LLM as $O(1)$, the time complexity of a search round would then be $O(L^2N^2)$, i.e., the square of the number of layers times the number of components. If we can overlook this time-consuming process, $\mathcal{G}*$ for each input would effectively facilitate training. In other words, $\mathcal{G}*$ could directly instruct LLM which paths are essential and which are not, thus reducing the training process. Despite the time complexity, we recall our contribution to the analysis of LMs, which is usually more challenging and does not require large-scale inference. 

We also recognize the limitations of testing on a single model and specific skills. Although many studies have validated the GPT-2 series to have public trustworthiness for mechanistic interpretability research, making us confident in its ability to support our contribution, pioneering work in discovering the theoretical foundation and experimental design of language skills, there remains ample scope for scalability across a variety of models and skills for future work. 

Finally, the lack of research at the representational level hinders our progress in answering more complex questions such as why certain samples fail to trigger a skill. Recognized that this is a rather challenging topic, we leave it as a promising future work. 

\end{document}